\documentclass[lettersize,journal]{IEEEtran}
\usepackage{amsmath,amssymb,amsfonts}
\usepackage{algorithmic}
\usepackage{algorithm}
\usepackage{array}
\usepackage{textcomp}
\usepackage{stfloats} 
\usepackage{amssymb}
\usepackage{url}
\usepackage{color}
\usepackage{verbatim}
\usepackage{graphicx}
\usepackage{epstopdf}
\usepackage{epsfig}
\usepackage{cite}
\usepackage{graphicx}
\usepackage{subfloat}
\usepackage{subfigure}
\usepackage{diagbox}
\usepackage{multirow}
\usepackage{booktabs}
\usepackage{makecell}
\usepackage{bm}
\usepackage{float}
\usepackage{tabularx} 
\usepackage[colorlinks, citecolor=red]{hyperref} 
\graphicspath{ {./authors/} }
\graphicspath{ {./PDF/} }
\usepackage{hyperref}

\begin{document}

\markboth{Journal of \LaTeX\ Class Files,~Vol.~xx, No.~xx, xxxx~xxxx} %
{Shell \MakeLowercase{\textit{et al.}}: A Sample Article Using IEEEtran.cls for IEEE Journals}

\IEEEpubid{0000--0000/00\$00.00~\copyright~20xx}

\title{A Robust Multisource Remote Sensing Image Matching Method Utilizing Attention and Feature Enhancement Against Noise Interference}

\author{Yuan~Li, 
		Dapeng~Wu,~\IEEEmembership{Senior Member,~IEEE,}
        Yaping~Cui,
        Peng~He,
		Yuan~Zhang,
        and~Ruyan~Wang  % <-this % stops a space

\thanks{Yuan Li, Dapeng Wu, Yaping Cui, Peng He, Yuan Zhang and Ruyan Wang are with the School of Communications and Information Engineering, Chongqing University of Posts and Telecommunications, Chongqing 400065, China, 
and with Advanced Network and Intelligent Connection Technology Key Laboratory of Chongqing Education Commission of China, 
Chongqing Key Laboratory of Ubiquitous Sensing and Networking. 
(e-mail: d210101010@stu.cqupt.edu.cn;  wudp@cqupt.edu.cn; cuiyp@cqupt.edu.cn; hepeng@cqupt.edu.cn; zhangyuan@cqie.edu.cn; wangry@cqupt.edu.cn).}
\vspace{-8mm}}

\maketitle 

\begin{abstract}
  Image matching is a fundamental and critical task of multisource remote sensing image applications. 
  However, remote sensing images are susceptible to various noises. 
  Accordingly, how to effectively achieve accurate matching in noise images is a challenging problem. 
  To solve this issue, we propose a robust multisource remote sensing image matching method utilizing attention and feature enhancement against noise interference. 
  In the first stage, we combine deep convolution with the attention mechanism of transformer to perform dense feature extraction, constructing feature descriptors with higher discriminability and robustness. 
  Subsequently, we employ a coarse-to-fine matching strategy to achieve dense matches. 
  In the second stage, we introduce an outlier removal network based on a binary classification mechanism, which can establish effective and geometrically consistent correspondences between images; 
  through weighting for each correspondence, inliers vs. outliers classification are performed, as well as removing outliers from dense matches. 
  Ultimately, we can accomplish more efficient and accurate matches. 
  To validate the performance of the proposed method, we conduct experiments using multisource remote sensing image datasets for comparison with other state-of-the-art methods under different scenarios, 
  including noise-free, additive random noise, and periodic stripe noise. 
  Comparative results indicate that the proposed method has a more well-balanced performance and robustness. 
  The proposed method contributes a valuable reference for solving the difficult problem of noise image matching.
\end{abstract}


\section{Introduction}

\IEEEPARstart{R}{EMOTE} sensing image matching refers to the pixel-wise identification and alignment of the same scene images acquired from different platforms or different sensors 
(e.g., satellites, airborne platforms, unmanned aerial vehicle (UAV), infrared, multispectral, and synthetic aperture radar (SAR)). 
It is a crucial step in image fusion, change detection, image navigation, and other related applications \cite{ref1,ref2,ref3}. %[1-3]

\begin{figure}[!t]
	\centering
	\hspace{-0.1cm}\subfigure[]{  % Optical-Depth
	\begin{minipage}[b]{0.23\textwidth}
	\includegraphics[width=1\linewidth,height=20mm]{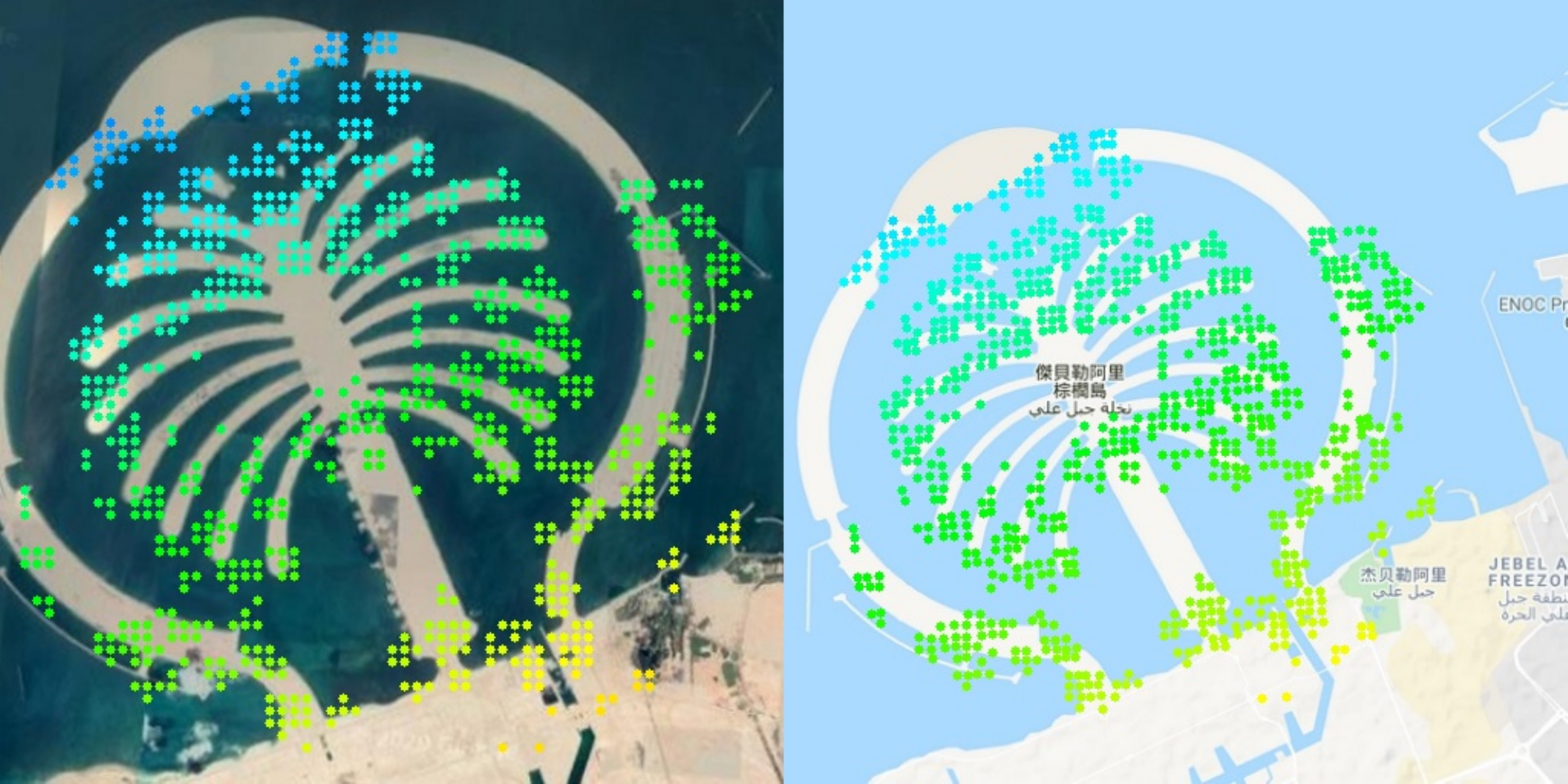}%\vspace{1pt}
	\end{minipage}}
	\hspace{-0.1cm}\subfigure[]{  %Infrared-Optical
	\begin{minipage}[b]{0.23\textwidth}
	\includegraphics[width=1\linewidth,height=20mm]{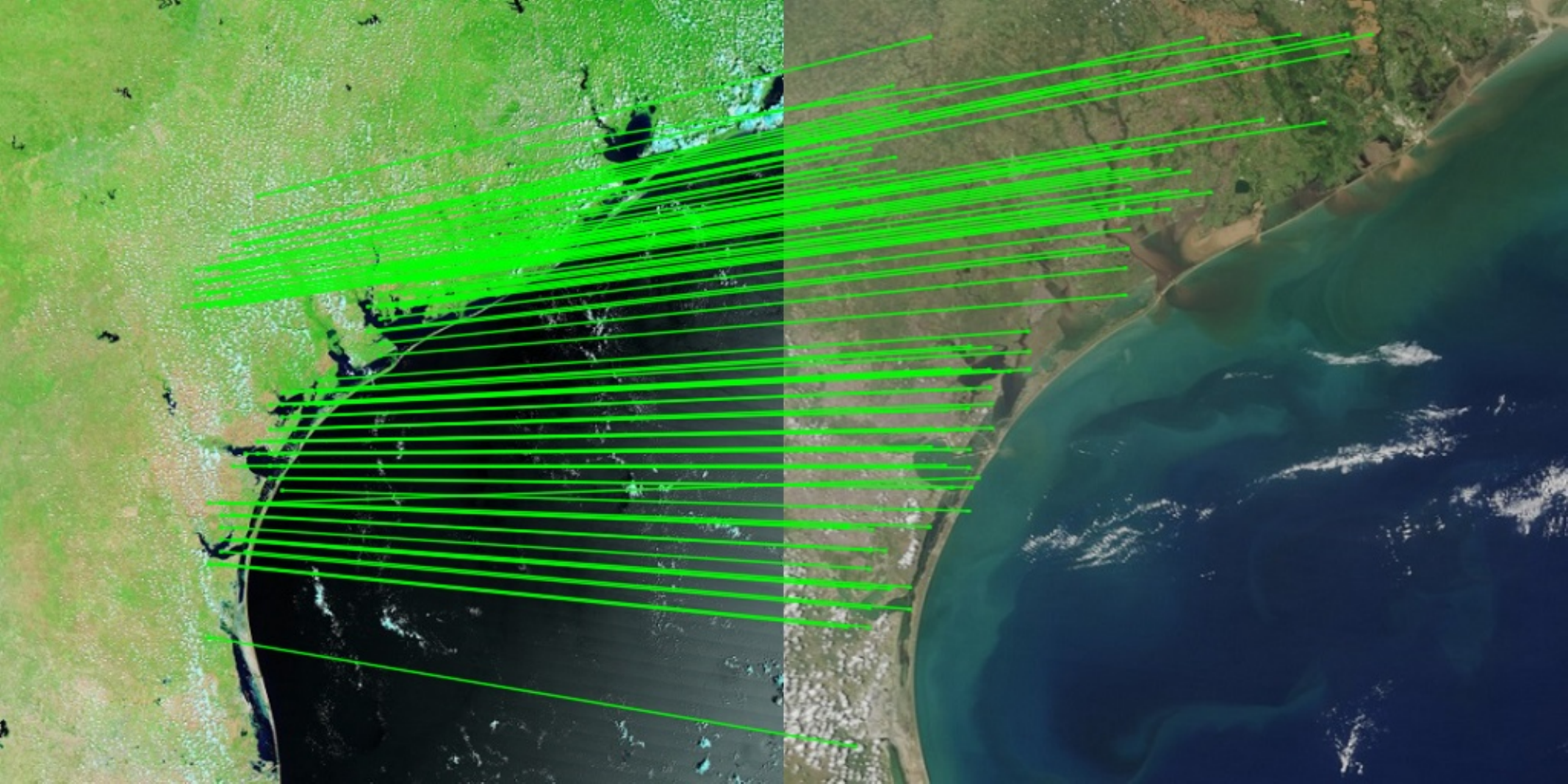}%\vspace{1pt}
	\end{minipage}}
	\hspace{-0.1cm}\subfigure[]{  % Optical-Map
	\begin{minipage}[b]{0.23\textwidth}
	\includegraphics[width=1\linewidth,height=20mm]{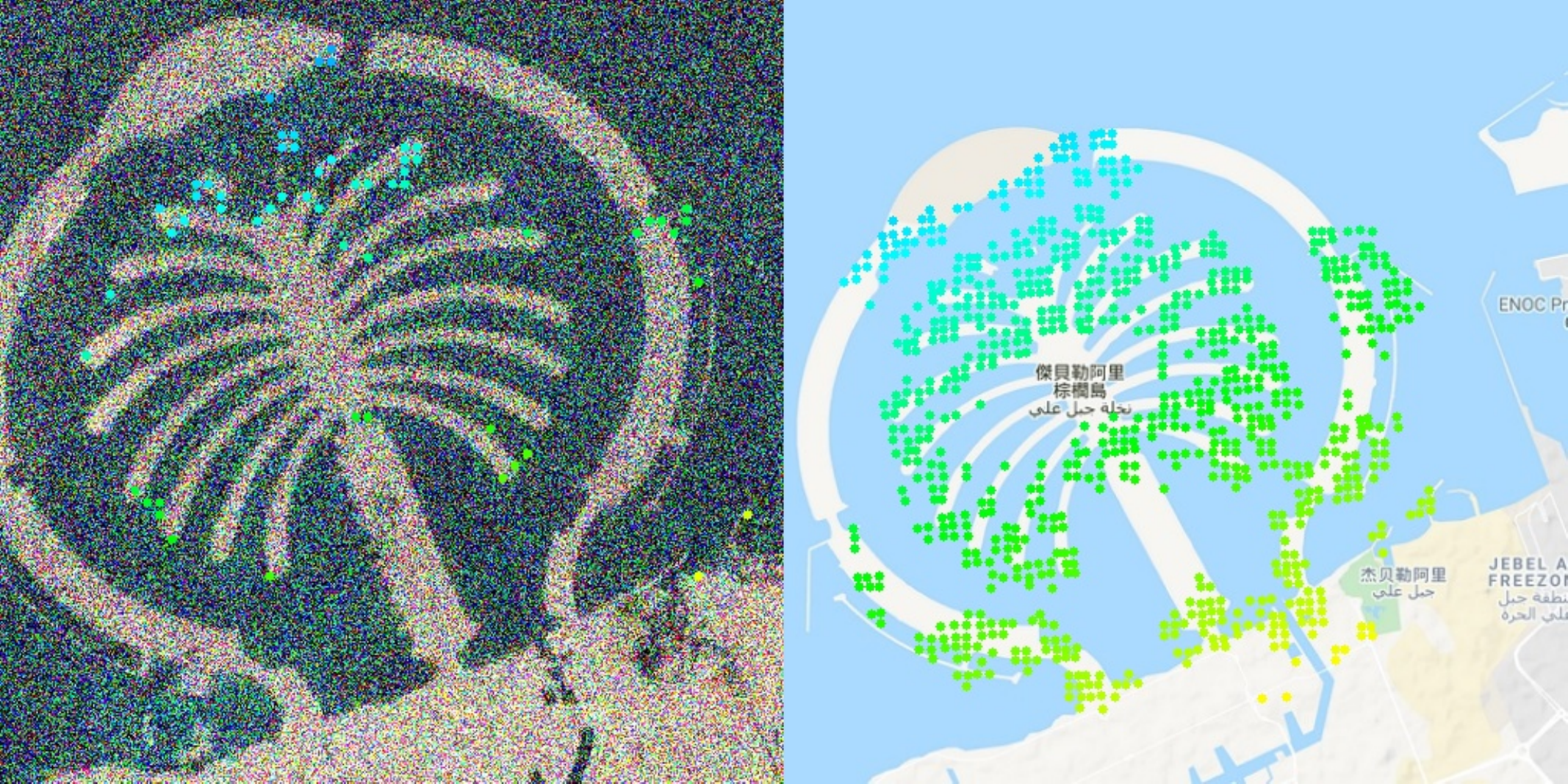}%\vspace{1pt}
	\end{minipage}}
	\hspace{-0.1cm}\subfigure[]{  % SAR-Optical
	\begin{minipage}[b]{0.23\textwidth}
	\includegraphics[width=1\linewidth,height=20mm]{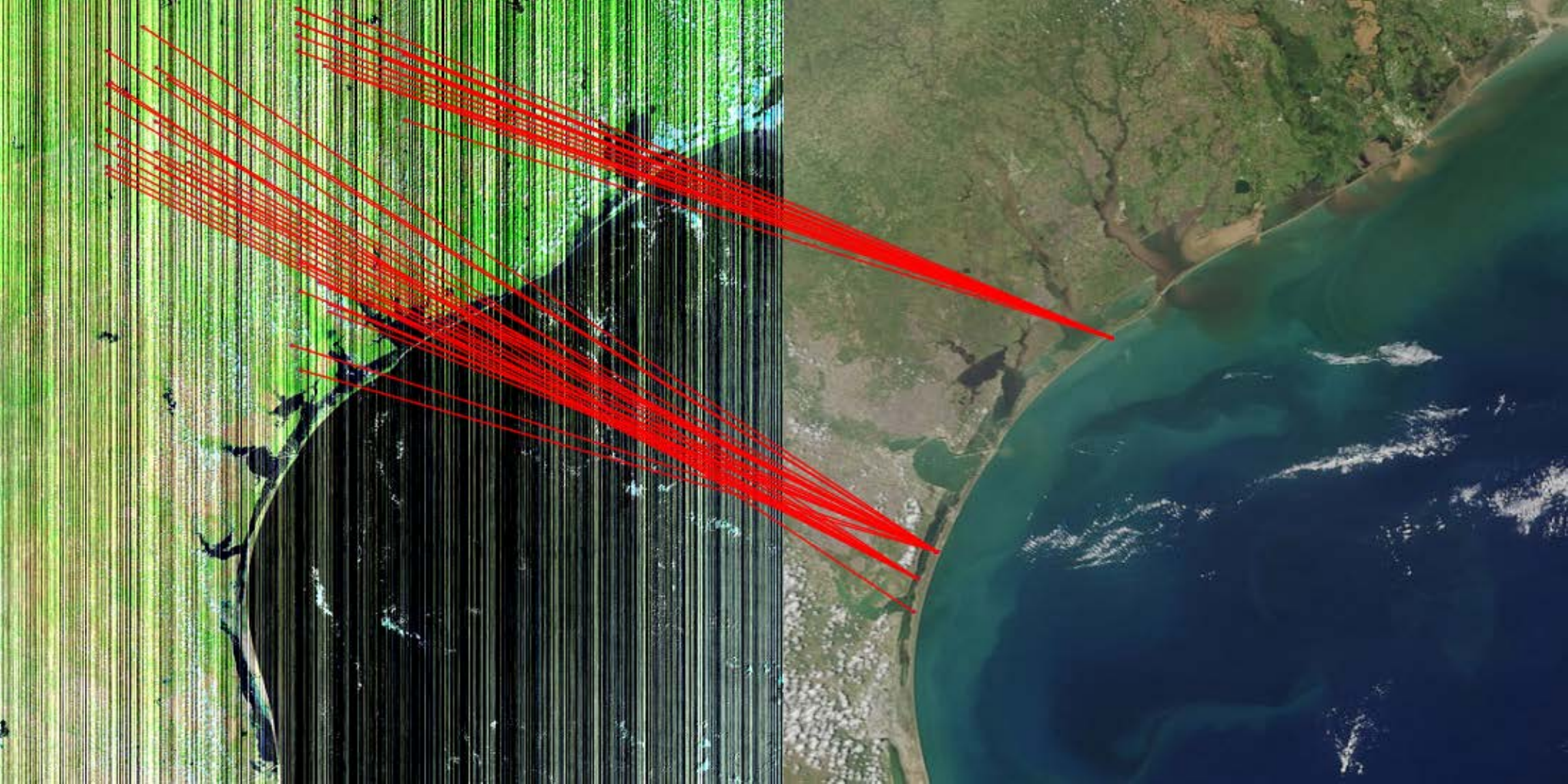}%\vspace{1pt}
	\end{minipage}}
	\caption{The effect of noise on image matching. Query image on the left, reference image on the right. (a) Noise-free feature extraction. (b) Noise-free feature matching. (c) Noise interferes with feature extraction. (d) Noise interferes with incorrect matching.}
	\label{Fig.0}
\end{figure}

In reality, a prominent problem in remote sensing image matching is noise interference. 
Noise in remote sensing images arises from various sources, including sensor thermal noise and electromagnetic interference noise \cite{ref4}, 
which primarily manifest in the form of  additive random noise, periodic stripes, and bright lines \cite{ref5}. 
As shown in Fig. \ref{Fig.0}, the existence of noise will degrade the quality of the images and impact the results of image matching. 
We summarize these effects in the following aspects:

\begin{itemize}
\item{1) Failure to extract effective features: Noise will obscure the structure and detail of the images and interfere with feature detection and extraction, which prevents the matching method from recognizing meaningful feature points in the noise image, rendering dense matching difficult to accomplish.} 
\item{2) Incorrect matches: Noise may bring in false features or change the descriptors of real features in the images. Matching methods would erroneously pair them with noise features in another image, leading to incorrect matches.} 
\IEEEpubidadjcol
\noindent 
\item{3) Reduce robustness: Robustness refers to the ability of image matching methods to perform consistently and accurately under varying conditions, including the presence of noise. Noise can introduce uncertainties and inconsistencies, decreasing the robustness of image matching methods. There will be more challenges, especially for multisource remote sensing images.}
\end{itemize}

These effects pose a daunting challenge to current image matching methods, which severely impact matching performance and hardly satisfy the demands of practical applications. 
Although image denoising is performed before image matching, different image denoising methods would have certain limitations, for example, denoising methods for optical images may not be applied to infrared or SAR images. 
In addition, validating the robustness of a matching method in the presence of noise interference is equivalent to the stability of the image processing system when the sensor equipment fails. 
Therefore, it is necessary to research more efficient, universal, and robust image matching methods for multisource remote sensing images subject to noise interference. 
In order to achieve universal and robust image matching, two key problems need to be addressed: 

\begin{itemize}
\item{1) how to achieve dense matching of images against noise interference.}
\item{2) how to design more efficient and accurate outlier removal methods.}
\end{itemize}

The handcrafted-based traditional matching methods for remote sensing images can be broadly divided into two categories: feature-based methods and area-based methods. 
Feature-based matching methods extract salient features (i.e., points, edges, or texture areas) from images and establish correspondences, 
representative ones of which include SAR-SIFT \cite{ref6}, LIFT \cite{ref7}, PSO-SIFT \cite{ref8}, and OS-SIFT \cite{ref9}. %RIFT \cite{ref6}, LNIFT \cite{ref7}, CoFSM \cite{ref8}, and R2FD2 \cite{ref9}. 
These methods can solve rotation and scale consistency issues in image matching. 
Nevertheless, they fail to be effectively applied to multisource remote sensing images with noise and distinct differences. 
Area-based methods (such as HOPC \cite{ref10}, PSCD \cite{ref11}, CFOG \cite{ref12}, and MS-HLMO \cite{ref13}) employ template matching to assess the similarity of the image patch and identify the center point with the best similarity as the corresponding point. 
These methods are able to mitigate the impact of noise on feature extraction, and the matching accuracy is further improved. 
However, it is hard for traditional area-based matching methods to fulfill the requirements of both accuracy and speed simultaneously. 

Currently, with the promotion of deep learning technology, an increasing number of deep learning networks are being explored for image matching. 
Deep learning-based image matching networks can directly predict pixel-level matching from images without explicitly detecting and describing features, and to accommodate the impact of nonlinear radiation and noise factors \cite{ref14,ref15}. 
As a representative example, the latest transformer method \cite{ref16} utilizes attention mechanisms to focus on specific elements and attributes, and perform global and local feature aggregation at the same time, which can effectively improve the performance of image matching methods. %\cite{ref16,ref17}
Xie \emph{et al} \cite{ref17} proposed a global-local multiscale perception module (GMPM), which incorporates abundant multiscale local context information into global feature representation by employing multiple depth-wise convolutions with varying kernel sizes, thereby generating discriminative features that are robust to scale shifts.  
Zhang \emph{et al} \cite{ref16} proposed the transformer-based structure can extract features including local gray-level information and global context information, and has strong robustness for multimodal image matching. 
Ye \emph{et al} \cite{ref18} take advantage of the attention module based on network learning to obtain a more refined and robust structural feature description model. Such an approach can enhance the shared feature presentation in weak structure or texture regions. 
Instead of performing image feature detection, description, and matching sequentially, Sun \emph{et al} \cite{ref19} proposed to first establish pixel-wise dense matches at a coarse level and later refine the good matches at a fine level. 
Liu \emph{et al} \cite{ref20} proposed that mismatch removal is achieved by using a neighborhood consistency strategy to eliminate outliers in the putative correspondence. The key to this strategy is to add physical constraints to localized regions to construct unique matching representations for feature points.

To tackle the key problems in noise image matching, based on the research and analysis of the literature \cite{ref16,ref17,ref18,ref19,ref20} and the extensive experiments we have conducted, we design the following strategies for implementing noise image matching:
\begin{itemize}
\item{Strategy 1. Multi-level spatial details are conducive to accurate localization of keypoints. we can use depth-separable convolution for feature preprocessing to increase multi-scale feature information and highlight local features on different scales \cite{ref17}.}
\item{Strategy 2. With the introduction of the transformer attention mechanism, alternate combinations of multilevel self-attention and cross-attention are used to construct feature descriptors of higher discriminability and robustness \cite{ref16} \cite{ref18}.}
\item{Strategy 3. As noise interferes with the local feature distribution, which results in low matching accuracy. Adopting a coarse and fine matching strategy for selective suppression of constraints helps to extract as many reliable keypoints as possible in a noise image \cite{ref19}.}
\item{Strategy 4. Noise destroys the correspondence of local geometric consistency between images and increases false matches. Therefore, it is essential to introduce an outlier removal network, which establishes geometric consistency by extracting global-local information and identifies the true inliers from the established putative correspondences sets, as well as removing outliers from sets \cite{ref20}.}
\end{itemize}

In accordance with the above strategies, a novel robust method based on deep learning is presented in this paper. 
The main contributions of this work can be concluded as follows :
\begin{itemize}
\item{1) The existing works on multisource/multimodal remote sensing image matching have mainly focused on the following problems: (i) geometric distortion, (ii) non-rigid matching, and (iii) nonlinear radial distortion (NRD).  
The main core of this paper is anti-noise interference that exploring the robustness and accuracy of multisource image matching under noise interference. 
In this regard, we propose key problems, design strategies, and implementation framework for solving noise image matching based on the analysis of the noise interference mechanism.} 
\item{2) A robust image matching framework is proposed against noise interference, 
which combines deep convolution with the transformer-based attention mechanism for dense feature extraction, constructing feature descriptors with higher discriminability and robustness. 
Even if there is a large number of noise and periodic stripes in the image, it is still able to extract prominent feature points.}
\item{3) An outlier removal network based on a binary classification mechanism is developed, 
which formulates the outlier removal task as an inlier/outlier classification problem. 
The outlier removal network exploits multi-layer perceptrons to extract the feature point geometry information after fine matching estimates, 
and utilizes context normalization to obtain global and local information, establishing effective and geometrically consistent image correspondences; 
through weighting for each correspondence, inliers vs. outliers classification are performed, as well as removing outliers from dense matches.}
\item{4) Aiming at the common additive random noise and periodic stripe noise in remote sensing images, 
we design noise image matching experiments to validate the robustness of the matching methods by progressively increasing the noise intensity. 
Experimental results on high-intensity noise interference show that our method achieves a matching success rate of over 40\% in noise images. 
The proposed method has a more well-balanced performance and robustness, and contributes a valuable reference for solving the difficult problem of noise image matching.
% with an average correct matching rate of more than 5\% higher than competing methods.
}
\end{itemize}

The rest of this paper is organized as follows. 
Section II reviews the related works of image matching research. 
In section III, the proposed robust image matching method is introduced in detail. 
Section IV designs experiments on noisy image matching and conducts ablation studies. 
The conclusions are drawn in section V.

\section{Related Works}
In this section, we review the closely related works in robust image matching, 
including the attention mechanism, image dense matching, outlier removal methods, and relation to existing works and discussion.
% including the attention mechanism for enhancing the feature descriptors, image dense matching, and outlier removal methods. 

\subsection{Attention Mechanism}

Learning-based attention mechanism have shown great advantages in computer vision and image processing. 
The core idea is to highlight the important features of images, focus on specific parts, suppress other secondary information, and improve the effectiveness and efficiency of the network model. 
Vaswani \emph{et al}. \cite{ref21} first proposed the self-attention mechanism, which is widely used in natural language processing and computer vision due to its effectiveness and efficiency. 
It can effectively improve the performance for various computer vision tasks such as semantic segmentation, target detection and image generation \cite{ref22}. 
transformer-based SuperGLue \cite{ref23} and LoFTR \cite{ref24} were firstly proposed as representative works in the field of image matching. 
SuperGlue employs self-attention and cross-attention to extract intra-image and inter-image feature information in sparse feature matching tasks. 
LoFTR, considering the computational and memory costs, employs linear attention \cite{ref25} to aggregate global features in the coarse matching stage. 
Quan \emph{et al}. \cite{ref26} proposed a deep feature correlation learning network (Cnet) for multimodal remote sensing image registration. 
With the aid of deep convolutional network and attention learning module, Cnet can utilize invariant and discriminative features to improve the performance of deep descriptors in image patch matching and image registration. 
Guo \emph{et al}. \cite{ref27} constructed a graph context attention-based network with maximum pooling aggregation strategy and ring convolutional aggregation strategy to obtain graph context with highly reliable fine-grained structural information to improve the performance of feature matching. 
Qin \emph{et al}. \cite{ref28} proposed a geometric transformer to learn geometric features for robust point cloud matching. Distances and angles between feature points are encoded to be independent of rigid transformations and robust to low overlap.

Inspired by the above works, we adopt an alternate combination of transformer-based multilevel self-attention and cross-attention for dense feature enhancement, which fuses a wider range of local and global information into the generation of feature descriptors, enabling the enhanced feature descriptors to have higher discriminability and robustness.

\subsection{Image Dense Matching}
The purpose of image dense matching is to produce pixel-level correspondences. Compared to sparse methods, dense methods are able to preserve the inherent uncertainty and make more robust decisions, which is a necessary prerequisite for dense 3D surface reconstruction \cite{ref29}, pixel-level image fusion \cite{ref30}, and target detection \cite{ref31}. 
Dense matching methods are widely studied in the field of computer vision and remote sensing. 
Xiang \emph{et al}. \cite{ref32} proposed a dense image registration method based on optical flow that combines the advantages of sparse features and dense optical flow framework to realize the registration of optical images with synthetic aperture radar (SAR) images. 
Zhang \emph{et al}. \cite{ref33} proposed a robust deep optical flow framework for dense registration of optical and SAR images. 
Rocco \emph{et al}. \cite{ref34} proposed an end-to-end network for dense matching that identifies spatially consistent matching sets by analyzing the 4D spatial neighborhood consistency of all possible correspondences between pairs of images, and does not require global geometric modeling. 
However, the computational cost to establish dense correspondences between features is vast, especially for high-resolution images. 
Another technique to reduce the matching computation is to perform dense correspondence estimation in a coarse-to-fine strategy. 
Li \emph{et al}. \cite{ref35} proposed a dual-resolution correspondence network that performs 4D correlation between coarse-resolution feature maps, the selected coarse-resolution matching scores allow the fine-resolution features to focus on a limited number of possible matches with high confidence. 
This method improves the matching accuracy while reducing the computational cost for matching high-resolution images. 
Truong \emph{et al}. \cite{ref36} find pixel-level correspondences by using global and local features extracted from images of different resolutions. 
Zhou \emph{et al}. \cite{ref37} replaced pixel-level matching with an image patch matching method, which was then refined by a regression layer. 
Truong \emph{et al}. \cite{ref38} proposed an augmented probabilistic dense correspondence network, which is capable of estimating accurate dense correspondences as well as reliable confidence mappings. 
However, none of the above methods consider and analyze the effect of noise on image matching.

In this paper, we combine deep convolution with the attention mechanism for dense feature extraction and construction of feature descriptors with higher discriminability and robustness, which can generate dense feature points even with noise interference; meanwhile, we adopt the coarse-to-fine strategy to get dense matches. 

\subsection{Outlier Removal}

Traditional outlier removal methods, such as RANSAC \cite{ref39}, MLESAC \cite{ref40}, PROSAC \cite{ref41}, USAC \cite{ref42}, and MAGSAC \cite{ref43}, use a generative and validation framework to remove outliers. 
RANSAC assumes that two images are related by certain parametric geometric relationships, such as epipolar geometry and homography. 
This approach aims to find the smallest subset of inliers to estimate predefined model parameters by repeated sampling. 
MLESAC is used to solve the image geometry problem. PROSAC selects the putative correspondences with high similarity scores. 
USAC provides a generalized framework by integrating the multiple strengths of the RANSAC variants. 
MAGSAC proposed $\sigma-$consistency in place of user-defined internal thresholds. 
RANSAC and its variants are effective methods for removing outliers when the percentage of inliers is high. 
However, when the percentage of outliers dominates, the sampled subset will inevitably be filled with a large number of outliers, which will cause the method performance to degrade and instability. 
For adapting to more complex scenarios, scholars research various neighborhood consistency methods, also known as relaxation methods, 
with representative studies including local preserving matching (LPM) \cite{ref44}, its improved version guided local preserving feature matching (GLPM) \cite{ref45}, and frame-based local preserving matching (F-LPM) \cite{ref46}, etc. 
These methods majorly focus on the localization of each correspondence rather than the global transformation, since the potential true correspondences hold stable local geometric relations. 
Normally, neighborhood consistency-based methods firstly establish an appropriate manual descriptor to characterize the similarity of two local structures, then remove outliers with large differences in local structures by various detection techniques. 
Nonetheless, the above local preserving matching methods mostly consider the simple spatial neighborhood relationship of the feature points and ignore the differences of the neighborhood elements themselves. 
When there are many outliers around the inliers, the accuracy will be greatly reduced.

Recently, deep learning approaches have achieved good performance in the field of image matching. 
Yi \emph{et al}. \cite{ref47} made the first attempt to introduce a learning-based technique termed as learning to find good correspondences (LFGC-Net), which aims to train a network in an end-to-end fashion to label the correspondences as inliers or outliers. 
Zhang \emph{et al}. \cite{ref48} devised a hierarchical graph-based neural network to learn the establishment of the two-view geometric matching, called OA-Net. 
Ma \emph{et al}. \cite{ref49} transformed outlier removal into a binary classification problem by learning a generalized classification mechanism to determine the correctness of arbitrarily assumed matches, but it still used a traditional learning paradigm and relied heavily on handcrafted feature construction. 
Chen \emph{et al}. \cite{ref50} proposed a local structured visualization-attention network that transformed outlier detection into dynamic visual similarity evaluation. 
Guo \emph{et al}. \cite{ref51} proposed a graph context attention-based network (GCA-Net) to distinguish inliers from outliers by efficiently leveraging graph context. 
Conversely, these approaches only preserve the correct matches that conform to neighborhood consistency, resulting in incomplete and inaccurate matching results.

Different from the existing design methods, we utilizes a multi-layer perception to extract the feature point geometry information in the fine matching estimation, use context normalization to obtain global and local information, establish valid and geometrically consistent correspondences between images, and then eliminate the incorrect matches.

\begin{table*}[hbp]
	\centering
	\caption{IMPORTANT ABBREVIATIONS AND NOTATIONS}
	\begin{tabularx}{\textwidth}{|lX|lX|lX|}
		\toprule
		Symbol &Explanation &Symbol &Explanation &Symbol &Explanation \\
		\midrule
		FPM   &feature preprocessing module  &DSC  &depth-wise separable convolution &$D{{W}_{1}},D{{W}_{3}}$  &depth-wise convolution with kernel sizes of 1 and 3 \\
		PE   &position encoding  &NPE  &normalizing position encoding  &FPN  &feature pyramid network \\
		$I^A,I^B$    &input images  &${{\tilde{F}}^{A}},{{\tilde{F}}^{B}}$   &coarse-level features &${{\hat{F}}^{A}},{{\hat{F}}^{B}}$  &fine-level features \\
		$\tilde{C},\hat{C}$   &coarse-level and fine-level features dimensions  &$\tilde{F}_{FPM}^{A},\tilde{F}_{FPM}^{B}$ &preprocessed coarse-level features &${{\tilde{F}}^{A^{\ast}}},{{\tilde{F}}^{B^{\ast}}}$  &normalized position encoding features \\
		$\mathsf{\mathcal{A}}$   &attention operations  &${S}$  &correlation score matrix &$^{L}{{\tilde{F}}^{A^{\ast}}},^{L}{{\tilde{F}}^{B^{\ast}}}$ &enhanced coarse-level feature \\
		${P}_{i,j}^{c}$   &probability confidence matrix for coarse matching  &$\tau_c $  &coarse matching confidence threshold &$\mathsf{\mathcal{M}}^{c}$  &coarse matching sets   \\
		$\bm{\mathbf{\tilde{x}}_{i,j}^{A}},\bm{\mathbf{\tilde{x}}_{i,j}^{B}}$   &the pixel coordinates of coarse matching  &$\hat{P}_{i,j}^{A},\hat{P}_{i,j}^{B}$  &local feature patches in fine-level feature maps &$w_{f}$  &the size of local feature patch \\
		${\mathsf{\mathcal{Z}}}_{l}$  &local heatmap &$ \mathbb{E} \left( \cdot \right)$  &spatial expectation   &${\mathsf{\mathcal{M}}}$    &pixel-level matching sets \\
		$\mathsf{\mathcal{P}}_{i,j}^{f}$   &probabilistic confidence matrix for fine matching   &${\mathsf{\mathcal{W}}}$  &weight sets of outlier removal network &${{\mathsf{\mathcal{L}}}_{\text{coarse}}}$  &coarse matching loss \\
		${{\mathsf{\mathcal{L}}}_{\text{fine}}}$   &fine matching loss  &${{\mathsf{\mathcal{L}}}_{\phi}}$  &outlier removal network loss &${\mathsf{\mathcal{M}}}_{gt}^{c}$  &ground truth coarse matching sets \\ 
		${P}_{gt}^{c}$   &ground truth confidence matrix  &${\widehat{\xi}}_{i,j}$  &reprojection error &${\mathsf{\mathcal{L}}}_{cls}$  &classification loss \\
		$\hat{E}$  &regressed essential matrix &$E_{gt}$  &ground truth essential matrix &${\mathsf{\mathcal{L}}}_{ess}$   &regression loss of essential matrix \\  
		\bottomrule
	\end{tabularx}
	\label{TABLE 0}
\end{table*}%

\subsection{Relation to Existing Works and Discussion}

Multisource/multimodal remote sensing images may contain scale, rotation, radiation, noise, blurring, or temporal changes \cite{ref57}; these differences directly affect the accuracy of image matching. 
Most of the existing work on multisource/multimodal remote sensing image matching has investigated the following three aspects: (i) geometric distortion, (ii) non-rigid matching, and (iii) nonlinear radial distortion (NRD). We discuss each of them in turn.

Geometric distortion refers to that the geometric positions, shapes, sizes, dimensions, orientations, and other features on the acquired images are not consistent with the corresponding ground objects in different remote sensing imaging modes and states. 
Li \emph{et al}. \cite{ref52} proposed a scale and rotation invariant feature transform (SRIF) method for multimodal image matching. SRIF obtains the scale of the keypoints by projecting them into a simple pyramidal scale space. Then a local intensity binary transform (LIBT) is proposed to enhance the structure information inside multimodal images, which enables the feature descriptors to be well distinguishable. 
Quan \emph{et al}. \cite{ref53} proposed a novel coarse-to-fine deep learning registration framework for multimodal remote sensing images. In the coarse registration stage, an effective deep ordinal regression (DOR) network is designed to transforms the rotation correction task into a rotational ordinal regression problem, which utilizes the correspondence between rotational ordinals to improve the accuracy of rotation estimation. In the fine registration stage, a deep descriptor learning (DDL) network is employed to deal with image modal changes based on rotationally corrected images. This method effectively solves the global geometric deformation caused by large rotational transformations.

Non-rigid matching is an important type of image matching, which refers to the process of matching images that involves complex local deformation or bending between images and is not limited to rigid transformations (translation, rotation and scaling). 
Xu \emph{et al}. \cite{ref54} proposed a nonrigid bidirectional registration network (NBR-Net) to estimate the flow-based dense correspondence for remote sensing images, which designd a bidirectional network by registering Image A to Image B and then back to Image A to strengthen the geometric consistency and reversibility. 
Shi \emph{et al}. \cite{ref55} proposed an unsupervised content-focused hierarchical alignment network (CHA-Net) for multimodal remote sensing images nonrigid matching, which is constructed based on the theory of domain adaptation. CHA-Net is a hierarchical refinement model that utilizes the Field Calibration Module (FCM) to align features with different scales separately and progressively generate the registration fields. 
Xiao \emph{et al}. \cite{ref56} introduced a multimodal remote sensing image matching method, called affine and deformable registration networks (ADRNet), which consists of affine and deformable registration stages. Firstly, the image pairs go through the affine registration network to learn the affine transformation parameters for coarse registration of moving images. Secondly, the results are then fed into a deformable registration network to obtain the fine registration results. ADRNet is also able to address both small-scale and large-scale deformations between image pairs. 

Nonlinear radiometric distortion refers to the nonlinear difference between the spectral radiation of the ground objects and the sensor detection in the remote sensing imaging process, which is caused by the influence atmosphere and illumination conditions. 
Li \emph{et al}. \cite{ref57} proposed a radiation insensitive feature matching method based on phase congruency (PC) and a maximum index map (MIM), which is called radiation-variation insensitive feature transform (RIFT). RIFT reduces the dimensionality of the descriptors by summing the convolution results over multiple scales and constructs an end-to-end loop structure to achieve rotational invariance. 
In recent work, the locally normalized image feature transform (LNIFT) method \cite{ref58} utilizes a local normalization filter to detect and describe feature points for multimodal image matching. The filter transforms original images into normalized images to reduce the NRD between multimodal images. LNIFT achieves rotational invariance by using an improved ORB detector and a simplified HOG descriptor. However, the detector and descriptor of the LNIFT method similar to the ORB method cannot solve the scale distortion. 
Fan \emph{et al}. \cite{ref59} proposed a novel multisource, multitemporal remote sensing image matching method based on multi-oriented filters, named oriented filter-based matching (OFM) method. In OFM, salient feature points are first extracted based on a phase congruency (PC) model; then the rotation-invariant feature descriptors are constructed based on the index map of multidirectional convolution; and finally the reliable matching sets are determined by the RANSAC algorithm and a match set iterative optimization strategy. Although OFM is not invariant to affine deformation, the OFM method is unaffected by scale and rotational changes; which is robust to significant nonlinear intensity differences (NIDs). 
Zhang \emph{et al} \cite{ref16}. proposed a multimodal remote sensing images (RSIs) matching method named modal-independent consistency matching (MICM). The method utilizes the capability of deep convolutional neural network and the transformer attention mechanism to improve the matching performance, which has obvious effects and advantages in dealing with significant nonlinear radiometric differences and noise in multimodal RSIs. As with most deep learning-based approaches, the method faces the challenge of relatively high computational complexity. 

Distinct from existing work, this paper focuses on the impact of noise interference on image matching. Based on the in-depth analysis of interference mechanism, key problems and robust design strategy, an effective implementation architecture for solving noise image matching is proposed.
	
% The main core of this paper is to explore the robustness and accuracy of multisource image matching methods under noise interference. 
% In this regard, we propose key problems, design strategies, and implementation framework for solving noise image matching based on the analysis of the noise interference mechanism.}

\section{Proposed Method} 
\begin{figure*}[htbp]	
	\hspace{-0.42cm}\includegraphics[width=1\textwidth]{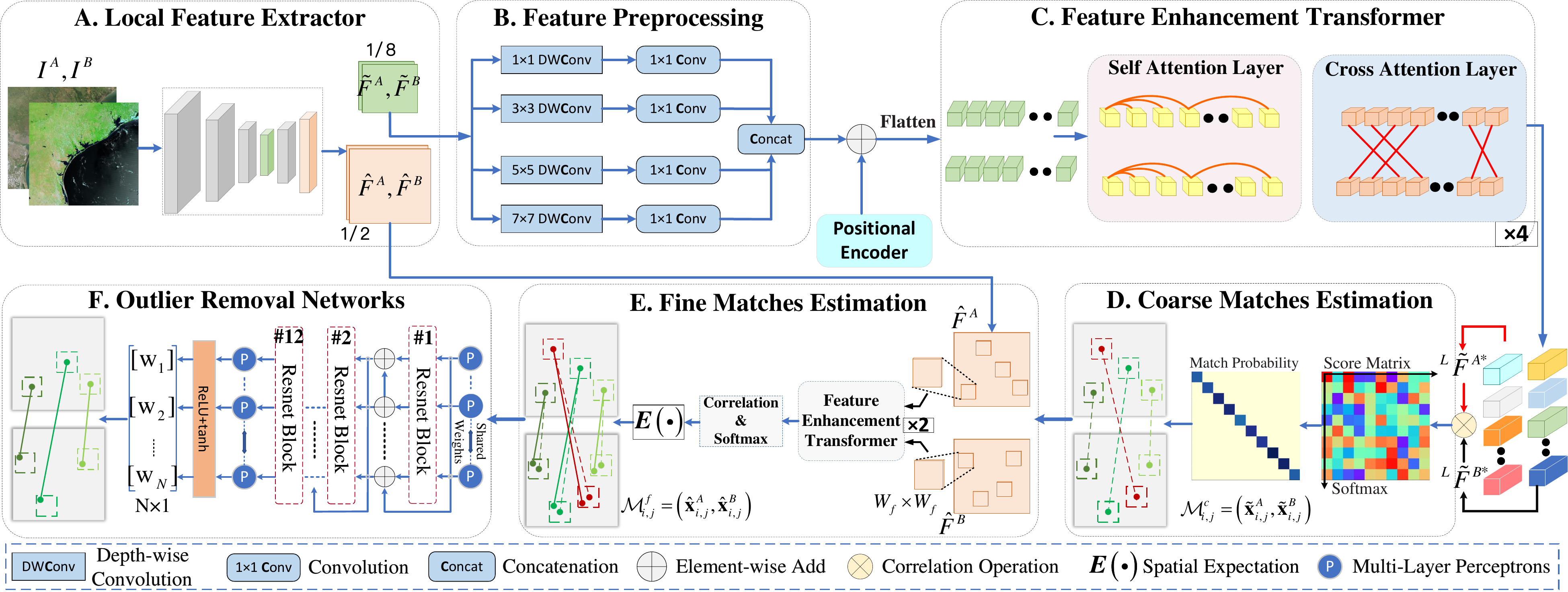}
	\caption{The framework of the proposed robust method for solving noise image matching}
	\label{Fig.1}
\end{figure*}

Remote sensing image matching based on deep learning improve adaptability to noise interference and nonlinear radiation \cite{ref14,ref15}. 
On this basis, we propose a robust matching method against noise interference, as shown in Fig. \ref{Fig.1}. It has six components: 
(i) For the given input images $I^A$ and $I^B$, we use a deep convolutional neural network with feature pyramid type for dense feature extraction to obtain coarse-level features ${{\tilde{F}}^{A}}$ and ${{\tilde{F}}^{B}}$, together with fine-level features ${{\hat{F}}^{A}}$ and ${{\hat{F}}^{B}}$ (Section \ref{Section III-A}). 
(ii) The depthwise separable convolution module serves as feature preprocessing, increasing the receptive field for feature extraction and then performing multi-scale feature aggregation (Section \ref{Section III-B}). 
(iii) We add normalized position encoding to the preprocessed features and then input the features with position information into the feature enhancement transformer for deep feature fusion (Section \ref{Section III-C}). 
(iv) The coarse matching estimation establishes dense matching on the enhanced coarse feature map and selects matches with high confidence from them (Section \ref{Section III-D}). 
(v) The fine matching estimation extracts the feature maps around the coarse matches and refines them to pixel-level coordinates by correlation methods (Section \ref{Section III-E}). 
(vi) The outlier removal network utilizes a multilayer perceptron to extract the depth information in the fine matching estimation and establish effective and geometrically consistent correspondences between images; 
the candidate matches are weighted by classification, and outliers are removed based on the correspondences (Section \ref{Section III-F}). 
In the end, the loss function and implementation details of the proposed method are described in Section \ref{Section III-G}. 

For better understanding, we also exhibit important abbreviations and notations in Table \ref{TABLE 0}. 
We use superscripts and subscripts to indicate the source and index of a variable, respectively. 
Scalars are shown in lowercase boldface, matrices and feature maps are denoted in uppercase boldface. Matching sets and loss functions are represented in fancy text.

\subsection{Local Feature Extraction} \label{Section III-A}

\begin{figure}[htbp]
	\centering
	\includegraphics[width=0.48\textwidth,height=33mm]{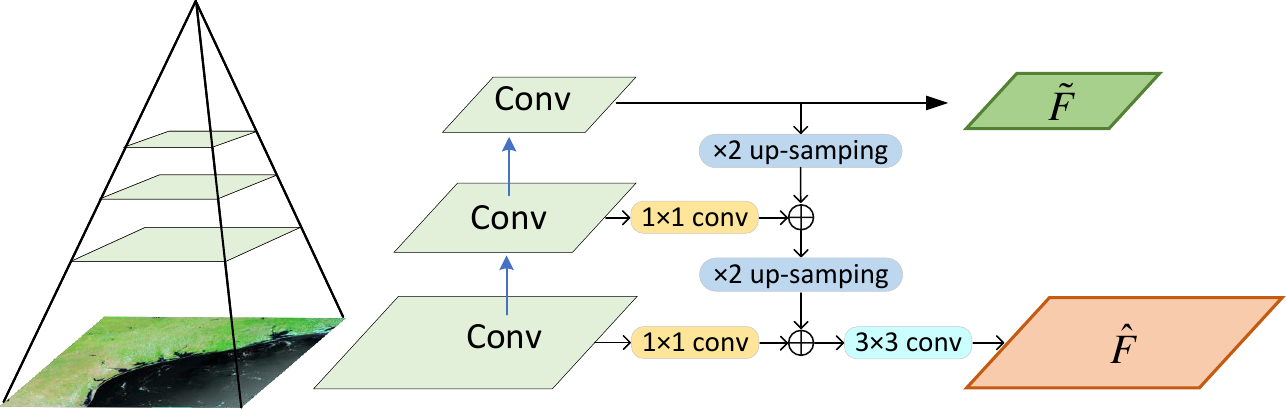}
	\caption{Feature pyramid network. As the convolutional layers increase, the resolution of the feature map decreases, whereas the richer the feature semantic information of the characterization will be.}
	\label{Fig.2}
\end{figure}

% \textbf{\emph{1)} Local Feature Extraction:} 
We use deep convolutional neural networks (CNN) as dense feature extractors. 
Given query images ${{I}^{A}}$ and reference images ${{I}^{B}}$, As shown in Fig. \ref{Fig.2}, 
we employ the standard convolutional architecture of a three-level feature pyramid network (FPN) \cite{ref60,ref61} to extract coarse-level features ${{\tilde{F}}^{A}},{{\tilde{F}}^{B}}\in {{\mathbb{R}}^{H/8\times W/8\times \tilde{C}}}$ 
and fine-level features ${{\hat{F}}^{A}},{{\hat{F}}^{B}}\in{{\mathbb{R}}^{H/2\times W/2\times \hat{C}}}$, where $H$ and $W$ represent the height and width of the image, respectively, 
and $\tilde{C}=256,\hat{C}=128$ represent the dimensions of the coarse-level and fine-level features.

The design of FPN enables the network to capture features at different scales.
Firstly, the coarse feature maps ${{\tilde{F}}^{A}}$ and ${{\tilde{F}}^{B}}$ are obtained through two convolutional layers. 
As the convolutional layers deepen, feature maps will contain richer semantic information. 
Subsequently, through double upsampling and fusion with shallow features from the CNN network, the fused fine-level feature map ${{\hat{F}}^{A}}$ and ${{\hat{F}}^{B}}$ have rich semantic information and high resolution. 
In the subsequent matching process, a quick coarse matching is performed on the smaller-sized coarse feature maps, followed by fine matching on the fine-level feature maps. 
Through feature fusion and hierarchical search, this approach enhances matching speed and produces more accurate results. 

\begin{figure*}[htbp]
	\centering
	\subfigure[]{
	\includegraphics[width=2.5in, height=70mm]{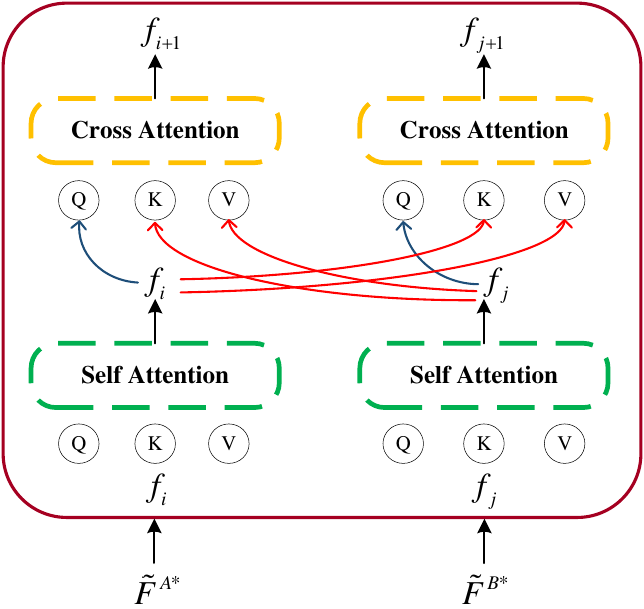}
	\label{Fig.3(a)}
	}
	% \quad
	\subfigure[]{
	\includegraphics[width=2.5in]{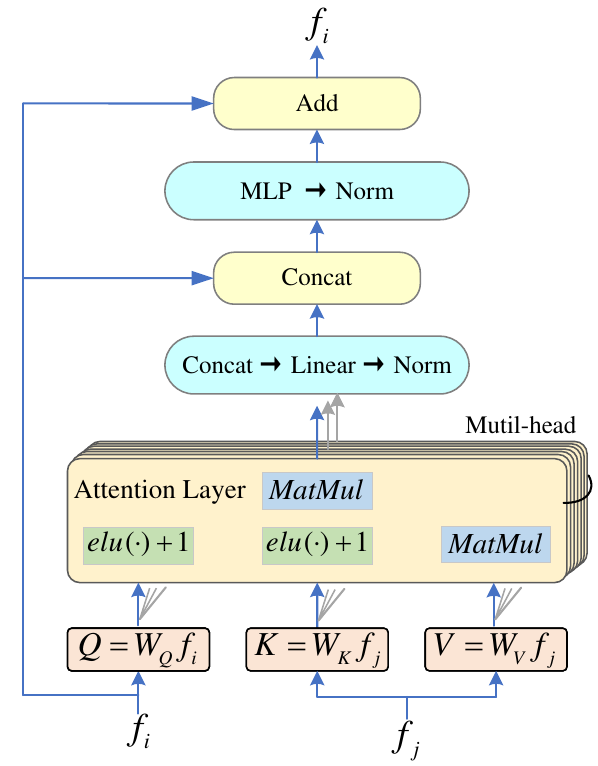}
	\label{Fig.3(b)}
	}
	\caption{Feature enhancement transformer. (a) Alternating combinations of self-attention and cross-attention to form the feature enhancement transformer module. (b) Illustration of the multi-head linear attention layer.}
	\label{Fig.3}       % Give a unique label
\end{figure*}

\subsection{Feature Preprocessing} \label{Section III-B}
A multiscale feature representation is crucial for distinguishing objects or regions of different sizes and preserving the local features of various scales. 
Therefore, we propose a feature preprocessing module (FPM) between the local feature extractor and the feature enhancement transformer to reinforce local feature information and ensure effective deep feature interaction in the feature enhancement transformer. 
The FPM adopts depthwise separable convolution (DSC), initially performing depth convolutions in spatial dimensions $(1\times1, 3\times3, 5\times5, 7\times7)$, 
followed by a $1\times1$ point convolution. Spatial depth convolutions can enhance multiscale feature information for ${{\tilde{F}}^{A}}$ and ${{\tilde{F}}^{B}}$, focusing on local features at different scales.
Point convolution serves as feature fusion between channels, enabling the acquisition of multiscale feature representations. 
DSC helps reduce model parameters and computational complexity \cite{ref62}. 
Finally, the features ${{\tilde{F}}^{A}_{FPM}},{{\tilde{F}}^{B}_{FPM}}\in {{\mathbb{R}}^{H/8\times\ W/8\times\ \tilde{C}}}$ 
obtained by concatenating the convolved features along the channel dimension, which can be formulated as follows:
\begin{equation}
	\label{deqn_ex1}
	\begin{aligned}
		FPM(F)&=\left[ C_{1}^{1/4}\left( {D{{W}_{1}}(F)} \right)|| \ C_{1}^{1/4}\left( {D{{W}_{3}}(F)} \right) \right.\\
		&\left. \ \  || \ C_{1}^{1/4}\left( {D{{W}_{5}}(F)} \right)|| \ C_{1}^{1/4}\left( {D{{W}_{7}}(F)}  \right) \right]
	\end{aligned}
\end{equation}
\begin{equation}
	\label{deqn_ex2}
	\tilde{F}_{FPM}^{A}=FPM \left({{\tilde{F}}^{A}}\right),\ \ \tilde{F}_{FPM}^{B}=FPM \left({{\tilde{F}}^{B}}\right)
\end{equation}
where ${{C}^{1/4}_1}$ represents using $1\times1$ convolution to compress the channel dimension to ${\tilde{C} /4}$,  
$D{{W}_{1}},D{{W}_{3}},D{{W}_{5}},D{{W}_{7}}$ denotes depth-wise convolution with kernel sizes of 1, 3, 5, 7, respectively; 
$[\cdot || \cdot]$ indicates concatenation along the channel dimension.

\subsection{Feature Enhancement Transformer} \label{Section III-C}

\textbf{\emph{1)} Normalized Positional Encoding:} Position encoding (PE) can provide unique position information for each element in the feature map, 
preserving spatial positional information between input images \cite{ref63}. 
We adopt a sine format for position encoding, encoding the position information into specific vectors \cite{ref24}. 
These vectors are fused with preprocessed features $\tilde{F}_{FPM}^{A}$ and $\tilde{F}_{FPM}^{B}$, 
the fused features contain both positional and depth feature information. 
The position encoding is defined as follows:
\begin{equation}
	\label{deqn_ex3}
	P{{E}^{i}}\left( x,\ y \right)=\left\{\begin{matrix}
		\sin \left( {{\omega}_{k}}\cdot x \right),\ &i=4k \quad \ \ \   \\
		\cos \left( {{\omega}_{k}}\cdot x \right),\ &i=4k+1 \\
		\sin \left( {{\omega}_{k}}\cdot y \right),\ &i=4k+2 \\
		\cos \left( {{\omega}_{k}}\cdot y \right),\ &i=4k+3 \\
		\end{matrix}\ \right.
\end{equation}
where $\left( x,\ y \right)$ are the position coordinates of the elements in feature map.
${{\omega}_{k}}=\frac{1}{{10000}^{2k/d}}$, 
$d$ is the feature dimension for position coding, 
and $i$ is the index of feature dimension.

One drawback of this method is that when the testing resolution differs from the training resolution, uncertain coordinates may be used for encoding, which affects the precise localization and boundary perception capabilities of the network. 
To solve this problem, we utilize a simple normalization technique, expressed as: 
\begin{equation}
	\label{deqn_ex4}
	{NPE}^{i} \left( x,\ y \right)={PE}^{i}\left( x\ast\mu , \ y\ast\nu \right)
\end{equation}
\begin{equation}
	\label{deqn_ex5}
	\mu = {{W}_{train}}/{{W}_{test}}, \ \nu ={{H}_{train}}/{{H}_{test}}
\end{equation}
where $NPE(\cdot)$ represents the operation of normalizing position encoding, 
and $W/{{H}_{train/test}}$ denotes the width/height of the training/testing images. 
By adding the normalized position encoding to $\tilde{F}_{FPM}^{A}$ and $\tilde{F}_{FPM}^{B}$, the transformed features will become position-dependent.
and the updated feature set is ${{\tilde{F}}^{A^{\ast}}},{{\tilde{F}}^{B^{\ast}}}\in {{\mathbb{R}}^{N\times \tilde{C}}}; N=H/8\times W/8$, ie.:
\begin{equation}
	\label{deqn_ex6}
	{{\tilde{F}}^{A^{\ast}}}=NPE\left( {\tilde{F}_{FPM}^{A}} \right), \ \ {{{\tilde{F}}}^{B^{\ast}}}=NPE\left( {\tilde{F}_{FPM}^{B}} \right)
\end{equation}
which will serve as the input sequence for the feature enhancement module.

\textbf{\emph{2)} Feature Enhancement Transformer:} As shown in Fig. \ref{Fig.3(a)}, we utilize a combination of multi-level self-attention and cross-attention as the feature enhancement transformer module, referred to as FEFormer. 
FEFormer distinguishes features within images by self-attention, learns the similarity between image pairs by cross-attention, and aggregates the local information and global information within and between images into the generation of feature descriptors, 
which makes feature descriptors more discriminable and robust, improving the success rate of feature matching. 
Additionally, linear attention can reduce the computational complexity of attention operations \cite{ref25}, as shown in Fig. \ref{Fig.3(b)}. 
The formula for linear attention is as follows:
\begin{equation}
	\label{deqn_ex7}
	\begin{split}
		\mathsf{\mathcal{A}}(Q,K,V) &= \phi(Q) \cdot (\phi {{(K)}^{T}}\cdot V) \\
		\phi (\cdot) &= elu(\cdot) + 1   \\
	\end{split}		
\end{equation}
where $\mathsf{\mathcal{A}}$ denotes linear attention, $elu(\cdot)$ is an exponential linear unit. 
Compared to ordinary attention \cite{ref21}, 
the computational complexity is reduced from $\mathsf{\mathcal{O}} \left( {N}^2 \right)$ to $\mathsf{\mathcal{O}} \left( {N} \right)$. 

Specifically, for self-attention, the queries, keys, and values in the attention mechanism are derived from the same input features, 
such as $\left( {\tilde{F}}^{A^{\ast}},{\tilde{F}}^{A^{\ast}} \right)$ or $\left( {\tilde{F}}^{B^{\ast}},{\tilde{F}}^{B^{\ast}} \right)$. 
For cross-attention, to introduce mutual dependencies between feature maps \cite{ref64}, the query comes from the feature ${\tilde{F}}^{B^{\ast}}$ if the keys and values come from ${\tilde{F}}^{A^{\ast}}$,  
or the keys and values come from ${\tilde{F}}^{B^{\ast}}$ and the query comes from the feature ${\tilde{F}}^{A^{\ast}}$. 
Self-attention layers and cross-attention layers are alternately placed in the FEFormer for feature aggregation within and between images, and we perform $L$ times of FEFormer for feature enhancement. 
During the $l$-th feature enhancement, the input features ${\tilde{F}}^{A^{\ast}}$ and ${\tilde{F}}^{B^{\ast}}$ are first fed into the self-attention layers separately for feature aggregation within their respective images, 
the outputs of the self-attention layers are then alternately fed into the cross-attention layers for searching similar positions from the cross images for matching. 
The process is expressed as: 
\begin{equation}
	\label{deqn_ex8}
	\begin{split}
		^{l-1}{{\tilde{F}}^{A^{\ast}}} = \text{Self\_Attention} \left( ^{l-1}{{\tilde{F}}^{A^{\ast}}}, \ ^{l-1} {{\tilde{F}}^{A^{\ast}}}  \right)  \\
		^{l-1}{{\tilde{F}}^{B^{\ast}}} = \text{Self\_Attention} \left( ^{l-1}{{\tilde{F}}^{B^{\ast}}}, \ ^{l-1} {{\tilde{F}}^{B^{\ast}}}  \right)  \\
		^{l}{{\tilde{F}}^{A^{\ast}}} = \text{Cross\_Attention} \left( ^{l-1}{{\tilde{F}}^{A^{\ast}}}, \ ^{l-1} {{\tilde{F}}^{A^{\ast}}}  \right)  \\
		^{l}{{\tilde{F}}^{B^{\ast}}} = \text{Cross\_Attention} \left( ^{l-1}{{\tilde{F}}^{B^{\ast}}}, \ ^{l-1} {{\tilde{F}}^{B^{\ast}}}  \right)  \\	
	\end{split}	
\end{equation}
Ultimately, local information and global context information are integrated into the enhanced feature sequence $^{L}{{\tilde{F}}^{A^{\ast}}},^{L}{{\tilde{F}}^{B^{\ast}}}$.	 	

\subsection{Coarse Matching Estimation} \label{Section III-D}

We construct coarse matching using the enhanced coarse-level feature maps. 
First, the correlation score matrix $ S\left( i,j \right)$ is calculated from $^{L}{{\tilde{F}}^{A^{\ast}}}$ and $^{L}{{\tilde{F}}^{B^{\ast}}}$.
\begin{equation}
	\label{deqn_ex9}
	S\left( i,j \right)\ ={{\ }^{L}}{{\tilde{F}}^{A^{\ast}}}\otimes{{\ }^{L}}{{\tilde{F}}^{B^{\ast}}}
\end{equation}
where $\otimes$ represent correlation operator. 
$i$ and $j$ denotes row indexes and column indexes, respectively.
To directly compute the probability confidence matrix $ P_{i,j}^{c}$ for coarse matching, 
softmax (referred to as 2D-softmax) is used on both the row and column dimensions of $S\left( i,j \right)$. 
The expression is given by Equation (10):  
\begin{equation}
	\label{deqn_ex10}
	P_{i,j}^{c} = softmax \left( S\left( {i},{:} \right) \right)\odot  softmax \left( S\left( {:},{j}  \right) \right)
\end{equation}
where $\odot$ represent the pixel-wise product. 
Next, we select matching points in $ {P}_{i,j}^{c}$ 
that are greater than the threshold $\tau_c=0.2$ and fulfill the mutual nearest neighbor (MNN) criterion (i.e., the maximum values in rows and columns of ${P}_{i,j}^{c}$). 
Finally, we obtain coarse matching sets $\mathsf{\mathcal{M}}^{c}$, as shown:  
\begin{equation}
	\label{deqn_ex11}
	\mathsf{\mathcal{M}}^{c}=\left\{ \left( \bm{\mathbf{\tilde{x}}_{i,j}^{A}},\bm{\mathbf{\tilde{x}}_{i,j}^{B}} \right)|\left( P_{i,j}^{c}\ge {{\tau}_{c}} \right)\wedge \emph{\text{MNN}}(P_{i,j}^{c}) \right\}
\end{equation}	
where $\bm{\mathbf{\tilde{\bm{x}}}_{i,j}^{A}}=\left(\tilde{x}_{i}^{A},\tilde{x}_{j}^{A} \right)\in {{\mathbb{R}}^{2}}$ and $\bm{\mathbf{\tilde{\bm{x}}}_{i,j}^{B}}=\left(\tilde{x}_{i}^{B},\tilde{x}_{j}^{B} \right)\in {{\mathbb{R}}^{2}}$ 
denote the pixel coordinates of the matching points on their own image planes, respectively.

\begin{figure}
	\centering
	\includegraphics[width=0.48\textwidth]{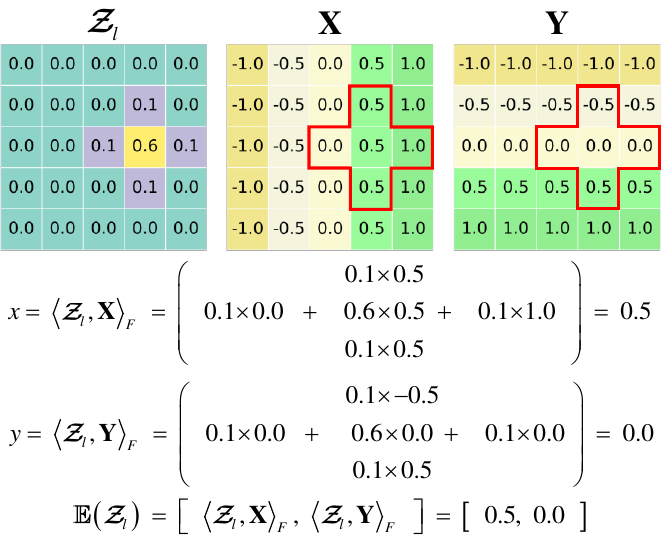}
	\caption{Fine-level coordinate regression using heatmap regression \cite{ref65}, where ${\left \langle \ {\cdot,\cdot} \ \right \rangle}_F$ denotes the Frobenius inner product; 
	\textbf{X} and \textbf{Y} are normalized grid matrix with $n \times m$, and ${{X}_{i,j}}=\frac{2j-n}{n}, {{Y}_{i,j}}=\frac{2i-m}{m}$, $i=1,2,\ldots ,n$, $j=1,2,\ldots ,m$, $n=m=5$ .}
	\label{Fig.4}
\end{figure}
\subsection{Fine Matching Estimation}  \label{Section III-E}

Coarse matching estimation is performed on $1/8$ feature map, resulting in low matching accuracy. 
Therefore, we designed a fine-matching estimation module to achieve precise pixel-level matching. 
For each pair of coarse matching $\left( \bm{\mathbf{\tilde{x}}_{i,j}^{A}}, \bm{\mathbf{\tilde{x}}_{i,j}^{B}}\right)$, 
firstly, we locate their positions $\left( \bm{\mathbf{\hat{x}}_{i,j}^{A}}, \bm{\mathbf{\hat{x}}_{i,j}^{B}}\right)$ on the fine-level feature maps ${{\hat{F}}^{A}}$ and ${{\hat{F}}^{B}}$. 
Secondly, with $\bm{\mathbf{\hat{x}}_{i,j}^{A}}$ and $\bm{\mathbf{\hat{x}}_{i,j}^{B}}$ as the centers, 
two sets of feature patches $\hat{P}_{i,j}^{A},\hat{P}_{i,j}^{B}\in {{\mathbb{R}}^{{{w}_{f}}\times {{w}_{f}}\times \hat{C}}}$ of size $w_{f}\times w_{f}$ are extracted as local dense feature maps. 
Thirdly, we input $\hat{P}_{i,j}^{A}$ and $\hat{P}_{i,j}^{B}$ into the fine-level feature enhancement transformer and perform $L_2$ times alternating operations to obtain the enhanced feature maps, 
denoted by $\hat{P}_{i,j}^{{A}^{*}}$ and $\hat{P}_{i,j}^{{B}^{*}}$. 
The probabilistic confidence matrix ${P}_{i,j}^{f}$ for fine matching is obtained by correlating the center vector ${{\vec{C}}^{\hat{P}_{i,j}^{A^{*}}}}$ of $\hat{P}_{i,j}^{A}$ with all the spatial vectors ${{\vec{S}}^{\hat{P}_{i,j}^{B^{*}}}}$ in $\hat{P}_{i,j}^{B}$:
\begin{equation}
	\label{deqn_ex12}
	{P}_{i,j}^{f}=softmax \left( {\vec{C}}^{\hat{P}_{i,j}^{A^{\ast}}} \otimes {{\vec{S}}^{\hat{P}_{i,j}^{B^{\ast}}}}  \right)
\end{equation}
where $\otimes$ denotes the correlation operation. 
Particularly, ${P}_{i,j}^{f}$ is equivalent to a local heatmap ${\mathsf{\mathcal{Z}}}_{l}\in {{\mathbb{R}}^{{{W}_{f}}\times {{W}_{f}}}}$. 
By calculating the probability distribution expectation, i.e., the spatial expectation on the local heatmap $ \mathbb{E} \left( {{\mathsf{\mathcal{Z}}}_{l}} \right)$ \cite{ref65}, 
the coordinate for fine matching between $\hat{P}_{i,j}^{A}$ and $\hat{P}_{i,j}^{B}$ can be determined, as shown in Fig. \ref{Fig.4}.  %\mathit{\mathbb{E}}
We iteratively process the full set of coarse-to-fine matches in the same way, and calculate the pixel-level coordinates of the matching points in the original resolution image. Finally, we obtain pixel-level matching sets $\mathsf{\mathcal{M}}$ between the image pairs $\left( {{I}^{A}},{{I}^{B}} \right)$.

\subsection{Outlier Removal Network} \label{Section III-F}

\begin{figure*}[htbp]
	\centering
	\includegraphics[width=1\textwidth]{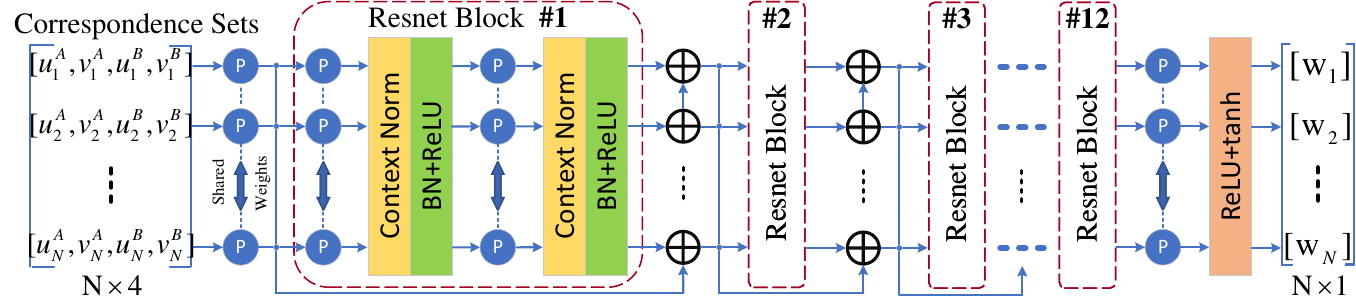}
	\caption{The architecture of the outlier removal network.}
	\label{Fig.5}
\end{figure*}

After fine matching estimation, assuming that there are $ N $ pairs of candidate matches in the set $\mathsf{\mathcal{M}}$.
For convenience, which can be formulated as:
\begin{equation}
	\label{deqn_ex13}
	\begin{split}
		\mathsf{\mathcal{M}} = [\bm{{{\mathbf{x}}_{1}}, \ldots, {{\mathbf{x}}_{N}}}]  \\
		\bm{{\mathbf{x}}_{i}} = [u_{i}^{A},v_{i}^{A},u_{i}^{B},v_{i}^{B}]  \\
	\end{split}		
\end{equation}	
where $\bm{{\mathbf{x}}_{i}}$ is a quaternion that represents the matching relationship between feature points $\left( u_{i}^{A},v_{i}^{A} \right)$ and $\left( u_{i}^{B},v_{i}^{B} \right)$, 
and $\mathsf{\mathcal{M}}$ is the set of all feature matching.

We formulate the outlier removal task as a binary classification problem. 
As described in Fig. \ref{Fig.5}, the outlier removal network leverages a multi-layer perceptron to extract feature point geometry information from fine-matching estimates. 
It utilizes context normalization to acquire both global and local information, establishing effective and geometrically consistent correspondences between images. 
Based on these correspondences, each pair of candidate matches is weighted through classification. 
If the matching confidence is high, the weight is larger; otherwise, the weight is more minor. 
The inputs to the network are the coordinates of a list of the candidate matches. 
Specifically, the candidate matches set $\mathsf{\mathcal{M}}$ is fed into multilayer perceptrons (denoted as \textcircled{P}), 
we use global and local contexts to calculate weights $\mathsf{\mathcal{W}}=[{{\text{w}}_{1}},\ldots ,{{\text{w}}_{N}}],\ {{\text{w}}_{i}}\in \left[ 0,1 \right]$, 
For each pair of candidate matches, as the weights get closer to 1, the probability of correct match is high; conversely, it indicates a wrong matching.  
Therefore, the architecture of the outlier removal network can be formulated as follows:
\begin{equation}
	\label{deqn_ex14}
	\mathsf{\mathcal{Y}}={{\mathsf{\mathcal{F}}}_{\phi}} (\mathsf{\mathcal{M}})
\end{equation}	
\begin{equation}
	\label{deqn_ex15}
	\mathsf{\mathcal{W}}=[{{w}_{1}}, \ldots, {{w}_{N}}]=tanh\left( ReLU\left( \mathsf{\mathcal{Y}} \right) \right)
\end{equation}
where $\mathsf{\mathcal{Y}}$ is the logical value for classification. 
${{\mathsf{\mathcal{F}}}_{\phi}} (\cdot)$ is an end-to-end network composed of multiple residual modules, 
and it is permutation-equivariant, meaning that the output of the network will not be affected by the unordered inputs. 
$\phi$ is the network parameters. $tanh$ and $ReLU$ are activation functions for removing outliers in classification \cite{ref66}. 
$\mathsf{\mathcal{W}}$ represents the weights assigned to candidate matches. 
${\mathbf{x}}_{i}$ is an outlier when ${{w}_{i}}=0$, and ${\mathbf{x}}_{i}$ is an inlier when ${{w}_{i}}=1$.

\subsection{Loss Function and Implementation Details}  \label{Section III-G}

\begin{figure*}[hbp]
	\centering
	\hspace{-0.1cm}\subfigure[]{  % Optical-Depth
	\begin{minipage}[b]{0.24\textwidth}
	\includegraphics[width=1\linewidth,height=20mm]{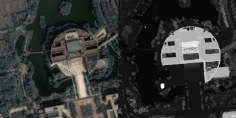}\vspace{4pt}
	\includegraphics[width=1\linewidth,height=20mm]{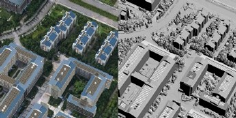}\vspace{4pt}
	\end{minipage}}
	\hspace{-0.1cm}\subfigure[]{  %Infrared-Optical
	\begin{minipage}[b]{0.24\textwidth}
	\includegraphics[width=1\linewidth,height=20mm]{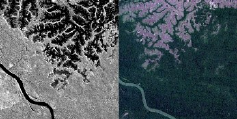}\vspace{4pt}
	\includegraphics[width=1\linewidth,height=20mm]{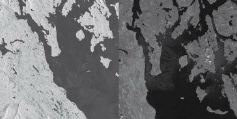}\vspace{4pt}
	\end{minipage}}
	\hspace{-0.1cm}\subfigure[]{  % Optical-Map
	\begin{minipage}[b]{0.24\textwidth}
	\includegraphics[width=1\linewidth,height=20mm]{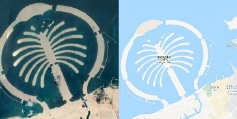}\vspace{4pt}
	\includegraphics[width=1\linewidth,height=20mm]{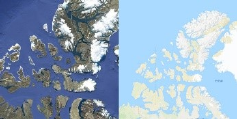}\vspace{4pt}
	\end{minipage}}
	\hspace{-0.1cm}\subfigure[]{  % SAR-Optical
	\begin{minipage}[b]{0.24\textwidth}
	\includegraphics[width=1\linewidth,height=20mm]{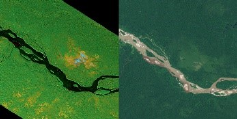}\vspace{4pt}
	\includegraphics[width=1\linewidth,height=20mm]{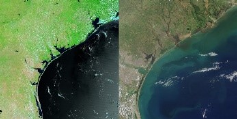}\vspace{4pt}
	\end{minipage}}
	\caption{Sample of the testing dataset. (a) Optical-Depth. (b) Infrared-Optical. (c) Optical-Map. (d) SAR-Optical.}
	\label{Fig.6}
\end{figure*}
\begin{figure*}[hbp]
	% \centering
	\begin{minipage}{1\linewidth}
		\centering
		\includegraphics[width=0.245\linewidth]{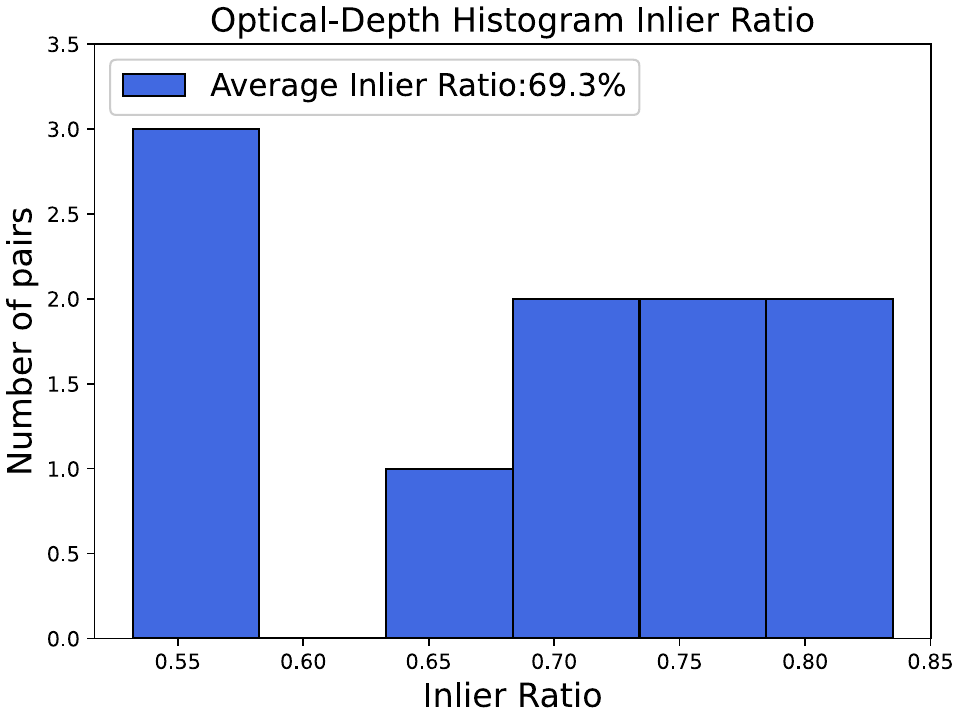}
		\label{fig7a}
		\includegraphics[width=0.245\linewidth]{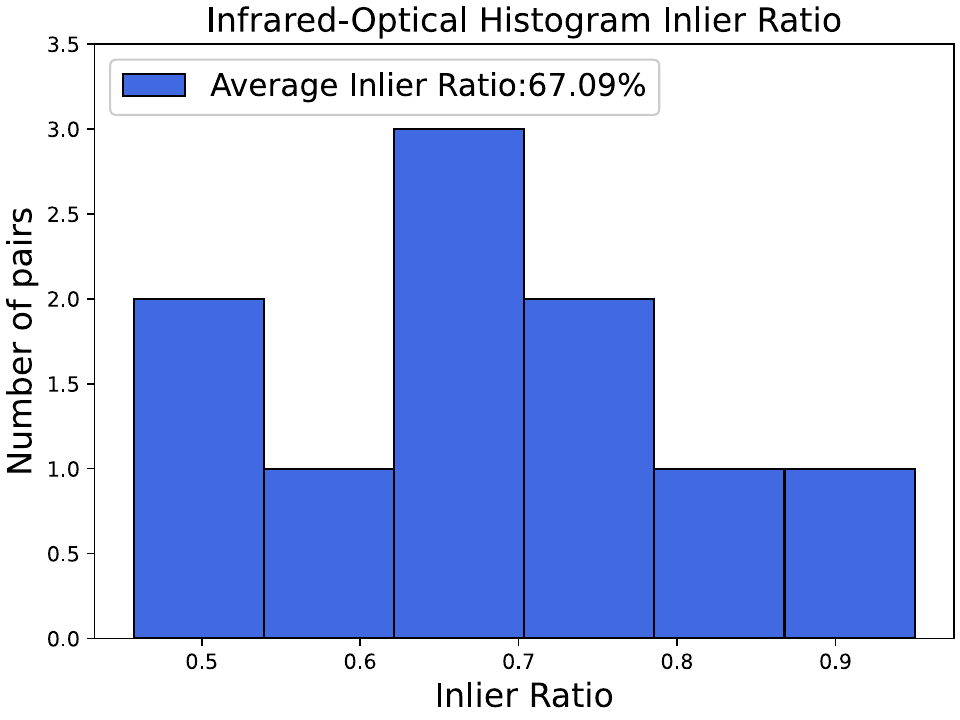}
		\label{fig7b}
		\includegraphics[width=0.245\linewidth]{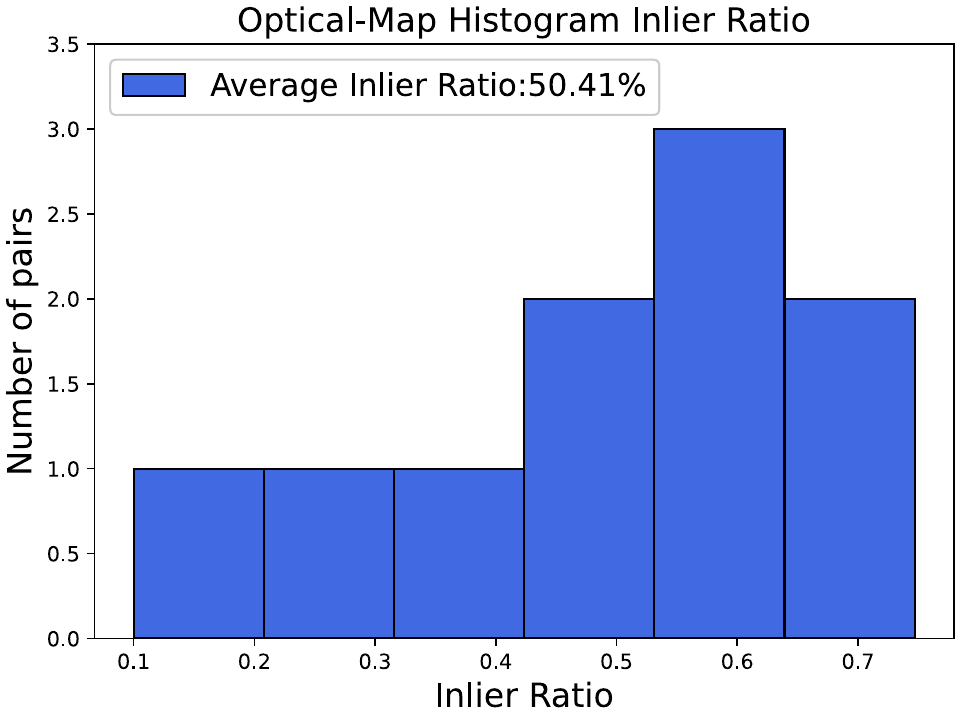}
		\label{fig7c}
		\includegraphics[width=0.245\linewidth]{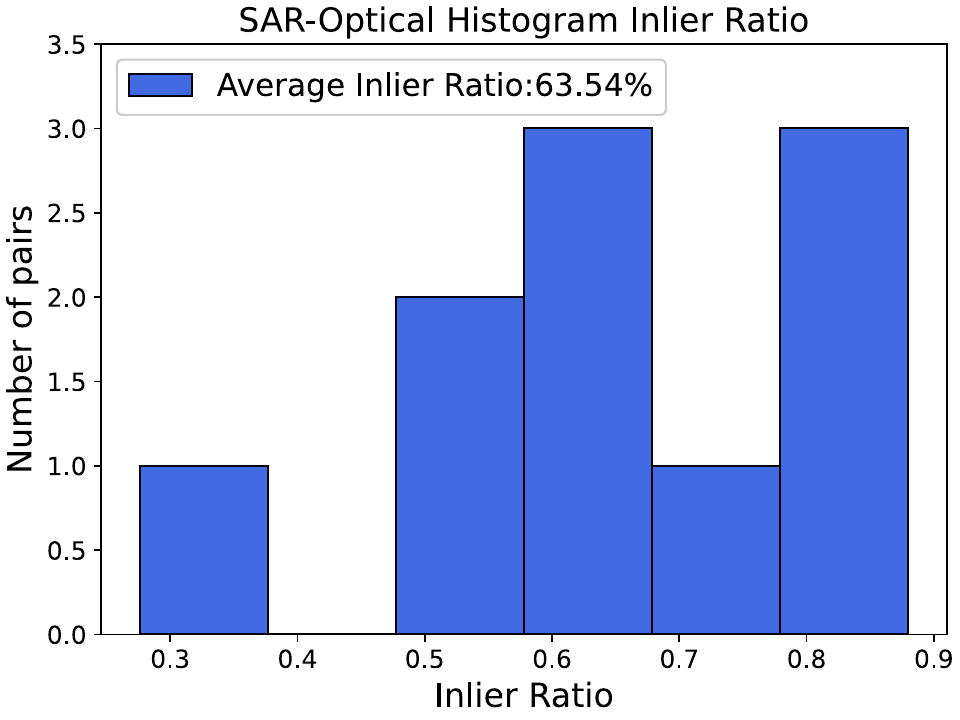}
		\label{fig7d}
		% \caption{Histogram}
	\end{minipage}
	\caption{The statistical information for the testing dataset. The histogram illustrates the distribution of inlier ratios for all image pairs in each dataset. It can be observed that the inlier ratios for some image pairs are relatively low, indicating that only a few assumed correspondences are correct. Therefore, it is challenging to find these correct ones.}
	\label{Fig.7}
\end{figure*}

\textbf{\emph{1)} Loss Function:} The final loss consists of three parts: 
coarse matching loss ${{\mathsf{\mathcal{L}}}_{\text{coarse}}}$, 
fine matching loss ${{\mathsf{\mathcal{L}}}_{\text{fine}}}$, 
and outlier removal network loss ${{\mathsf{\mathcal{L}}}_{\phi}}$; 
it can be expressed as:
\begin{equation}
	\label{deqn_ex16}
	\mathsf{\mathcal{L}}_{all}={{\mathsf{\mathcal{L}}}_{\text{coarse}}}+{{\mathsf{\mathcal{L}}}_{\text{fine}}}+{{\mathsf{\mathcal{L}}}_{\phi}}
\end{equation}

Coarse Matching Loss: The coarse matching sets ${\mathsf{\mathcal{M}}}^{c}$ is defined as the mutual nearest neighbors of two sets of 1/8 resolution grids \cite{ref24}, 
and the matching error is measured through the reprojected distance at their center positions. 
Based on the depth maps and camera poses provided by the datasets, the ground truth coarse matching sets ${\mathsf{\mathcal{M}}}_{gt}^{c}$ is determined through the reprojection of the depth maps and camera poses onto the image pairs. 
The ground truth confidence matrix ${P}_{gt}^{c}$ can be inversely inferred from ${\mathsf{\mathcal{M}}}_{gt}^{c}$. 
The cross-entropy loss between the inferred ${P}_{i,j}^{c}$ and its ground truth label ${P}_{gt}^{c}$ is utilized as the coarse matching loss function, it can be expressed as: 
\begin{equation}
	\label{deqn_ex17}  
	\begin{split}
		{{\mathsf{\mathcal{L}}}_{\text{coarse}}}=\frac{1}{N_{c}^{2}} \sum\limits_{i=1,j=1}^{{{N}_{c}},{{N}_{c}}} 
	\Big(&-{P}_{gt}^{c}\log \left({P}_{i,j}^{c}\right) \\
		&-( 1-{P}_{gt}^{c})\log{\left(1-{P}_{i,j}^{c}\right)} \Big)
	\end{split}	
\end{equation}
where $ N_{c}=H/8 \times W/8 $.

Fine Matching Loss: The fine matching is centered on the established coarse matching and searches for feature points within the window $ w_{f} \times w_{f} $. 
The supervision for fine matching is the reprojection error ${\widehat{\xi}}_{i,j}$ between the feature point $\mathbf{\hat{x}}_{i,j}^{A}$ of the query image and the feature point $\mathbf{\hat{x}}_{i,j}^{B}$ of the reference image. 
It can be formulated as follows:
\begin{equation}
	\label{deqn_ex18}
	{\widehat{\xi}}_{i,j}=\mathbb{T}\left(\mathbf{\hat{x}}_{i,j}^{A}\right)-\mathbf{\hat{x}}_{i,j}^{B}
\end{equation}	
where $\mathbb{T}\left( \cdot \right)$ is projective transformation from $\mathbf{\hat{x}}_{i,j}^{A}$ to $\mathbf{\hat{x}}_{i,j}^{B}$, 
depending on the camera pose provided by the datasets, according to \cite{ref66}. 
For each query point, its uncertainty can be measured by calculating the total variance ${{\sigma}^{2}}\left( {{\mathsf{\mathcal{Z}}}_{l}} \right)$ of the corresponding heatmap. 
Our goal is to optimize fine positions with low uncertainty to generate the final weighted loss function. 
\begin{equation}
	\label{deqn_ex19}
	{{\mathsf{\mathcal{L}}}_{\text{fine}}}=\frac{1}{{{N}_{f}}}\sum\limits_{i,j=1}^{{{N}_{f}}}{\frac{1}{{{\sigma}^{2}}\left( {{\mathsf{\mathcal{Z}}}_{l}} \right)}{{\left\| \mathbb{E}({{\mathsf{\mathcal{Z}}}_{l}})-{{\widehat{\xi}}_{i,j}} \right\|}_{2}}}
\end{equation}
where ${{N}_{f}}$ is the number of fine matching. 
When calculating ${{\mathsf{\mathcal{L}}}_{\text{fine}}}$, 
if the position of $\mathbf{\hat{x}}_{i,j}^{A}$ after the projection transformation falls outside the local window of $\hat{P}_{i,j}^{B}$, the corresponding matching will be excluded. 

Outlier removal network loss: The outlier removal networks employ a mixed loss function to supervise the network: 
\begin{equation}
	\label{deqn_ex20}
	{\mathsf{\mathcal{L}}}_{\phi}=\alpha {{\mathsf{\mathcal{L}}}_{cls}}(\mathsf{\mathcal{W}},{{L}_{gt}})+\beta {{\mathsf{\mathcal{L}}}_{ess}}(\hat{E},E_{gt})
\end{equation}
where ${\mathsf{\mathcal{L}}}_{\phi}$ is also the hybrid loss function, 
${\mathsf{\mathcal{L}}}_{cls}$ denotes the classification loss function that computes the binary cross-entropy of the weight sets $\mathsf{\mathcal{W}}$ and the ground truth label sets ${L}_{gt}$, 
${\mathsf{\mathcal{L}}}_{ess}$ represents the regression loss of the essential matrix constrained by geometric distance, 
$\alpha$ is the weight for the classification loss to balance the inlier/outlier ratio, and $\beta$ is a hyperparameter for balancing the classification loss and essential matrix regression loss. 
The formulations for the classification loss function and essential matrix regression loss function are expressed as follows: 
\begin{equation}
	\label{deqn_ex21}
	{{\mathsf{\mathcal{L}}}_{cls}}=\frac{1}{N}\sum\limits_{i=1}^{N}{\mathsf{\mathcal{H}}({{\text{w}}_{i}},{{L}_{gt}})}
\end{equation}	
where ${\mathsf{\mathcal{H}}}(\cdot,\cdot)$ denotes the binary cross entropy, 
which is used to compute the weight sets and label sets. 
The ground truth label values are derived through epipolar geometry constraints. 
In simple terms, a feature point in a given image is considered an incorrect match if its matching point is not on the corresponding epipolar line in another image.
\begin{equation}
	\label{deqn_ex22}
	{{\mathsf{\mathcal{L}}}_{ess}}(\hat{E},E_{gt})=\frac{{{({p^{\prime}}{^{T}}\hat{E}_{gt}p)}^{2}}}{(E_{gt}p)_{[1]}^{2}+(E_{gt}p)_{[2]}^{2}+({{E}^{T}_{gt}}p^{\prime} )_{[1]}^{2}+({{E}^{T}_{gt}}p^{\prime} )_{[2]}^{2}}
\end{equation}	
where $p=\left( u_{i}^{A},v_{i}^{A}, 1 \right)$ and ${p}^{\prime}=\left( u_{i}^{B},v_{i}^{B},1 \right)$ 
represent the putative correspondences between images $I^{A}$ and $I^{B}$ in homogeneous coordinates. 
$(\cdot)_{[i]}$ denotes the i-th element of vector. 
$\hat{E}=g\left( \mathsf{\mathcal{W}},\mathsf{\mathcal{M}} \right)$ is the regressed essential matrix, 
$g\left( \cdot,\cdot \right)$ represents the weighted 8-point algorithm; 
which selects more than eight corresponding points and their weights, 
and computes the essential matrix $\hat{E}$ by self-adjoint eigenvalue decomposition \cite{ref49}.

\textbf{\emph{2)} Implementation details:} We set the coarse matching confidence threshold ${{\tau}_{c}}=0.2$, 
the size of local patch for fine matching is ${{w}_{f}}=5$, and the feature enhancement operations for coarse-levels and fine-levels are ${{L}_{1}}=4$ and ${{L}_{2}}=2$, respectively. 
The hyperparameter in the outlier removal network are set to $\alpha=0.5$ and $\beta=0.1$ in order to balance classification loss and regression loss.

\section{Experiments and Discussion}

In this section, we select several multisource datasets for qualitative and quantitative evaluation. 
We compare our method with four state-of-the-art algorithms under various situations, 
including noise-free, additive random noise, and periodic stripe noise scenarios. 
At the end, we perform ablation study and complexity analyses. 

\subsection{Training and Testing Datasets}

\textbf{\emph{1)} Training Dataset:} Following the same approach as \cite{ref16}, \cite{ref24}, we initially adopt the MegaDepth dataset \cite{ref67} to train our network model. 
The MegaDepth dataset provides training data for a large number of different viewing angles and lighting conditions. 
The dataset uses COLMAP software to build 196 different scenes from 1,070,468 Internet photos. 
We randomly sample 100 pairs from each subscene during each epoch of training, the sampling scheme also balance the sample difficulties and scene varieties of each iteration, which facilitate convergence of networks. 
After the first step of training described above, the second step is to train on a dedicated remote sensing dataset(i.e., SUIRD dataset \cite{ref68}) and refine the existing network parameters to enhance the robustness of the model performance. 
Small UAV Image Registration Dataset (SUIRD): SUIRD is a publicly available dataset for image registration and feature-matching research. 
The SUIRD v2.2 includes 60 pairs of UAV images and provides their ground truth. 
These images are divided into five categories according to various perspectives: horizontal rotation, vertical rotation, mixture, scaling, and extreme.

Our network is trained in two stages. 
Firstly, on MegaDepth dataset, we set the learning rate to $1\times 10^{-3}$, the batch size to 8, and 30 epochs training is conducted. 
Secondly, the fine-tuning stage lasts for 10k iterations with a learning rate of $1\times 10^{-5}$ and a batch size to 1. 
The proposed network is implemented in PyTorch, and the PyTorch-Lighting framework is employed for distributed training. 
The experimental platform comprises an Intel(R) Xeon(R) Gold 5128 CPU @ 2.30 GHz, Nvidia GeForce RTX 3090 GPU, and 16GB RAM.

\textbf{\emph{2)} Testing Dataset:} To evaluate the performance of the proposed method in this paper, we collected four types of multisource remote sensing image datasets from publicly available literature \cite{ref58,ref69}. 
These include optical-depth, infrared-optical, optical-electronic map, and synthetic aperture radar (SAR)-optical image pairs. 
Each type of dataset comprises ten image pairs, each with a size of $512 \times 512$, as illustrated in Fig. \ref{Fig.6}.
The depth images are derived from airborne LiDAR data, infrared images are sourced from Landsat TM-5 or airborne infrared sensors, 
electronic maps are obtained from Google Maps, and SAR data is acquired from the GF-3 satellite. 
The average inlier ratios are 69.30\%, 67.09\%, 50.41\%, and 63.54\% for each dataset type, as depicted in the histogram in Fig. \ref{Fig.7}. 
These image pairs exhibit significant non-linear radiometric distortions, contrast differences, and geometric transformations such as scale variations, rotations, and displacements. 
These characteristics pose considerable challenges to image matching methods. 
Instead, these challenges can test the effectiveness and robustness of the proposed matching methods in a more comprehensive ways.

\begin{figure*}[hbp]
	\centering
	\hspace{-0.1cm}\subfigure[]{  % Optical-Depth
	\begin{minipage}[b]{0.24\textwidth}
	\includegraphics[width=1\linewidth,height=17mm]{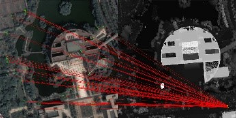}\vspace{1pt}
	\includegraphics[width=1\linewidth,height=17mm]{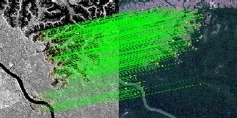}\vspace{1pt}
	\includegraphics[width=1\linewidth,height=17mm]{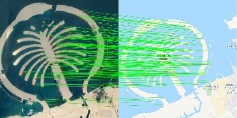}\vspace{1pt}
	\includegraphics[width=1\linewidth,height=17mm]{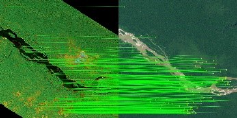}\vspace{1pt}
	\end{minipage}}
	\hspace{-0.1cm}\subfigure[]{  %Infrared-Optical
	\begin{minipage}[b]{0.24\textwidth}
	\includegraphics[width=1\linewidth,height=17mm]{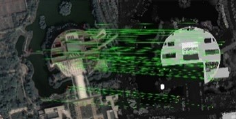}\vspace{1pt}
	\includegraphics[width=1\linewidth,height=17mm]{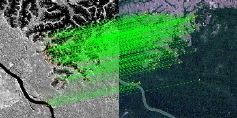}\vspace{1pt}
	\includegraphics[width=1\linewidth,height=17mm]{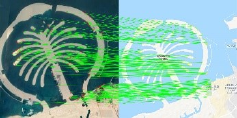}\vspace{1pt}
	\includegraphics[width=1\linewidth,height=17mm]{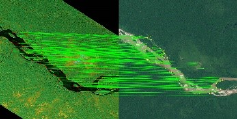}\vspace{1pt}
	\end{minipage}}
	\hspace{-0.1cm}\subfigure[]{  % Optical-Map
	\begin{minipage}[b]{0.24\textwidth}
	\includegraphics[width=1\linewidth,height=17mm]{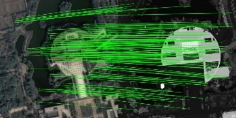}\vspace{1pt}
	\includegraphics[width=1\linewidth,height=17mm]{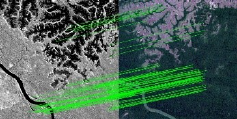}\vspace{1pt}
	\includegraphics[width=1\linewidth,height=17mm]{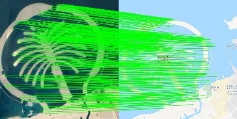}\vspace{1pt}
	\includegraphics[width=1\linewidth,height=17mm]{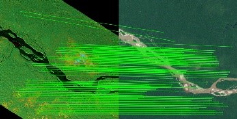}\vspace{1pt}
	\end{minipage}}
	\hspace{-0.1cm}\subfigure[]{  % SAR-Optical
	\begin{minipage}[b]{0.24\textwidth}
	\includegraphics[width=1\linewidth,height=17mm]{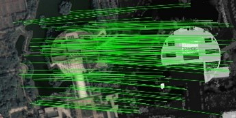}\vspace{1pt}
	\includegraphics[width=1\linewidth,height=17mm]{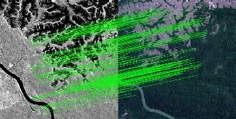}\vspace{1pt}
	\includegraphics[width=1\linewidth,height=17mm]{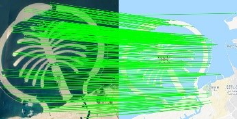}\vspace{1pt}
	\includegraphics[width=1\linewidth,height=17mm]{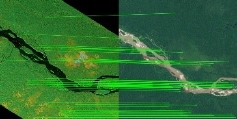}\vspace{1pt}
	\end{minipage}}
	\\
	\hspace{-0.1cm}\subfigure[]{  % SAR-Optical
	\begin{minipage}[b]{0.24\textwidth}
	\includegraphics[width=1\linewidth,height=17mm]{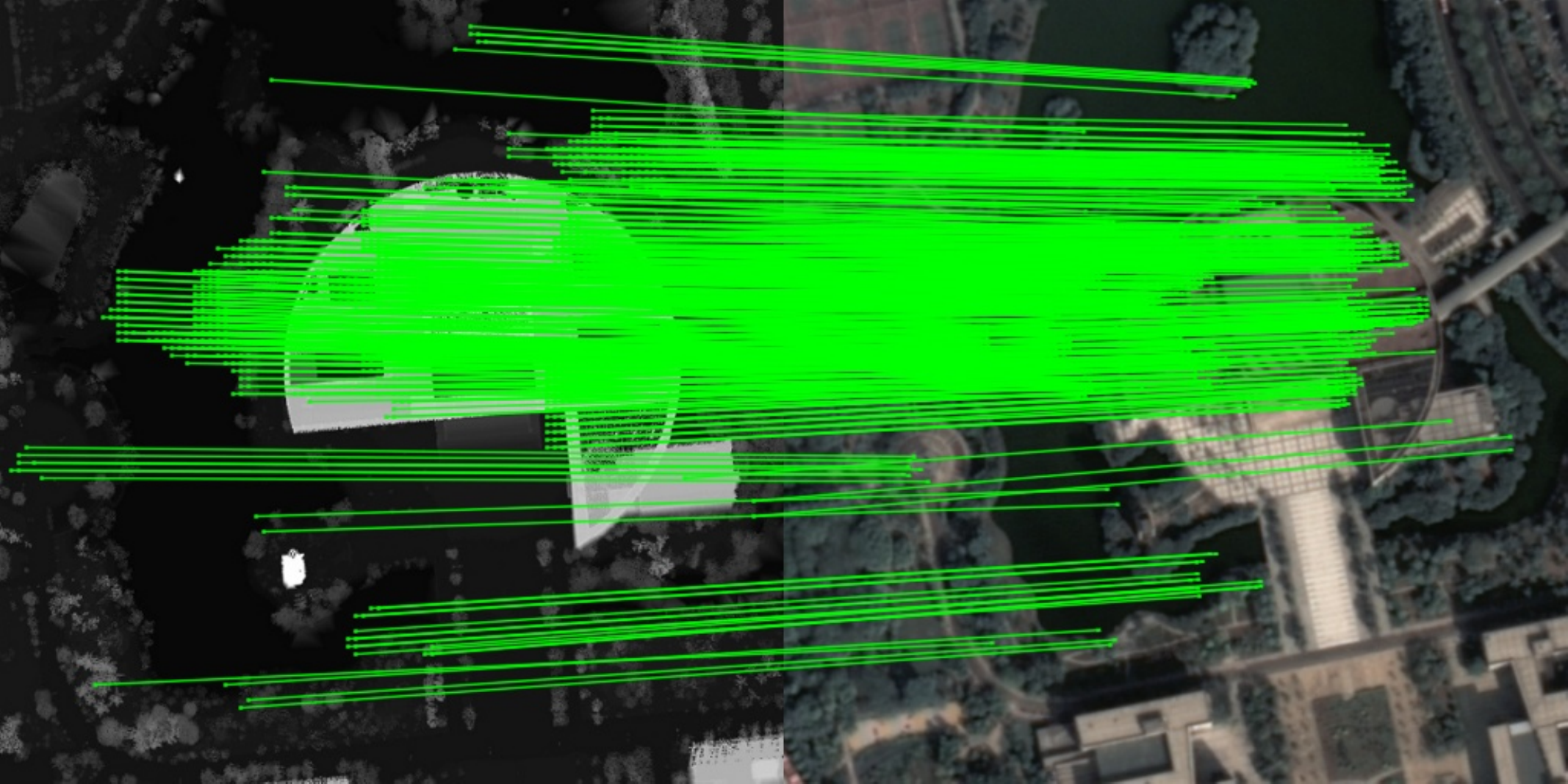}\vspace{1pt}
	\includegraphics[width=1\linewidth,height=17mm]{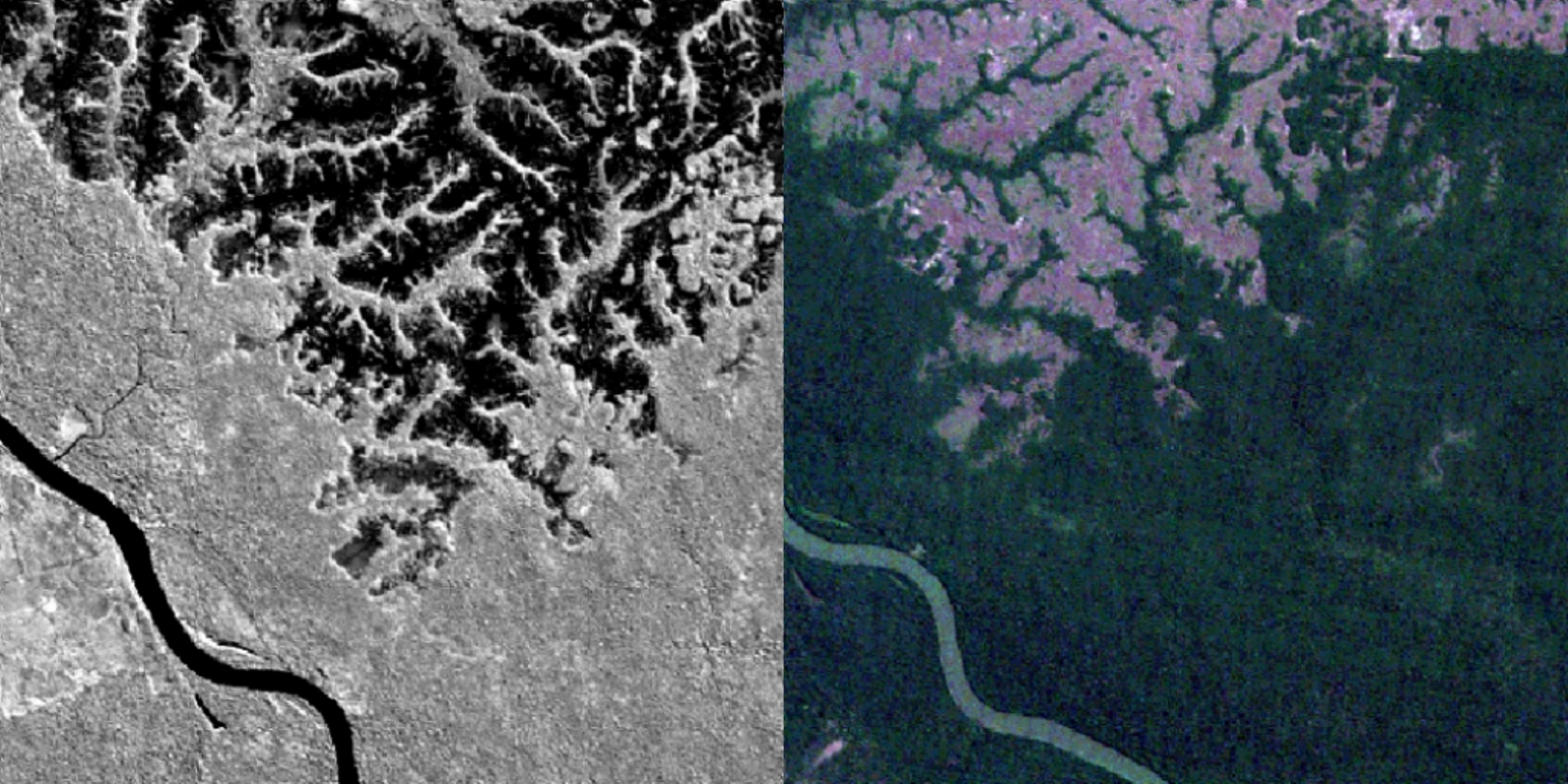}\vspace{1pt}
	\includegraphics[width=1\linewidth,height=17mm]{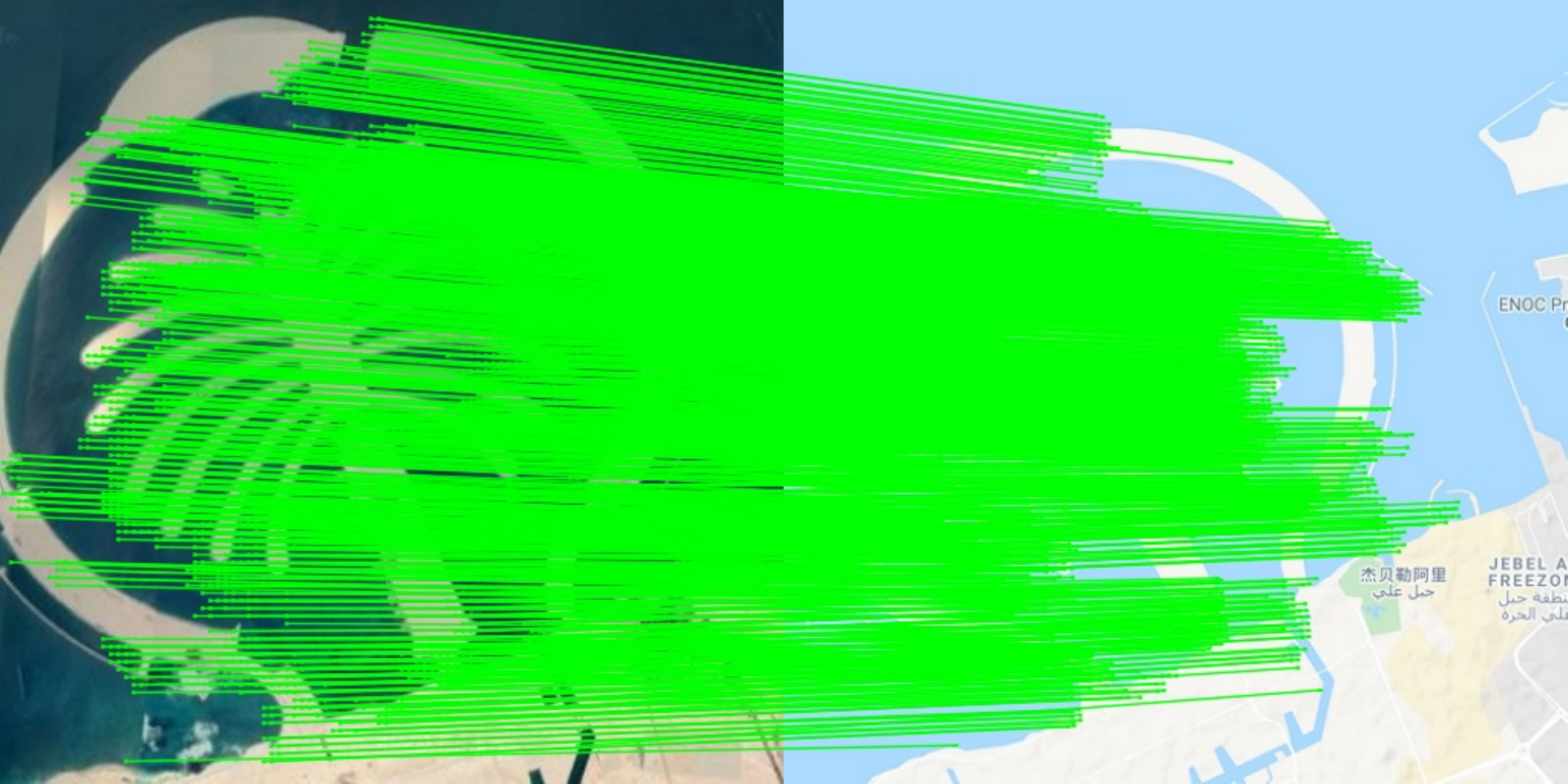}\vspace{1pt}
	\includegraphics[width=1\linewidth,height=17mm]{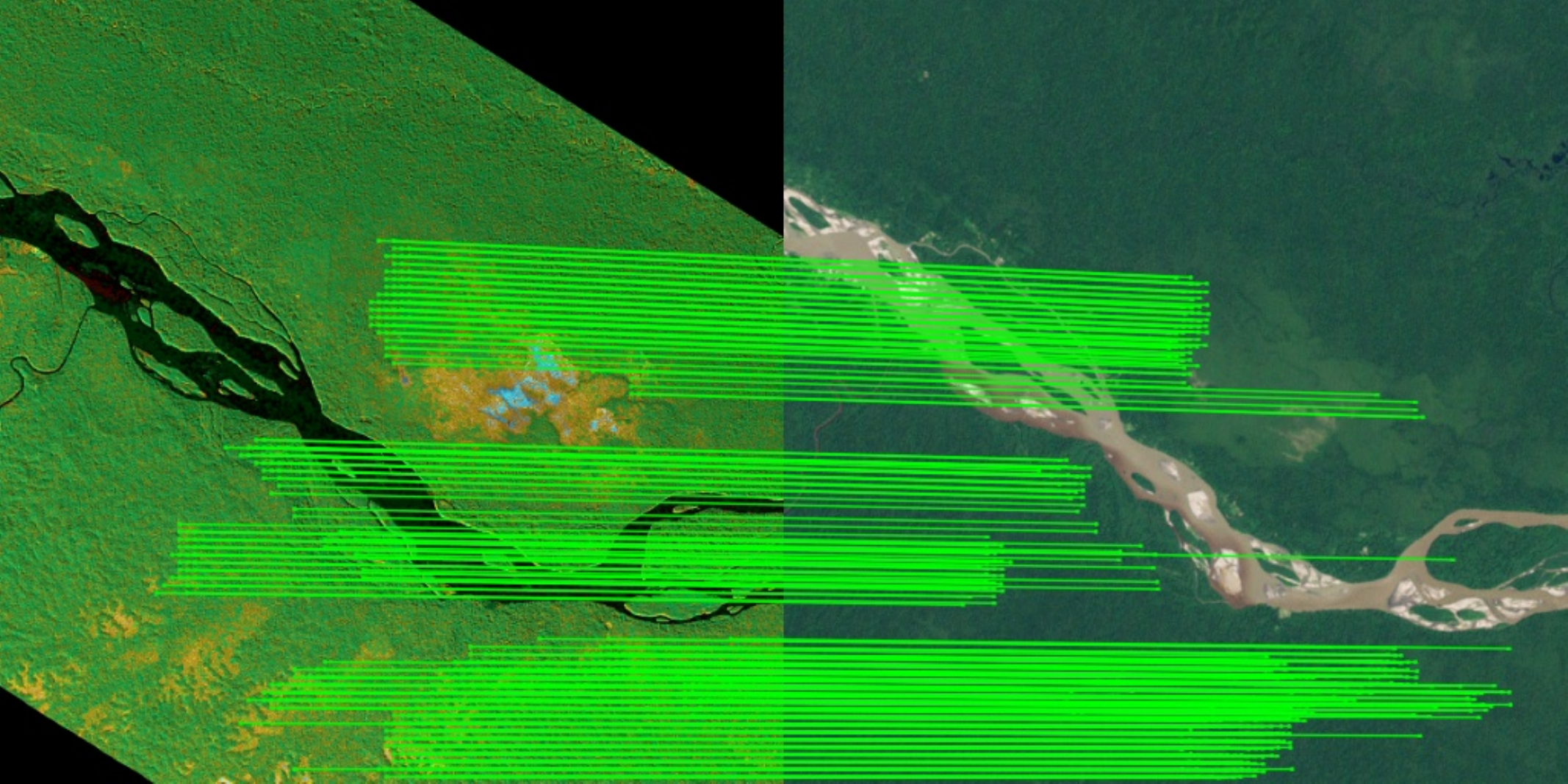}\vspace{1pt}
	\end{minipage}}
	\hspace{-0.1cm}\subfigure[]{  %Infrared-Optical
	\begin{minipage}[b]{0.24\textwidth}
	\includegraphics[width=1\linewidth,height=17mm]{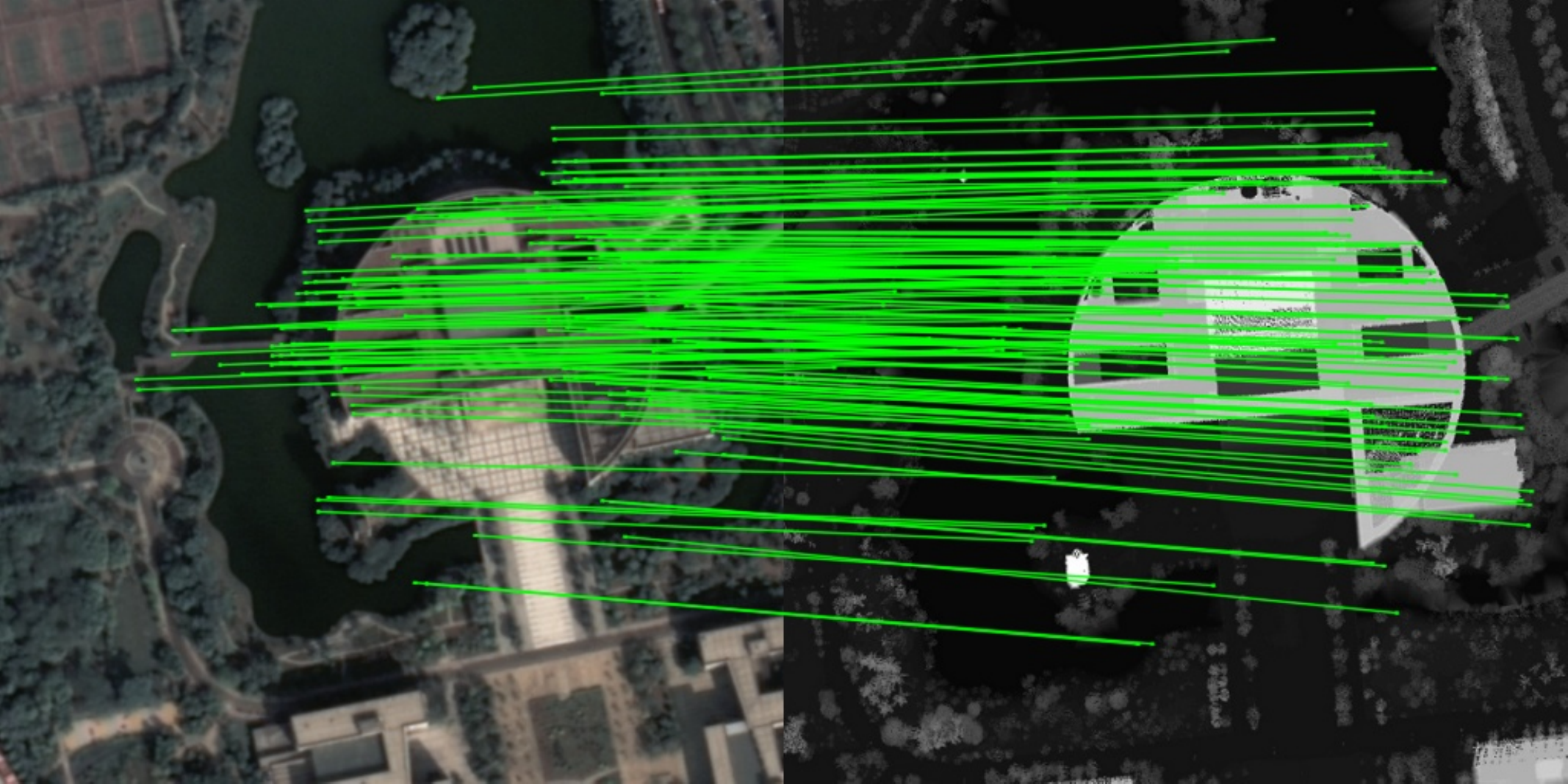}\vspace{1pt}
	\includegraphics[width=1\linewidth,height=17mm]{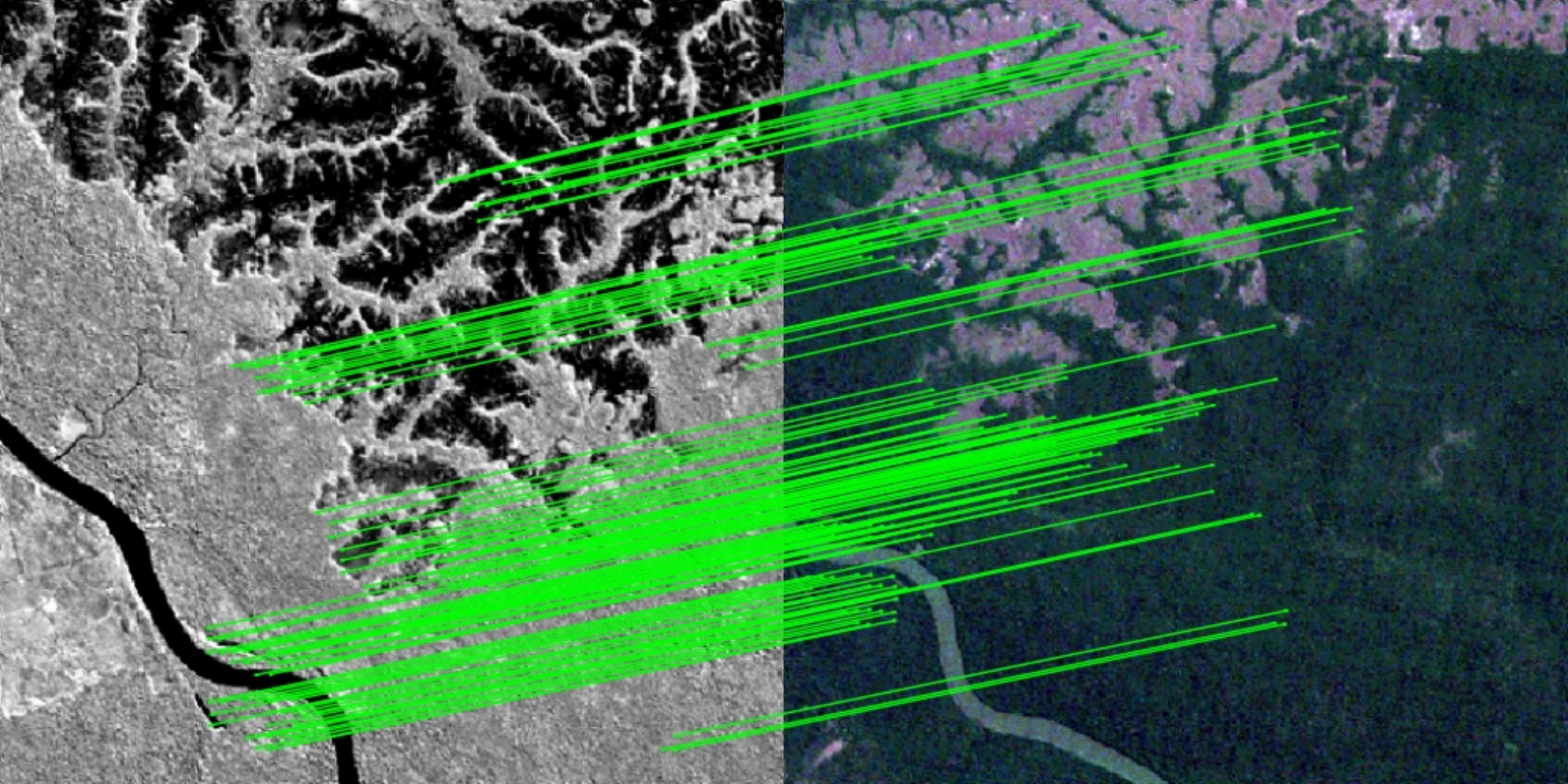}\vspace{1pt}
	\includegraphics[width=1\linewidth,height=17mm]{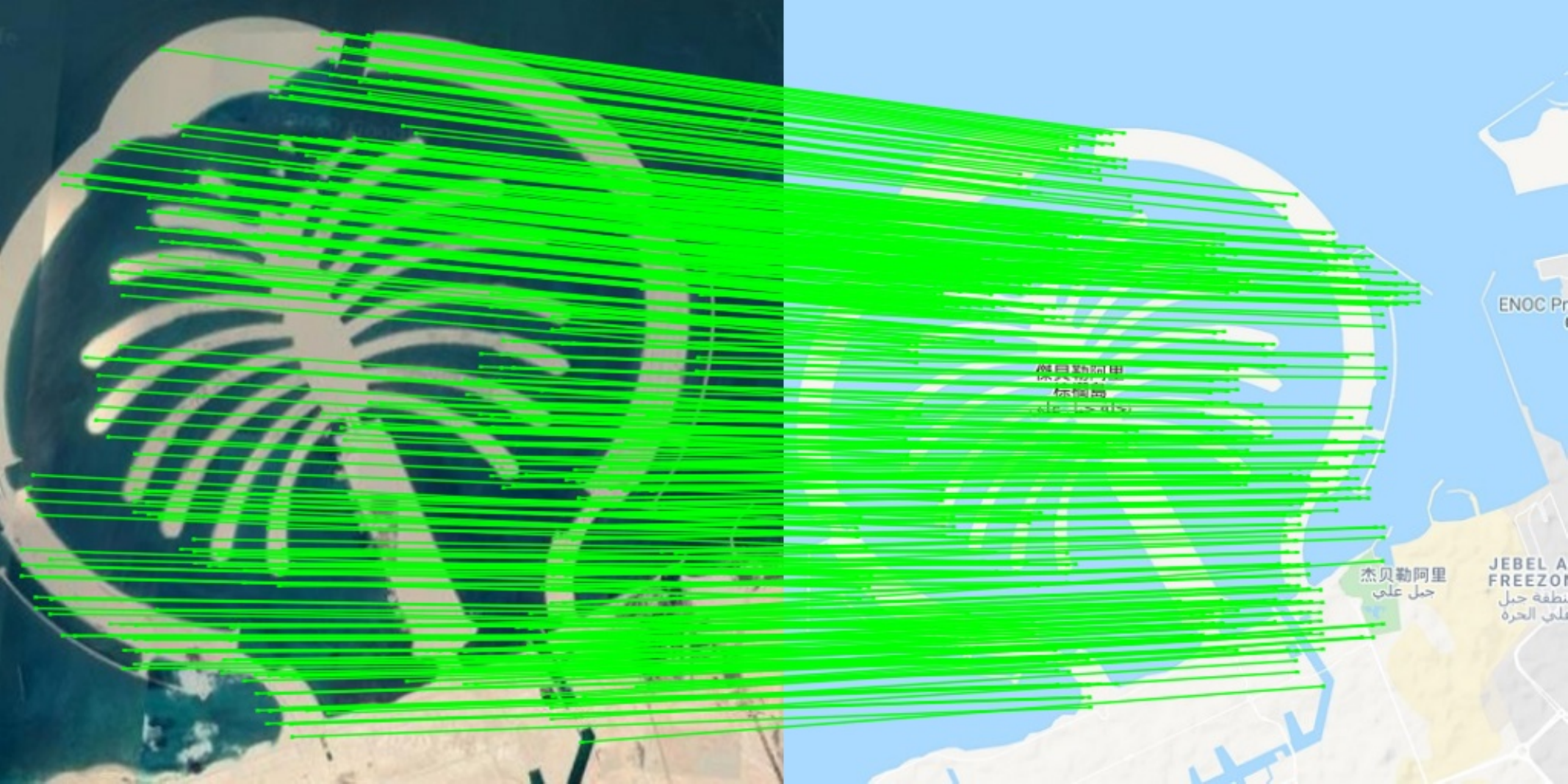}\vspace{1pt}
	\includegraphics[width=1\linewidth,height=17mm]{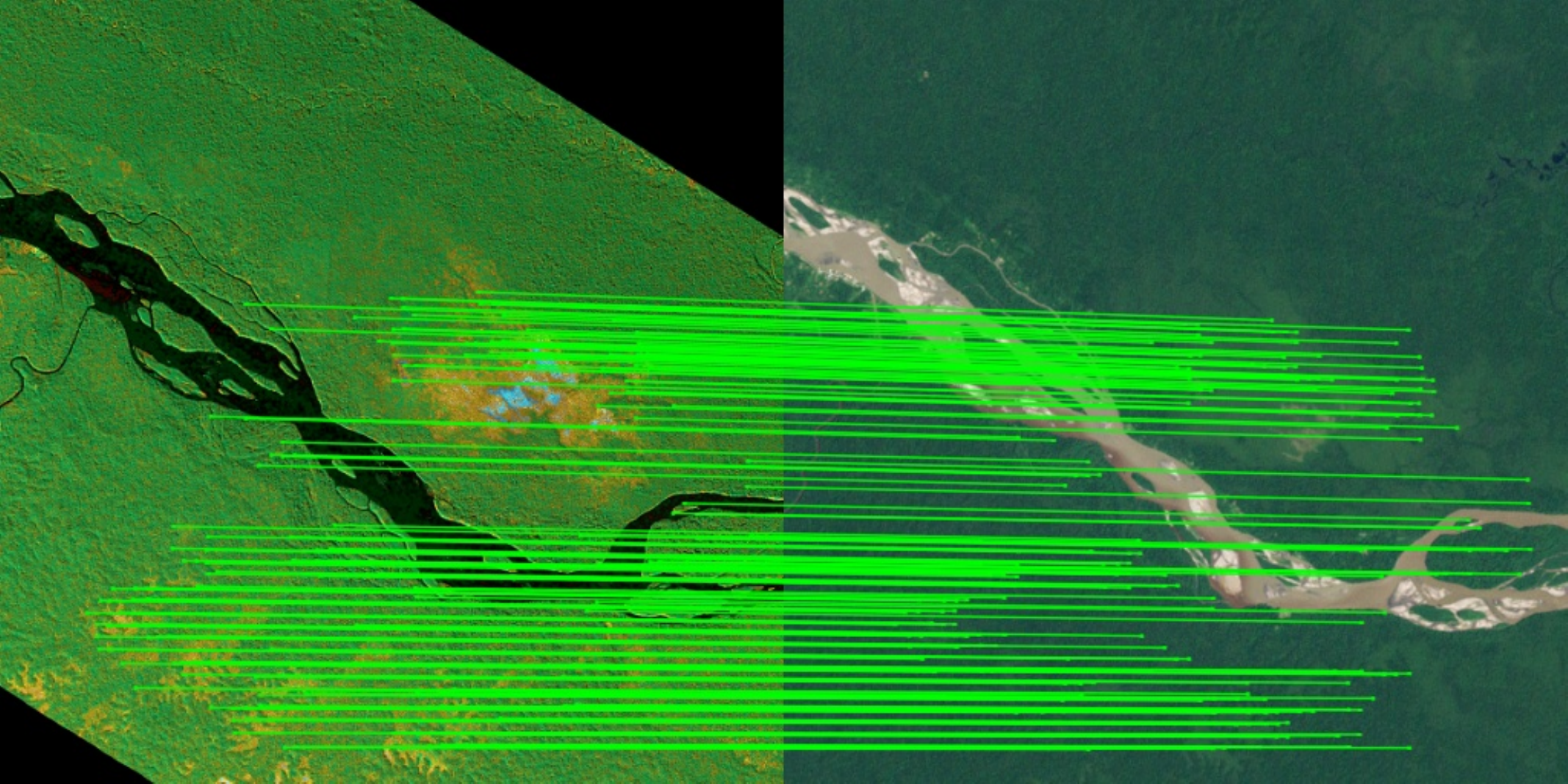}\vspace{1pt}
	\end{minipage}}
	\hspace{-0.1cm}\subfigure[]{  % Optical-Map
	\begin{minipage}[b]{0.24\textwidth}
	\includegraphics[width=1\linewidth,height=17mm]{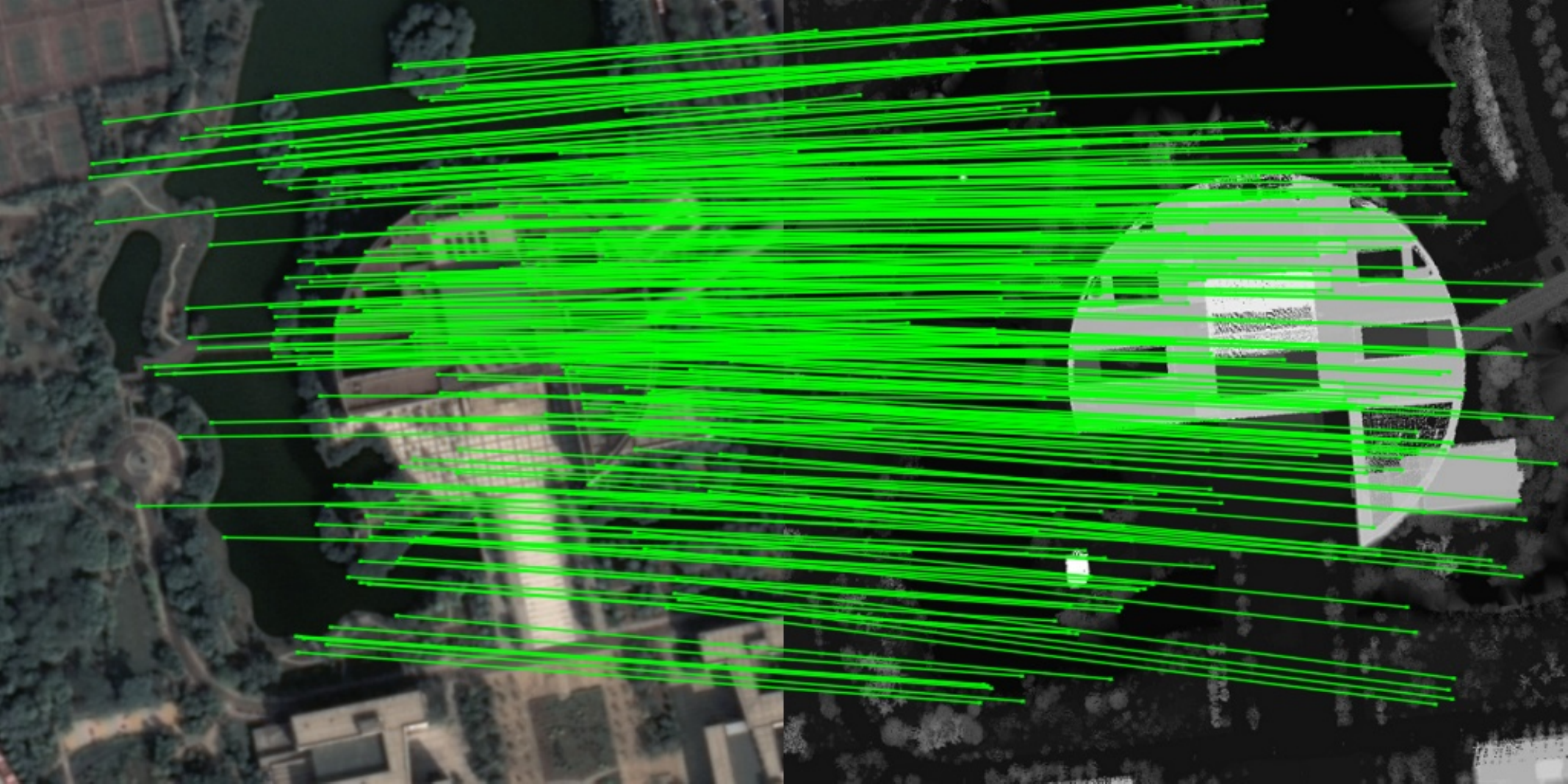}\vspace{1pt}
	\includegraphics[width=1\linewidth,height=17mm]{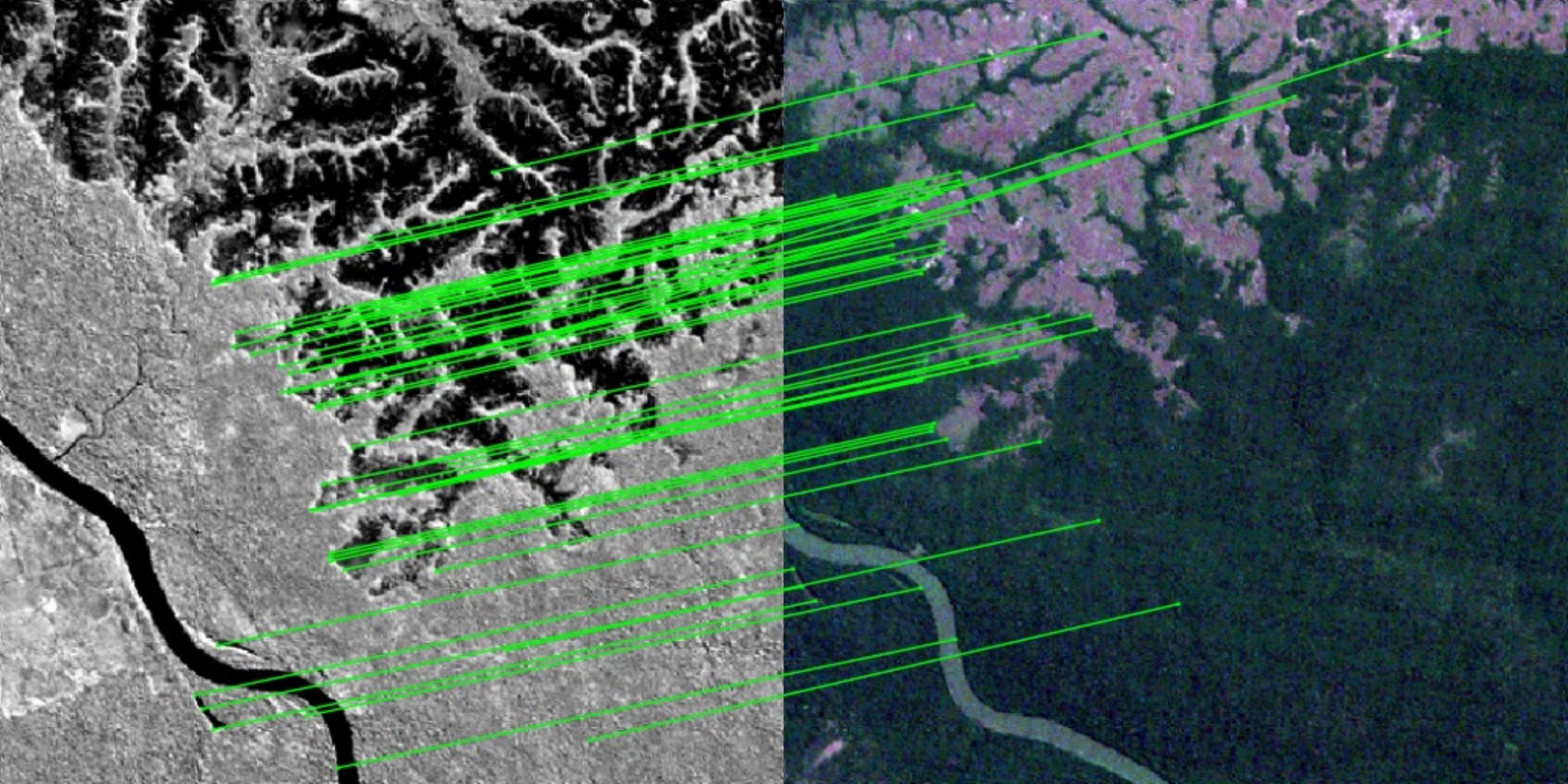}\vspace{1pt}
	\includegraphics[width=1\linewidth,height=17mm]{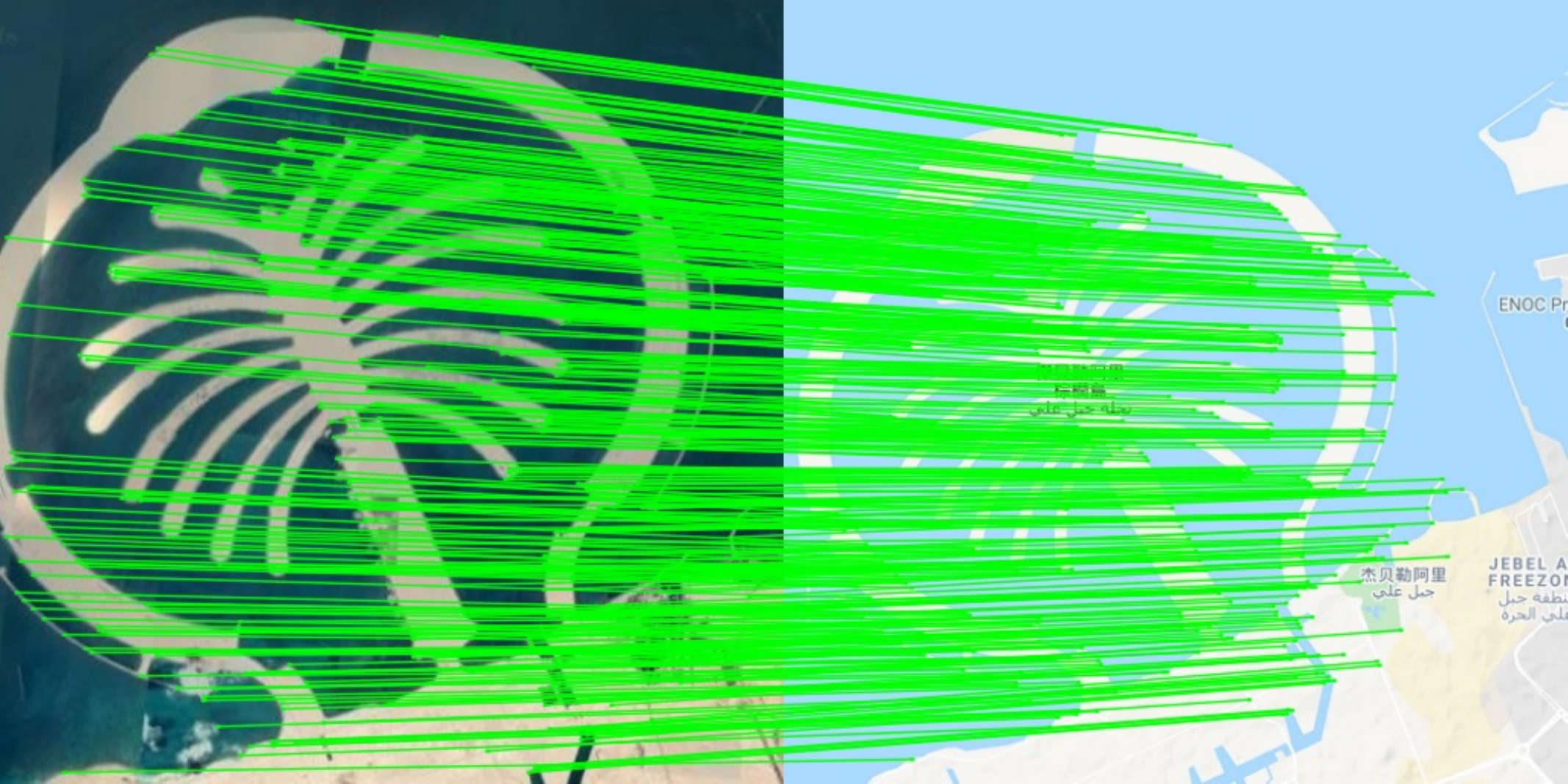}\vspace{1pt}
	\includegraphics[width=1\linewidth,height=17mm]{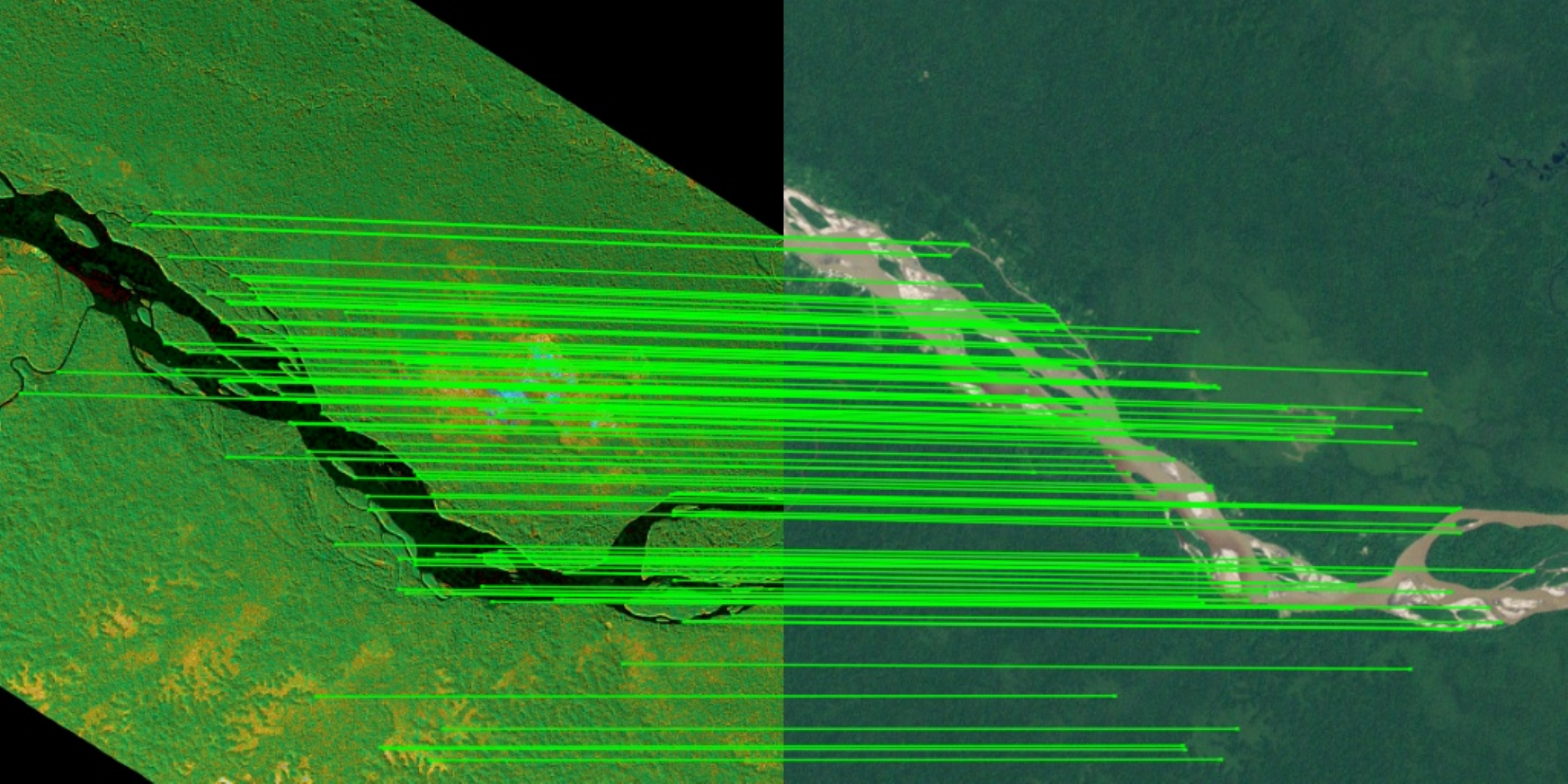}\vspace{1pt}
	\end{minipage}}
	\hspace{-0.1cm}\subfigure[]{  % SAR-Optical
	\begin{minipage}[b]{0.24\textwidth}
	\includegraphics[width=1\linewidth,height=17mm]{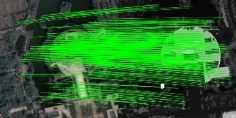}\vspace{1pt}
	\includegraphics[width=1\linewidth,height=17mm]{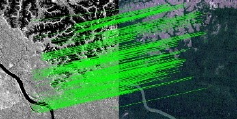}\vspace{1pt}
	\includegraphics[width=1\linewidth,height=17mm]{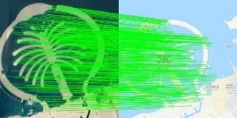}\vspace{1pt}
	\includegraphics[width=1\linewidth,height=17mm]{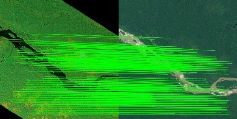}\vspace{1pt}
	\end{minipage}}

	\caption{Comparison results of eight methods on the optical-depth, infrared-optical, optical-map, and SAR-optical datasets without noise. (a) HAPCG. (b) RIFT. (c) LoFTR. (d) MatchFormer. (e) ECO-TR. (f) TopicFM. (g) LightGlue. (h) Ours.}
	\label{Fig.8}
\end{figure*}

\subsection{Performance Metric}
\begin{table*}[htbp]
	\caption{THE QUANTITATIVE COMPARISON RESULTS ON MULTISOURCE REMOTE SENSING IMAGES}
		\centering
		\fontsize{8.5}{11}\selectfont  
		\resizebox{\linewidth}{!}{		
			\begin{tabular}{|p{1.9cm}<{\centering}|p{2.0cm}<{\centering}|p{2.2cm}<{\centering}|p{2.2cm}<{\centering}|p{2.2cm}<{\centering}|p{2.2cm}<{\centering}|p{2.2cm}<{\centering}|}			
			\hline	
			\multicolumn{2}{|c|}{\diagbox[width=4.5cm]{Method\ /\ Metric}{Data}} & \makecell{ Dataset\ 1 \\ Optical-Depth} & \makecell{ Dataset\ 2 \\ Infrared-Optical} & \makecell{ Dataset\ 3 \\ Optical-Map} & \makecell{Dataset\ 4 \\ SAR-Optical} & Average  \cr
			\hline	
			\multirow{4}*{\makecell{\textbf{HAPCG}}}
			&NCM \bm{$\uparrow$}           & 219	& 256  & 259  & 124   & 214.5   \\
			&SR(\%)\bm{$\uparrow$}         & 90	    & 100  & 70	  & 80	  & 85	    \\
			&RMSE(pixels)\bm{$\downarrow$} & 3.92	& 1.87 & 7.26 & 5.55  & 4.65    \\
			&RT(s)\bm{$\downarrow$}        & 7.16	& 6.39 & 6.44 & 6.89  & 6.72    \\
			\hline
			\multirow{4}*{\makecell{\textbf{RIFT}}}
			&NCM \bm{$\uparrow$}           & 268	& 147	& 254	& 168	& 209.3   \\
			&SR(\%)\bm{$\uparrow$}         & 100	& 100	& 90	& 90	& 95	\\
			&RMSE(pixels)\bm{$\downarrow$} & 1.94	& 1.90	& 3.67  & 3.72  & 2.81  \\
			&RT(s)\bm{$\downarrow$}        & 6.23	& 6.08	& 6.63  & 6.73  & 6.42  \\
			\hline
			\multirow{4}*{\makecell{\textbf{LoFTR} \\ \textbf{+\ FSC}}}
			&NCM \bm{$\uparrow$}           & 387	& 406	& 239	& 246   & 319.5   \\
			&SR(\%)\bm{$\uparrow$}         & 100    & 100	& 80	& 100   & 95    \\
			&RMSE(pixels)\bm{$\downarrow$} & 2.69	& 2.92  & 8.11	& 2.15  & 3.97  \\
			&RT(s)\bm{$\downarrow$}        & 2.96	& 3.06  & 3.03	& 2.98  & 3.01  \\
			\hline
			\multirow{4}*{\makecell{\textbf{MatchFormer} \\ \textbf{+\ FSC}}}
			&NCM \bm{$\uparrow$}           & 535	& 434   & 299 	& 341	& 402.3   \\
			&SR(\%)\bm{$\uparrow$}         & 100    & 100   & 90    & 90	& 95    \\
			&RMSE(pixels)\bm{$\downarrow$} & 2.72	& 2.40  & 4.69	& 2.16	& 2.99  \\
			&RT(s)\bm{$\downarrow$}        & 3.00	& 2.95  & 2.89	& 3.03	& 2.97  \\
			\hline

			\multirow{4}*{\makecell{\textbf{ ECO-TR} \\ \textbf{ +\ FSC}}}     
			& NCM \bm{$\uparrow$}            &  1326  & 861   &  1703  &  221   &  1027.8  \\
			& SR(\%)\bm{$\uparrow$}          &  100   & 80    &  100   &  80    &  90      \\
			& RMSE(pixels)\bm{$\downarrow$}  &  2.58  & 5.98  &  2.59  &  6.08  &  4.31    \\
			& RT(s)\bm{$\downarrow$}         &  6.09  & 6.21  &  6.03  &  6.25  &  6.14    \\
			\hline
			\multirow{4}*{\makecell{\textbf{ TopicFM} \\ \textbf{ +\ FSC}}}
			& NCM \bm{$\uparrow$}            &  307   & 438   &  412   &  208   &  341.2   \\
			& SR(\%)\bm{$\uparrow$}          &  100   & 100   &  100   &  100   &  100     \\
			& RMSE(pixels)\bm{$\downarrow$}  &  2.28  & 2.32  &  2.03  &  2.42  &  2.26    \\
			& RT(s)\bm{$\downarrow$}         &  2.05  & 2.08  &  2.02  &  2.02  &  2.04    \\
			\hline
			\multirow{4}*{\makecell{\textbf{ LightGlue} \\ \textbf{ +\ FSC}}}
			& NCM \bm{$\uparrow$}            &  384   & 221   &  315   &  154   &  268.4   \\
			& SR(\%)\bm{$\uparrow$}          &  100   & 100   &  100   &  80    &  95      \\
			& RMSE(pixels)\bm{$\downarrow$}  &  2.03  & 2.38  &  1.96  &  5.84  &  3.05    \\
			& RT(s)\bm{$\downarrow$}         &  4.23  & 3.85  &  3.95  &  3.89  &  3.98    \\
			\hline
			\multirow{4}*{\makecell{\textbf{Ours}}}
			&NCM \bm{$\uparrow$}           & 993   & 551   & 737   & 685   & 741.5  \\
			&SR(\%)\bm{$\uparrow$}         & 100   & 100   & 80    & 100   & 95     \\
			&RMSE(pixels)\bm{$\downarrow$} & 2.84  & 2.87  & 6.92  & 2.90  & 3.88   \\
			&RT(s)\bm{$\downarrow$}        & 3.26  & 3.20  & 3.10  & 3.12  & 3.17   \\
			\hline	
			\end{tabular}
		}
		% \end{center}
		\label{TABLE I}
\end{table*}
According to the evaluation methods for multimodal remote sensing images in references \cite{ref57} and \cite{ref58}, 
we use the number of correct matches (NCM), success rate (SR), root mean square error (RMSE), and run time (RT) as the quantitative evaluation metrics.

\emph{1)} The NCM is defined as the number of matches with matching errors of less than 3 pixels in horizontal and vertical directions.

\emph{2)} The SR is the ratio between correctly matched image pairs and the total number of image pairs. 
If the NCM of image pairs exceeds 10, the image pair is considered a successful match; conversely, too small NCM reflects the lack of robustness for the method.

\emph{3)} RMSE is calculated as follows:

\begin{equation}
	\label{deqn_ex23}
	\text{RMSE}=\sqrt{\frac{\sum\nolimits_{i=1}^{N}{{{\left( u_{i}^{B}-u_{i}^{A\to B} \right)}^{2}}+{{\left( v_{i}^{B}-v_{i}^{A\to B} \right)}^{2}}}}{N}}
\end{equation}	
where $N$ is the number of correct matches, 
$\left( u_{i}^{B},v_{i}^{B} \right)$ denotes the coordinates of the i-th feature point $\mathbf{x}_{i}^{B}$ in the reference image $I^{B}$, 
$\left( u_{i}^{A\to B},v_{i}^{A\to B} \right)$ is the coordinates of the i-th feature point $\mathbf{x}_{i}^{A}=\left( u_{i}^{A},v_{i}^{A} \right)$ 
in the query image $I^{B}$ mapped to the reference image through matching correspondence transformation. 
Here, we use the homography transformation $H$ as the transformation relationship, i.e., $\mathbf{x}_{i}^{A\to B}=H \mathbf{x}_{i}^{A}$; 
and the estimation of the homography matrix $H$ is computed by using the library function cv2.findHomography from OpenCV. 
If an image pair is not successfully matched (NCM $<10$), its RMSE is set to be 20 pixels.

\subsection{Matching Performance Evaluation}

To evaluate the effectiveness of the proposed method, we designed performance comparison experiments 
to compare our method with state-of-the-art image matching methods e.g. HAPCG \cite{ref70}, RIFT \cite{ref57}, LoFTR \cite{ref24}, MatchFormer \cite{ref71}, ECO-TR \cite{ref72}, TopicFM\cite{ref73}, and LightGlue\cite{ref74}. 
To sum up, HAPCG is a matching method that considers anisotropic weighted moments and the histogram of absolute phase consistency gradients. 
RIFT uses phase coherence for feature point detection and maximum index map for feature description, the method exhibits rotational invariance and is suitable for matching multi-modal images. 
Local Feature Transformer (LoFTR) utilizes a transformer with self-attention and cross-attention layers to transform local features into information with context and position relevance. 
MatchFormer propose a hierarchical extraction and matching transformer that synchronously performs multi-level feature extraction and feature similarity learning by interleaving cross-attention and self-attention. 
ECO-TR propose a new coarse-to-fine transformer-based framework for finding correspondence that can be applied to both sparse and dense matching tasks.
TopicFM present a robust method that fuses local context and high-level semantic information into latent features using a topic modeling strategy. 
LightGlue aims to significantly increase inference speed while reinforcing matching performance. 
The code for these comparison methods is obtained from the original authors' publications. 
All method parameters are set according to the original papers. 

For the above methods, HAPCG and RIFT use the fast sample consensus (FSC) \cite{ref75} algorithm to eliminate incorrect matches, 
while LoFTR, MatchFormer, ECO-TR, TopicFM and LightGlue based on deep learning methods do not consider additional outlier removal strategies. 
To ensure fairness in the experiments, we add the FSC outlier removal algorithm to the above deep learning methods. 
Compared with the classic RANSAC algorithm, FSC can get more correct matches in less number of iterations. 
Additionally, HAPCG and RIFT are implemented in the MATLAB R2021b, while other deep learning-based methods are implemented in the PyTorch framework. 
Testing and training were conducted on the same experimental platform.

Results analysis and comparison: Fig. \ref{Fig.8} shows the qualitative results of HAPCG, RIFT, LoFTR, MatchFormer, and our proposed method. 
Compared to other methods, our method demonstrates more dense and accurate matching, particularly in scenes with scale differences and low textures (third and fourth columns of Fig. \ref{Fig.8}). 
Table \ref{TABLE I} summarizes the quantitative results obtained by the eight methods on each dataset, where lower values of RMSE and RT are preferable, and higher values of SR and NCM are desirable. 

\begin{figure*}[hbp]
	\centering
	\hspace{-0.1cm}\subfigure[]{  % Optical-Depth
	\begin{minipage}[b]{0.24\textwidth}
	\includegraphics[width=1\linewidth,height=17mm]{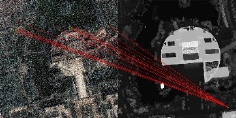}\vspace{1pt}
	\includegraphics[width=1\linewidth,height=17mm]{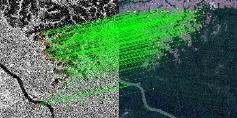}\vspace{1pt}
	\includegraphics[width=1\linewidth,height=17mm]{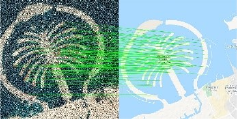}\vspace{1pt}
	\includegraphics[width=1\linewidth,height=17mm]{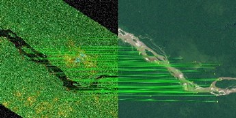}\vspace{1pt}
	\end{minipage}}
	\hspace{-0.1cm}\subfigure[]{  %Infrared-Optical
	\begin{minipage}[b]{0.24\textwidth}
	\includegraphics[width=1\linewidth,height=17mm]{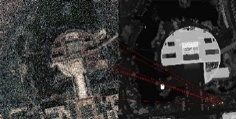}\vspace{1pt}
	\includegraphics[width=1\linewidth,height=17mm]{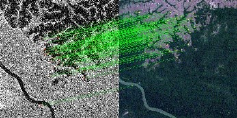}\vspace{1pt}
	\includegraphics[width=1\linewidth,height=17mm]{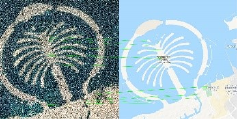}\vspace{1pt}
	\includegraphics[width=1\linewidth,height=17mm]{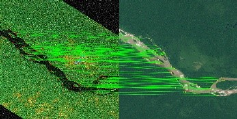}\vspace{1pt}
	\end{minipage}}
	\hspace{-0.1cm}\subfigure[]{  % Optical-Map
	\begin{minipage}[b]{0.24\textwidth}
	\includegraphics[width=1\linewidth,height=17mm]{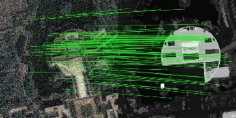}\vspace{1pt}
	\includegraphics[width=1\linewidth,height=17mm]{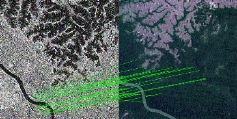}\vspace{1pt}
	\includegraphics[width=1\linewidth,height=17mm]{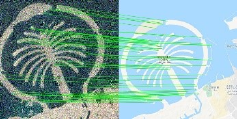}\vspace{1pt}
	\includegraphics[width=1\linewidth,height=17mm]{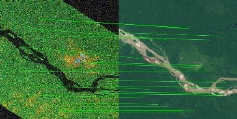}\vspace{1pt}
	\end{minipage}}
	\hspace{-0.1cm}\subfigure[]{  % SAR-Optical
	\begin{minipage}[b]{0.24\textwidth}
	\includegraphics[width=1\linewidth,height=17mm]{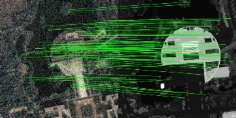}\vspace{1pt}
	\includegraphics[width=1\linewidth,height=17mm]{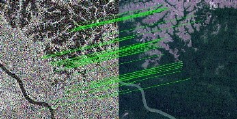}\vspace{1pt}
	\includegraphics[width=1\linewidth,height=17mm]{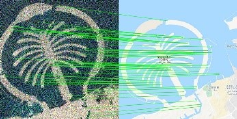}\vspace{1pt}
	\includegraphics[width=1\linewidth,height=17mm]{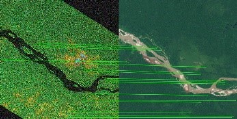}\vspace{1pt}
	\end{minipage}}
	\\
	\hspace{-0.1cm}\subfigure[]{  % SAR-Optical
	\begin{minipage}[b]{0.24\textwidth}
	\includegraphics[width=1\linewidth,height=17mm]{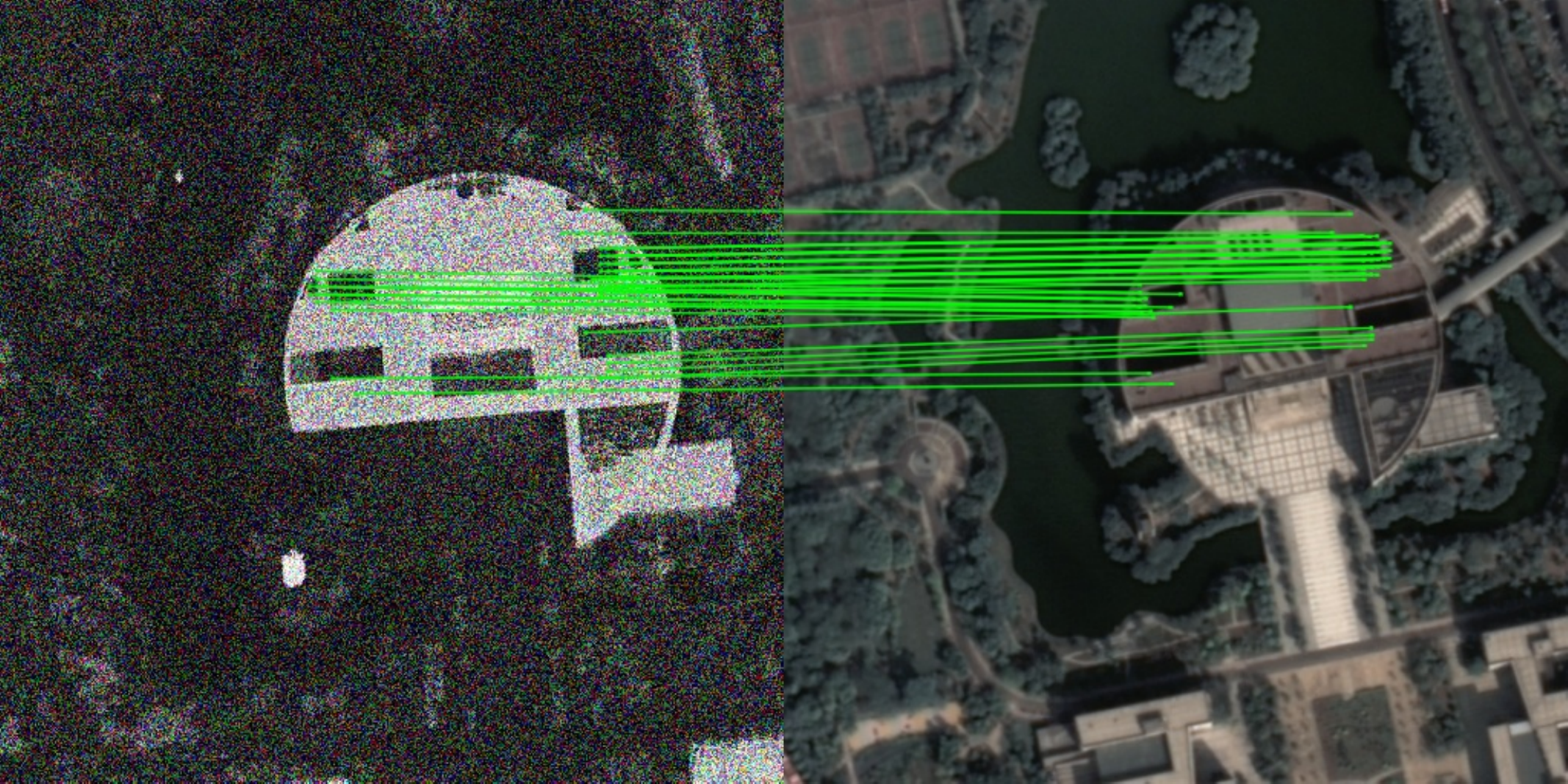}\vspace{1pt}
	\includegraphics[width=1\linewidth,height=17mm]{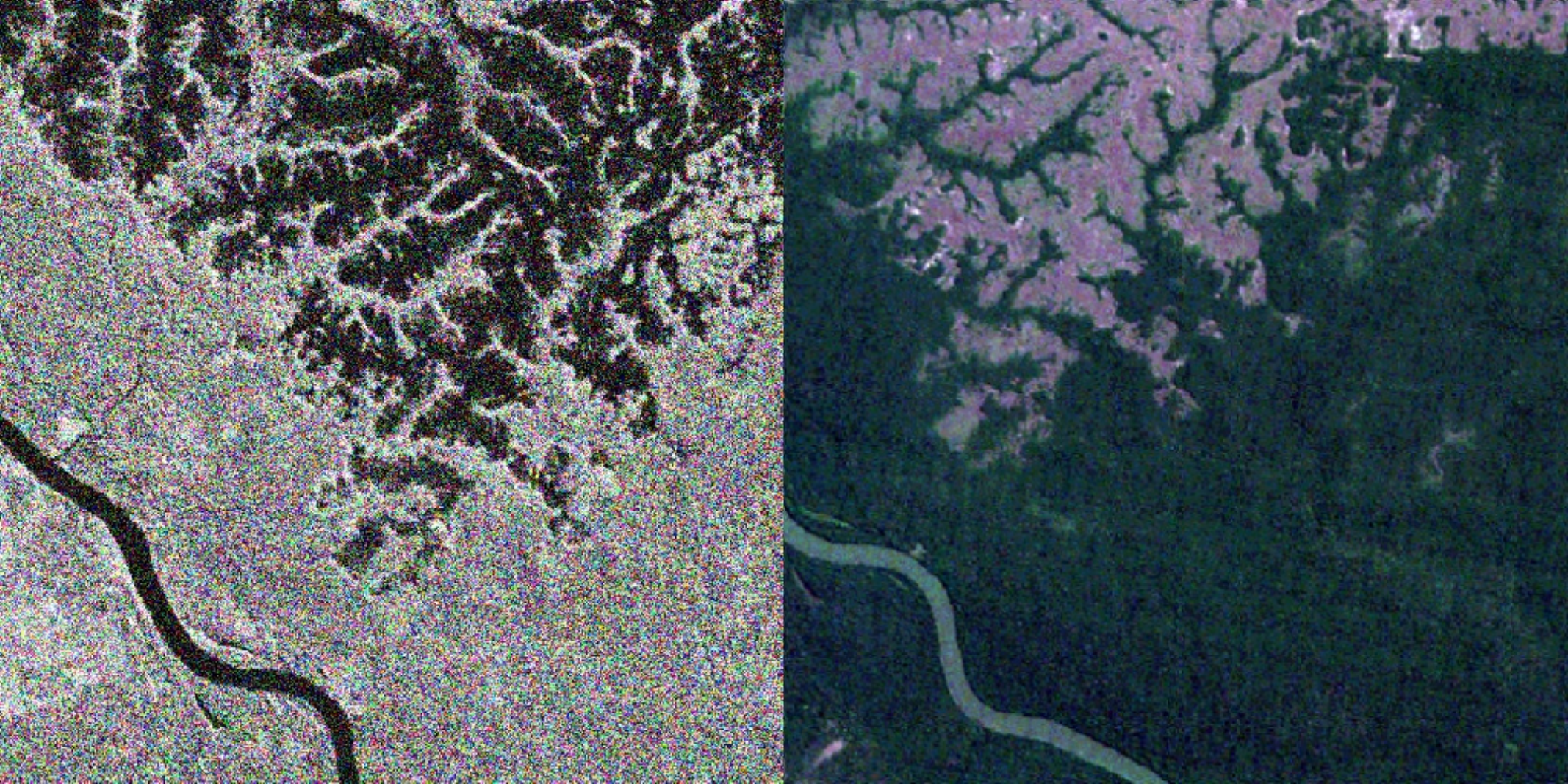}\vspace{1pt}
	\includegraphics[width=1\linewidth,height=17mm]{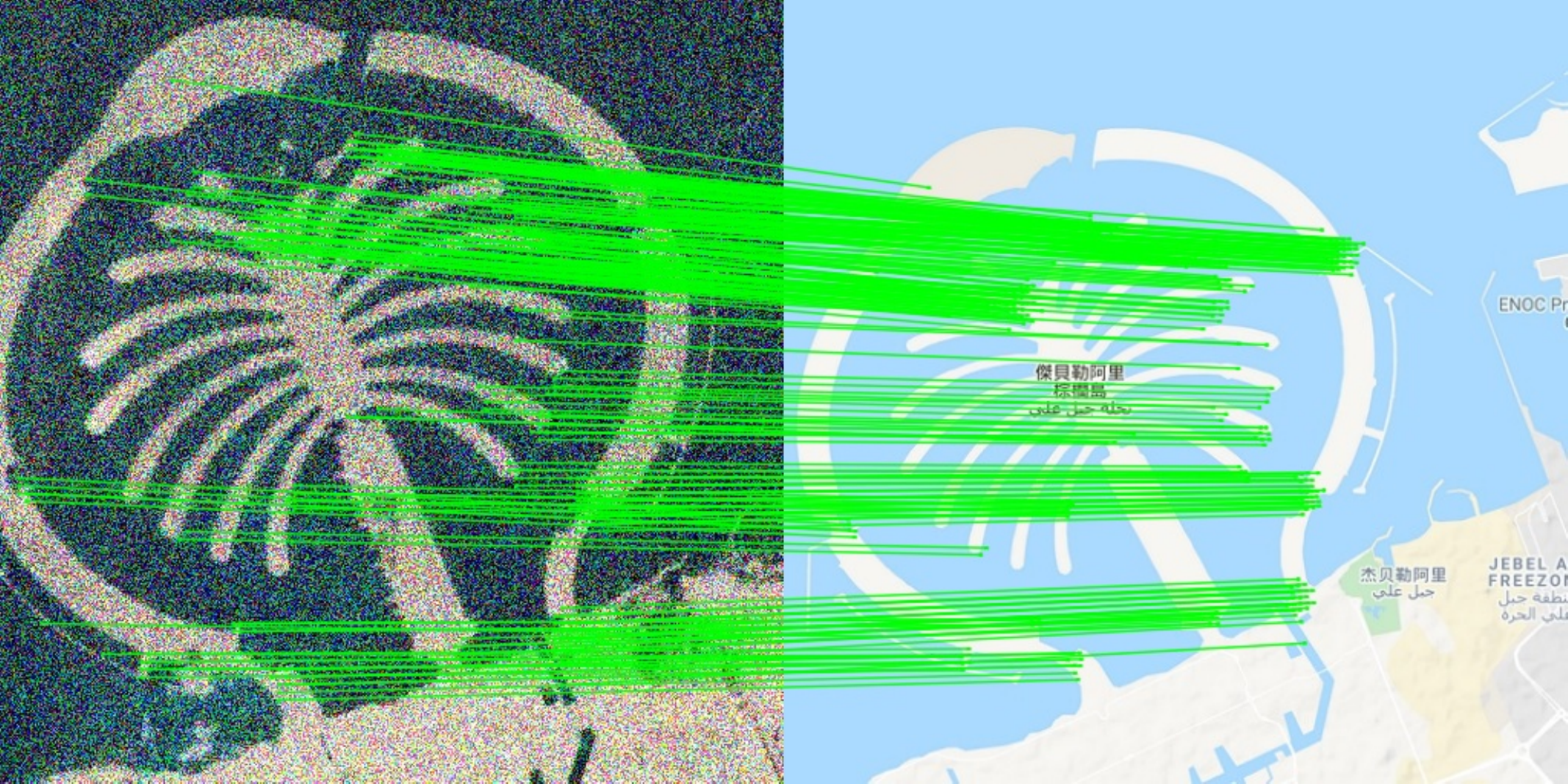}\vspace{1pt}
	\includegraphics[width=1\linewidth,height=17mm]{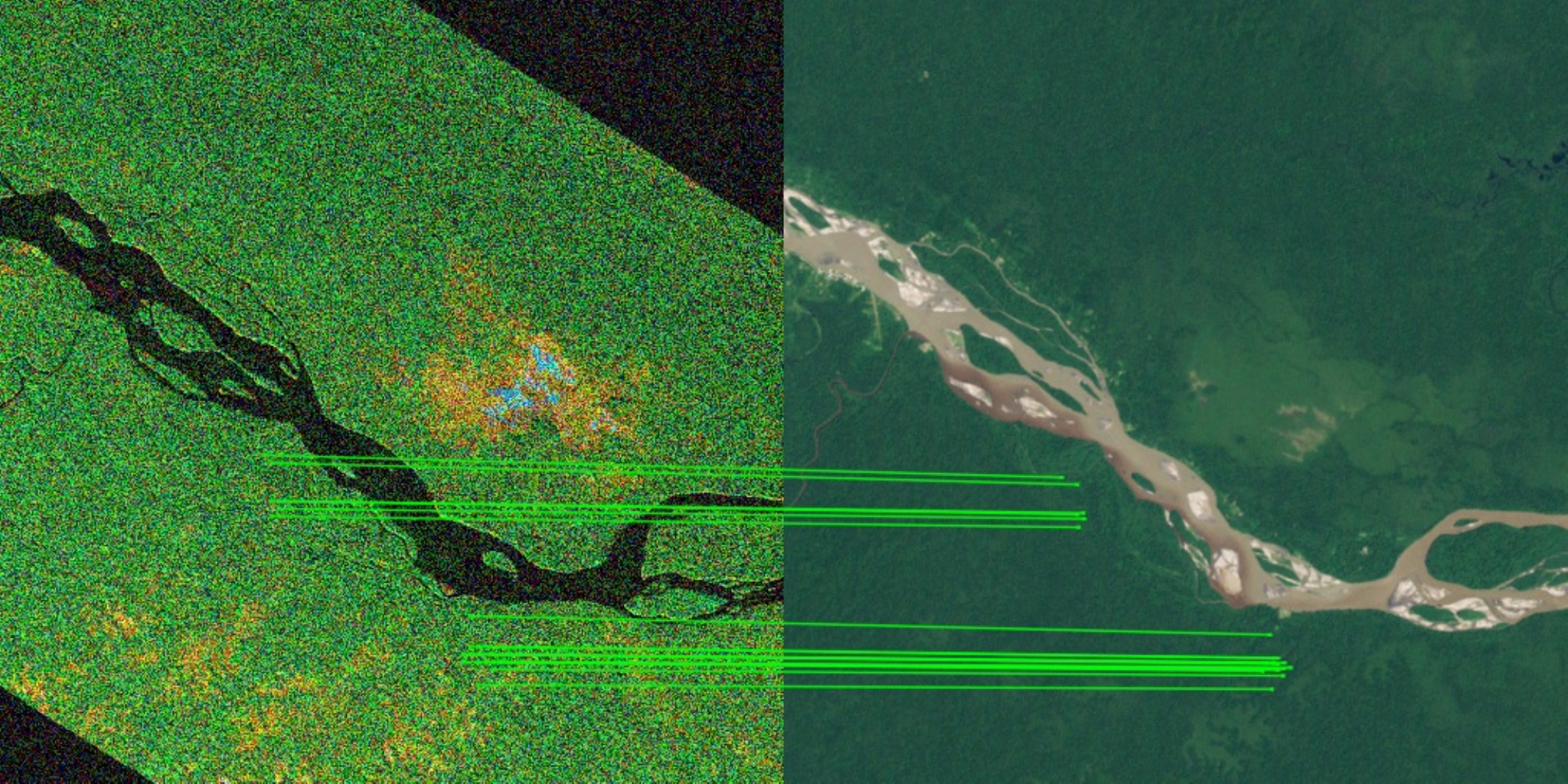}\vspace{1pt}
	\end{minipage}}
	\hspace{-0.1cm}\subfigure[]{  %Infrared-Optical
	\begin{minipage}[b]{0.24\textwidth}
	\includegraphics[width=1\linewidth,height=17mm]{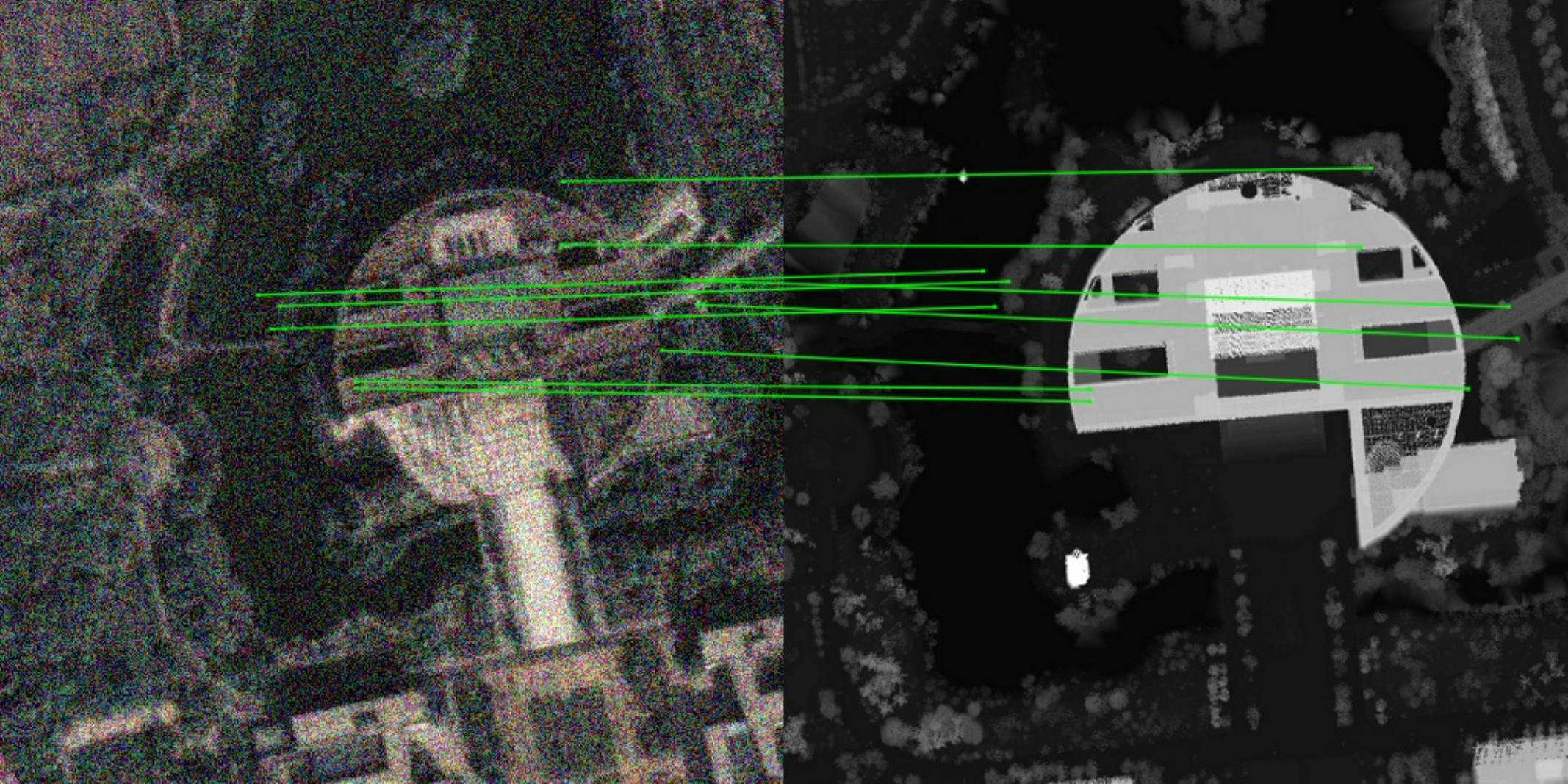}\vspace{1pt}
	\includegraphics[width=1\linewidth,height=17mm]{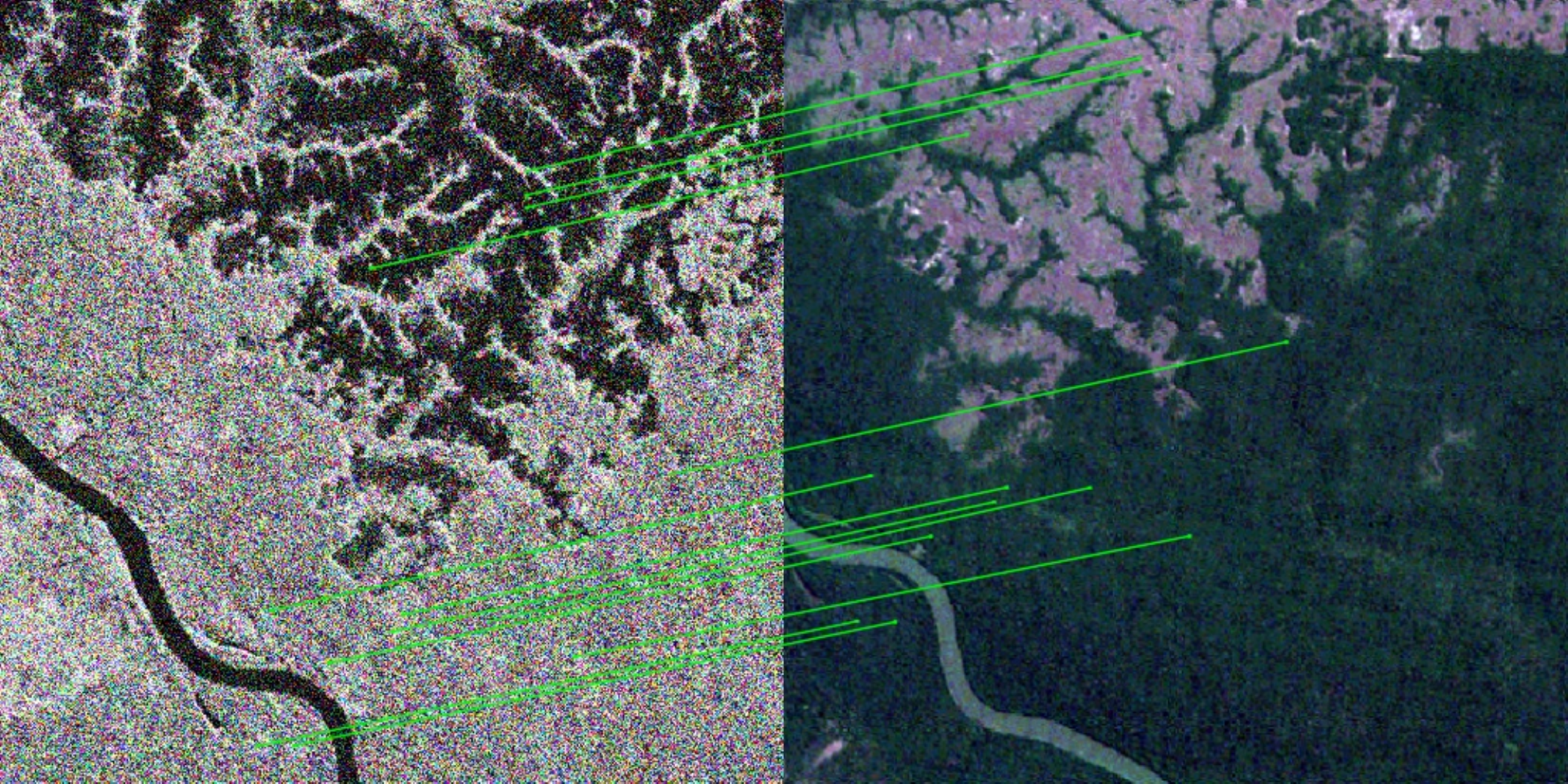}\vspace{1pt}
	\includegraphics[width=1\linewidth,height=17mm]{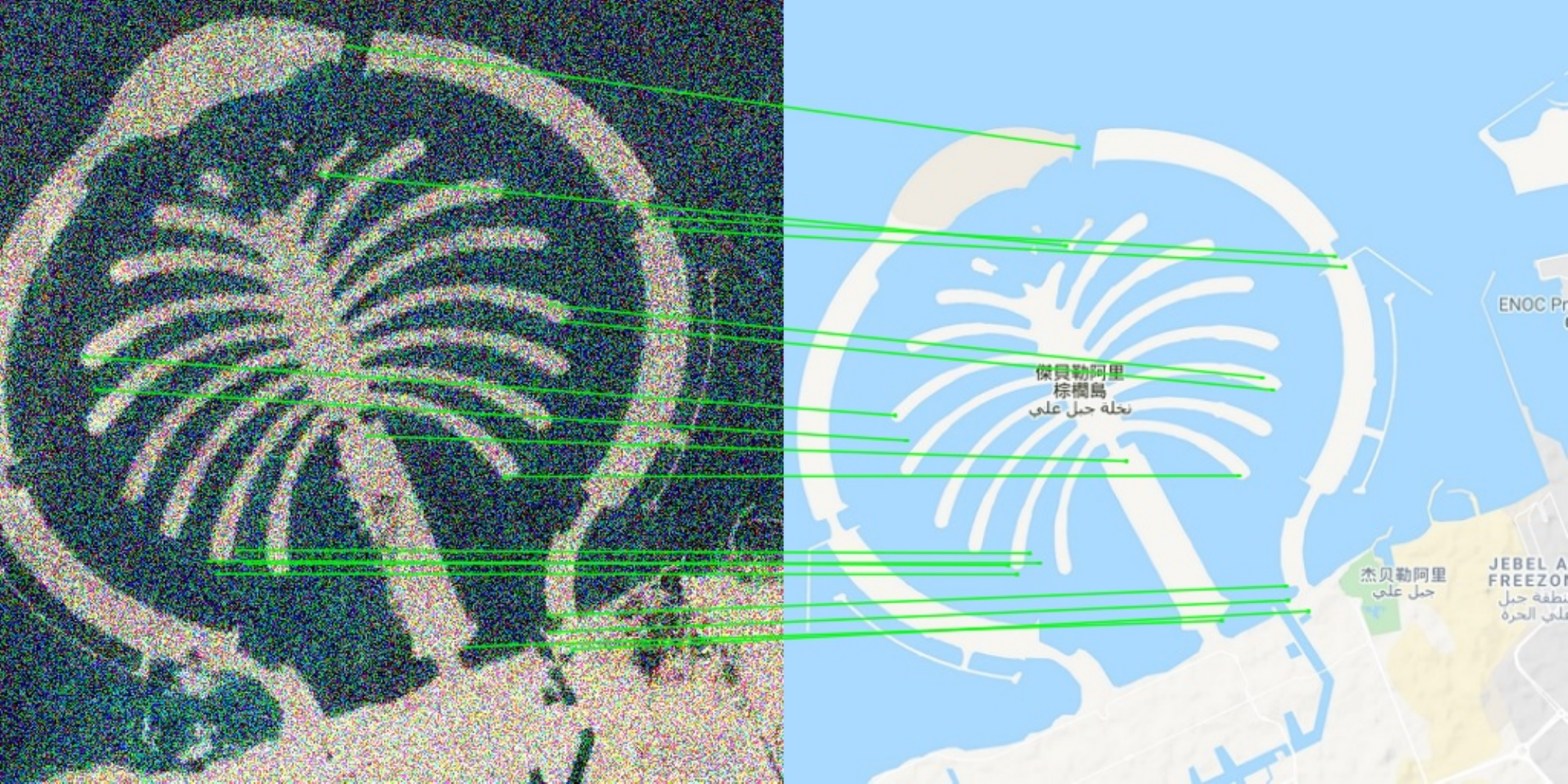}\vspace{1pt}
	\includegraphics[width=1\linewidth,height=17mm]{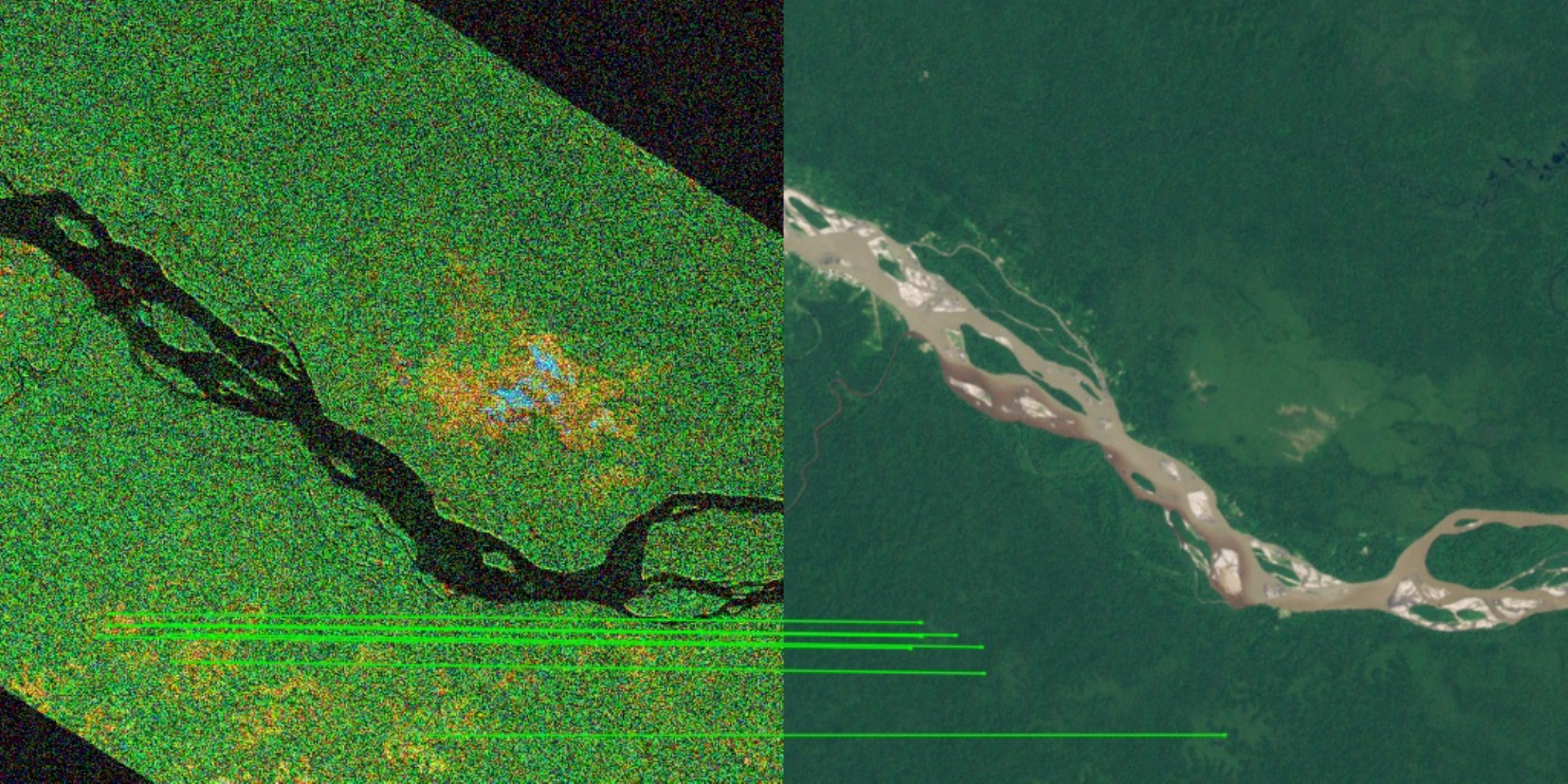}\vspace{1pt}
	\end{minipage}}
	\hspace{-0.1cm}\subfigure[]{  % Optical-Map
	\begin{minipage}[b]{0.24\textwidth}
	\includegraphics[width=1\linewidth,height=17mm]{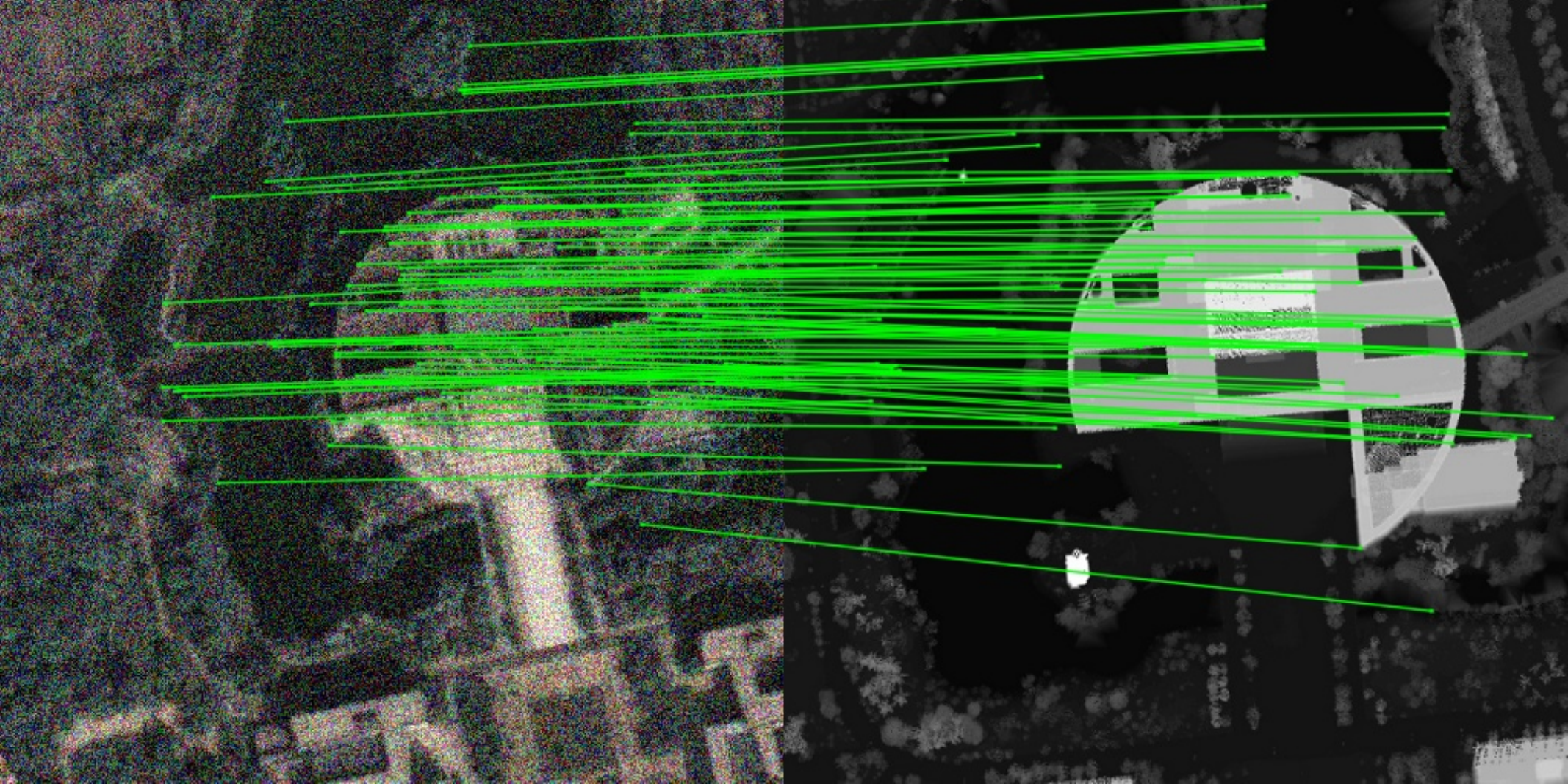}\vspace{1pt}
	\includegraphics[width=1\linewidth,height=17mm]{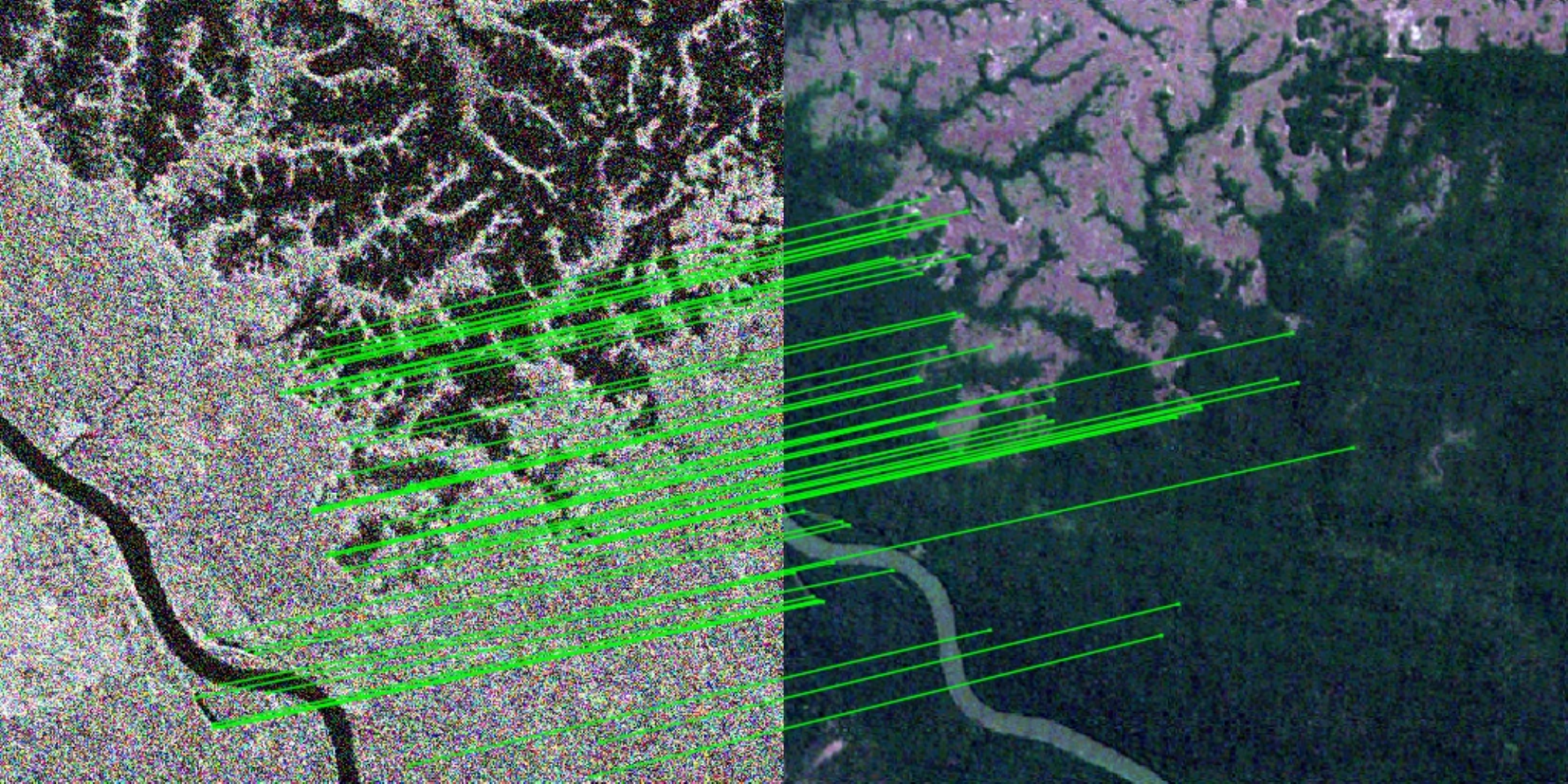}\vspace{1pt}
	\includegraphics[width=1\linewidth,height=17mm]{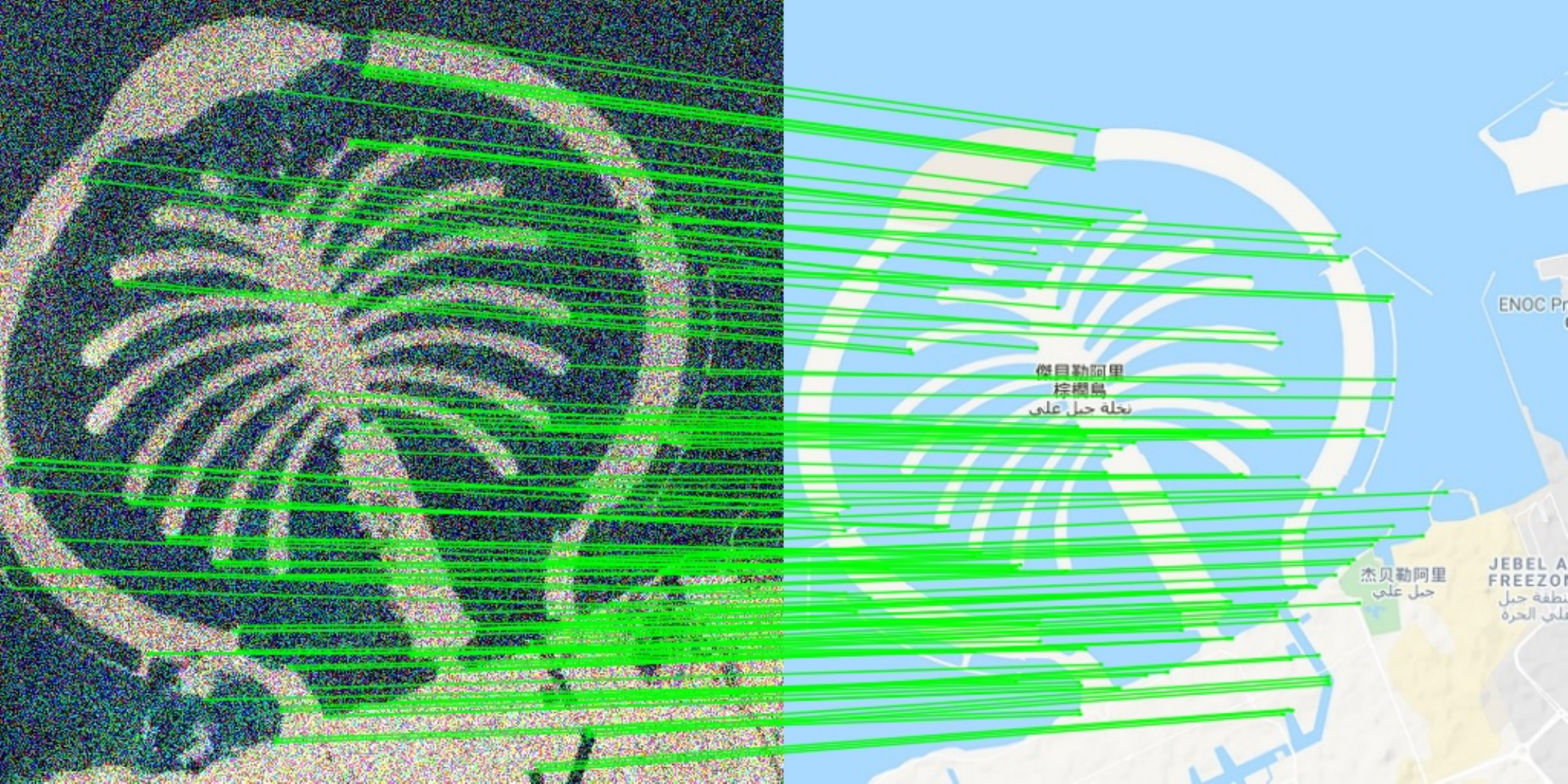}\vspace{1pt}
	\includegraphics[width=1\linewidth,height=17mm]{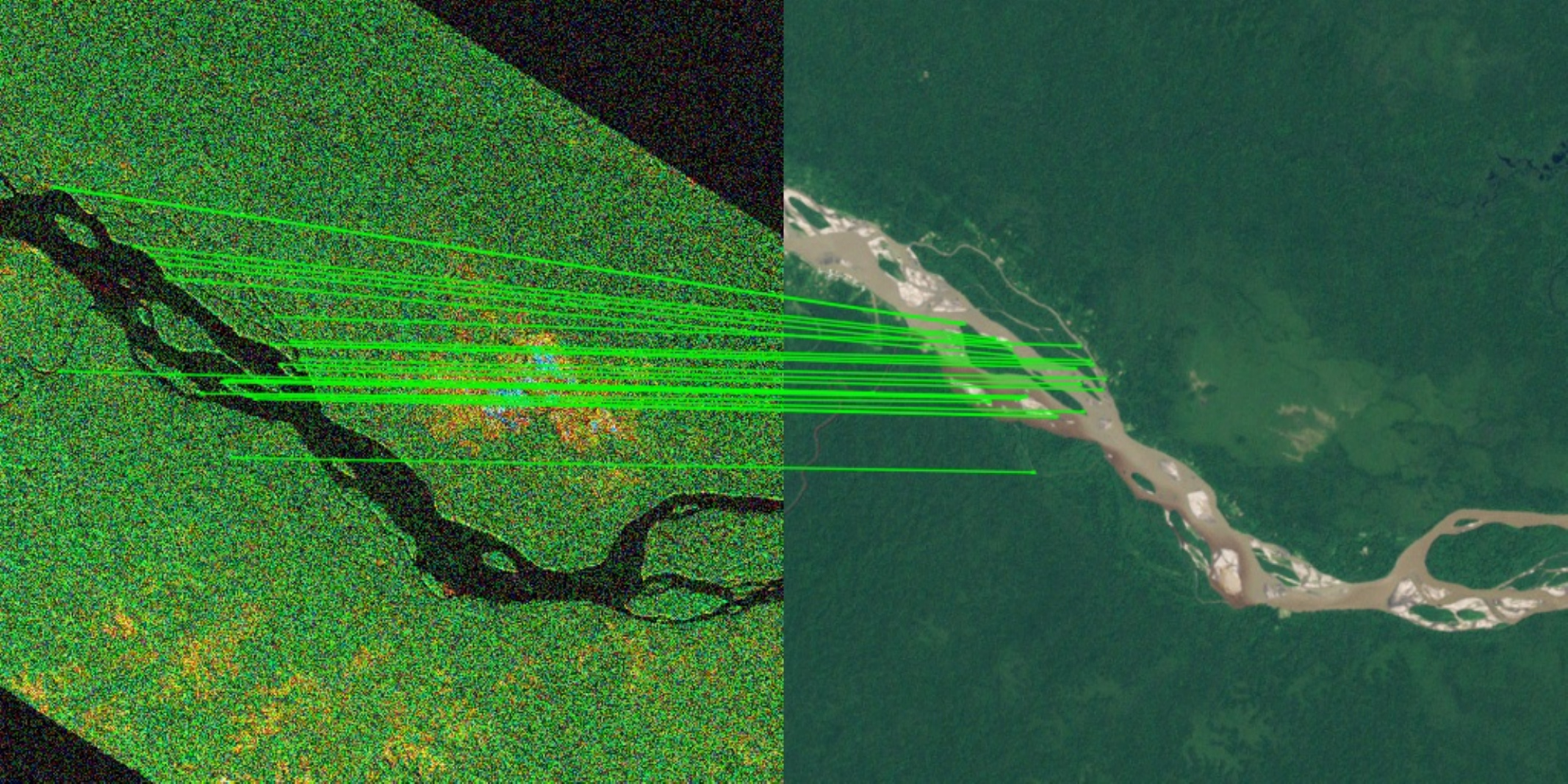}\vspace{1pt}
	\end{minipage}}
	\hspace{-0.1cm}\subfigure[]{  % SAR-Optical
	\begin{minipage}[b]{0.24\textwidth}
	\includegraphics[width=1\linewidth,height=17mm]{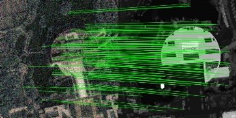}\vspace{1pt}
	\includegraphics[width=1\linewidth,height=17mm]{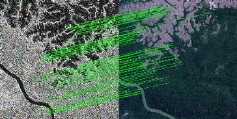}\vspace{1pt}
	\includegraphics[width=1\linewidth,height=17mm]{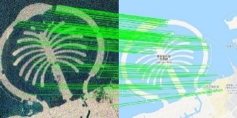}\vspace{1pt}
	\includegraphics[width=1\linewidth,height=17mm]{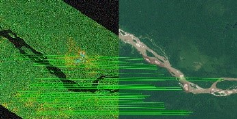}\vspace{1pt}
	\end{minipage}}

	\caption{Comparison results of eight methods when overlaying additive random noise with SNR of 0 dB onto the query image (left image). 
	The red lines represent incorrect matches, and the green lines indicate correct matches. 
	Compared to the noise-free scenario, it can be observed that our method is still capable of achieving dense and accurate correspondences. 
	(a) HAPCG. (b) RIFT. (c) LoFTR. (d) MatchFormer. (e) ECO-TR. (f) TopicFM. (g) LightGlue. (h) Ours.}
	\label{Fig.9}
\end{figure*}

\begin{itemize}
	\item{1) HAPCG achieves 90\% and 100\% success rates on datasets 1 and 2, respectively, exhibiting good performance with the minimum RMSE on dataset 2. 
	However, on datasets 3 and 4, success rates drop to 70\% and 80\%. 
	This is attributed to significant scene variations and scale differences among the images. 
	HAPCG performs well in handling scale differences and small rotational variances,  whereas it may not be the optimal solution for handling larger rotational differences.}
	\item{2) RIFT demonstrates excellent adaptability in multisource image matching. 
	In the optical-map dataset, there were three pairs of images with significant differences that caused all methods to fail to match in varying degrees, 
	RIFT still achieves a success rate of 90\% with an RMSE of 3.72 pixels, which is the best of all the results. 
	RIFT has an average SR of 95\% and an average RMSE of 2.80 pixels across the four datasets, 
	but is not competitive on the average NCM due to the low number of matches on datasets 2 and 4, 
	which is attributed to the fact that RIFT does not support scale variance, and therefore performs suboptimal performance in datasets with scale variations.}	
	\item{3) The average success rate of LoFTR+FSC is 95\%, 
	with performance not as good as MatchFormer+FSC but higher than HAPCG and RIFT in terms of NCM.}
	\item{4) MatchFormer+FSC achieved impressive results on the full dataset, with match success rates above 90\%. 
	Among all the methods compared, MatchFormer ranked third in average NCM and second in runtime.}
	\item{5) ECO-TR adopts multiple transformer modules in cascade to complete the coarse-to-fine matching process in a staged mode. 
	ECO-TR can obtain a higher NCM for some images with strong texture features in datasets 1 and 3, which greatly exceeds the other methods; and for datasets 2 and 4, which have some of weak texture images, both SR and NCM are decreased. 
	However, the average performance analysis suggests that ECO-TR has an advantage in NCM among all methods.} 
	\item{6) TopicFM improves the robustness of matching by focusing on the same semantic regions between images. It has SR of 100\% for all datasets, which is the highest success rate among all methods.}
	\item{7) The LightGlue architecture, which augments visual descriptors with context in each layer based on self-attention and cross-attention units with positional encoding. It performs well on datasets 1, 2 and 3 but performs poorly on dataset 4.}	
	\item{8) Our proposed method performs exceptionally well on all datasets, achieving a success rate of 100\% on all datasets except for dataset 3, 
	where the success rate is 80\%. Moreover, the accuracy metrics on datasets 1 and 4 are the highest, 
	attributed to the combined use of dense matching methods and the outlier removal network.}
\end{itemize}

\begin{table*}[hpb]
	\centering
	\caption{THE QUANTITATIVE COMPARISON RESULTS WITH ADD RANDOM NOISE}
	\fontsize{9}{11}\selectfont
	\resizebox{\linewidth}{!}{
		\begin{tabular}{|p{2.0cm}<{\centering}|p{3.0cm}<{\centering}|p{2.1cm}<{\centering}|p{2.1cm}<{\centering}|p{2.1cm}<{\centering}|p{2.1cm}<{\centering}|p{2.1cm}<{\centering}|}
		\hline	
		\multicolumn{2}{|c|}{\diagbox[width=6cm]{\makecell{Data / Method / Metric \\ (NCM / SR / RMSE)}}{\makecell \\ \\ {SNR/dB}}} & 5 & 2 & 0 & -2 & -5 \cr
		\hline	
		\multirow{8}*{\makecell{Dataset\ 1 \\ Optical-Depth}}
		&\textbf{HAPCG}                  & 147 / 80 / 5.51	& 119 / 70 / 7.12	& 92 / 70 / 7.36	& 37 / 40 / 12.74	& 3 / 10 / 18.27    \\
		&\textbf{RIFT}                   & 164 / 90 / 3.10	& 111 / 90 / 3.36	& 83 / 90 / 3.67	& 47 / 90 / 3.89	& 11 / 60 / 9.01    \\
		&\textbf{LoFTR\ +\ FSC}          & 131 / 70 / 7.33	& 92 / 70 / 7.79	& 68 / 50 / 11.32	& 21 / 30 / 14.27	& 7 / 30 / 14.85    \\
		&\textbf{MatchFormer\ +\ FSC}    & 73 / 80 / 5.58	& 55 / 80 / 6.08	& 38 / 60 / 9.48	& 27 / 60 / 9.95	& 8 / 30 / 14.75    \\
		
		&\textbf{ ECO-TR\ +\ FSC}       &  587 / 100 / 2.56   &  370 / 100 / 2.58   &  251 / 100 / 2.79   &  131 / 90 / 4.32  &  29 / 80 / 5.99    \\
		&\textbf{ TopicFM\ +\ FSC}      &  175 / 100 / 2.34   &  132 / 100 / 2.37   &  94 / 90 / 4.10     &  57 / 80 / 5.92   &  20 / 40 / 11.26   \\
		&\textbf{ LightGlue\ +\ FSC}    &  273 / 100 / 2.16   &  182 / 100 / 2.28   &  83 / 100 / 2.47    &  33 / 40 / 13.02  &  12 / 20 / 16.57   \\		
		&\textbf{Ours}                   & 759 / 90 / 4.60	& 615 / 80 / 6.89	& 434 / 70 / 8.44	& 286 / 60 / 10.26	& 109 / 40 / 12.93  \\
		\hline
		\multirow{8}*{\makecell{Dataset\ 2 \\ Infrared-Optical}}
		&\textbf{HAPCG}                  & 183 / 80 / 5.48	& 154 / 70 / 7.30	& 119 / 60 / 9.10	& 107 / 40 / 12.72	& 45 / 30 / 14.51  \\
		&\textbf{RIFT}                   & 90 / 100 / 1.68	& 60 / 80 / 5.38	& 46 / 60 / 8.98	& 30 / 40 / 12.68	& 12 / 30 / 14.56  \\
		&\textbf{LoFTR\ +\ FSC}          & 131 / 70 / 7.33	& 92 / 70 / 7.79	& 68 / 50 / 11.32	& 21 / 30 / 15.27	& 7 / 30 / 15.85   \\
		&\textbf{MatchFormer\ +\ FSC}    & 73 / 80 / 5.58	& 55 / 80 / 6.08	& 38 / 60 / 9.44	& 27 / 60 / 9.86	& 8 / 30 / 15.75   \\
		
		&\textbf{ ECO-TR\ +\ FSC}       &  566 / 70 / 7.80   &  520 / 70 / 7.92    &  395 / 70 / 8.47    &  286 / 50 / 11.24	&  88 / 40 / 13.02  \\
		&\textbf{ TopicFM\ +\ FSC}      &  331 / 100 / 2.28  &  280 / 100 / 2.34   &  231 / 100 / 2.39   &  166 / 100 / 2.50	&  80 / 70 / 7.63   \\
		&\textbf{ LightGlue\ +\ FSC}    &  136 / 100 / 2.21  &  109 / 100 / 2.46   &  84 / 80 / 5.85     &  64 / 60 / 9.31      &  20 / 40 / 12.93  \\		
				
		&\textbf{Ours}                   & 415 / 100 / 2.89	& 297 / 90 / 4.75	& 253 / 80 / 6.02	& 201 / 70 / 7.81	& 163 / 60 / 9.98  \\
		\hline
		\multirow{8}*{\makecell{Dataset\ 3 \\ Optical-Map}}
		&\textbf{HAPCG}                  & 159 / 60 / 9.06	& 100 / 30 / 14.31	& 74 / 30 / 14.85	& 9 / 10 / 18.18	& 4 / 10 / 18.26   \\
		&\textbf{RIFT}                   & 79 / 80 / 5.48	& 38 / 70 / 7.29	& 20 / 60 / 9.10	& 10 / 40 / 12.66	& 1 / 10 / 18.17   \\
		&\textbf{LoFTR\ +\ FSC}          & 208 / 70 / 7.14	& 175 / 70 / 7.65	& 139 / 70 / 7.89	& 111 / 60 / 9.36	& 76 / 40 / 13.44  \\
		&\textbf{MatchFormer\ +\ FSC}    & 192 / 70 / 7.12	& 155 / 70 / 7.34	& 114 / 70 / 7.53	& 89 / 70 / 7.99	& 45 / 40 / 12.25  \\
		
		&\textbf{ ECO-TR\ +\ FSC}       &  787 / 100 / 3.04   &  572 / 90 / 4.33   &  418 / 80 / 5.98   &  267 / 80 / 6.15  &  97 / 60 / 9.47   \\
		&\textbf{ TopicFM\ +\ FSC}      &  390 / 90 / 3.68    &  333 / 80 / 4.34   &  298 / 80 / 5.65   &  181 / 80 / 5.95  &  56 / 70 / 7.67   \\
		&\textbf{ LightGlue\ +\ FSC}    &  251 / 90 / 3.82    &  173 / 90 / 3.87   &  102 / 80 / 5.64   &  57 / 80 / 5.86   &  19 / 40 / 12.96  \\ 		

		&\textbf{Ours}                   & 656 / 80 / 6.11	& 537 / 70 / 7.98	& 443 / 70 / 8.38	& 366 / 60 / 9.70	& 270 / 50 / 12.03 \\
		\hline
		\multirow{8}*{\makecell{Dataset\ 4 \\ SAR-Optical}}
		&\textbf{HAPCG}                  & 73 / 60 / 9.12	& 47 / 40 / 12.76	& 35 / 30 / 14.28	& 19 / 30 / 14.66	& 8 / 20 / 16.34   \\
		&\textbf{RIFT}                   & 119 / 90 / 3.68	& 86 / 80 / 5.36	& 75 / 80 / 5.43	& 55 / 70 / 7.30	& 19 / 50 / 10.94  \\
		&\textbf{LoFTR\ +\ FSC}          & 102 / 90 / 4.22	& 42 / 80 / 5.92	& 22 / 70 / 7.63	& 12 / 50 / 11.29	& 6 / 40 / 13.08   \\
		&\textbf{MatchFormer\ +\ FSC}    & 288 / 90 / 4.45	& 226 / 80 / 5.86	& 156 / 80 / 6.02	& 101 / 80 / 6.33	& 36 / 40 / 13.31  \\

		&\textbf{ ECO-TR\ +\ FSC}       &  87 / 70 / 7.80   &  55 / 60 / 9.72	&  33 / 40 / 13.04  &  18 / 20 / 16.51   &  6 / 10 / 18.27   \\
		&\textbf{ TopicFM\ +\ FSC}      &  98 / 90 / 4.21   &  64 / 90 / 4.32	&  47 / 70 / 7.72   &  31 / 60 / 9.49    &  14 / 40 / 12.96  \\	
		&\textbf{ LightGlue\ +\ FSC}    &  106 / 80 / 6.13  &  71 / 80 / 6.15	&  31 / 60 / 9.54   &  21 / 50 / 11.31   &  12 / 30 / 14.82  \\
				
		&\textbf{Ours}                   & 561 / 90 / 4.19	& 468 / 90 / 4.83	& 349 / 80 / 5.86	& 246 / 70 / 8.33	& 129 / 60 / 9.97   \\
		\hline
		\end{tabular}
	}
	\label{TABLE II}
\end{table*}

As well, Table \ref{TABLE I} records the average run time of the eight matching methods in multisource remote sensing images. 
HAPCG and RIFT can handle most image matching tasks, but both of them have high matching time costs, which seriously affects the matching efficiency. 
LoFTR has a lower time cost than HAPCG and RIFT, its average RMSE value is relatively large. 
MatchFormer has the fastest execution time while is in the middle of the range in terms of average NCM, and is not optimal in terms of time efficiency. 
ECO-TR is a dense matching method, and its number of correct matches is very large, which is proportional to its RT; the larger its NCM, the higher the computation time RT. 
TopicFM has the fastest runtime of all methods due to its coarse-to-fine framework and the construction of lightweight networks for each stage. 
LightGlue continues the design idea of SuperGlue by detecting and describing key points throughout the image; this pipeline procedure reduces its running time. 
Our method is second only to TopicFM and MatchFormer in terms of run time, and is 2.1 times better than TopicFM and 1.8 times better than MatchFormer in terms of NCM. 
Generally, our method exhibits the best comprehensive time efficiency.

\subsection{Noise Image Matching Experiment}

The design of the noise image matching experiment aims to verify the robustness of the matching methods without considering noise preprocessing. 
In this experiment, the reference image (right) is fixed, and noise intensity is continuously increased to the query image (left) to assess the robustness of matching methods. 
We conducted experiments for two cases: additive random noise and periodic stripe noise. 
Additionally, the average correct rate (ACR) is used as the evaluation metric to evaluate the robustness of all methods when images are disturbed by noise. 
The expression is as follows:
\begin{equation}
	\label{deqn_ex24}
	\text{ACR}=\frac{\text{NCM}_\text{noise}} {\text{NCM}}
\end{equation}	
where ${\text{NCM}_\text{noise}}$ represents the average correct matching number in the presence of noise interference, 
and $\text{NCM}$ is the average correct matching number without noise. 
Ultimately, we use a curve graph to illustrate the robustness of all methods as noise intensity increases.
\begin{figure*}[htbp]
	\centering
	\begin{minipage}{1\linewidth}
		\centering
		\includegraphics[width=0.245\linewidth]{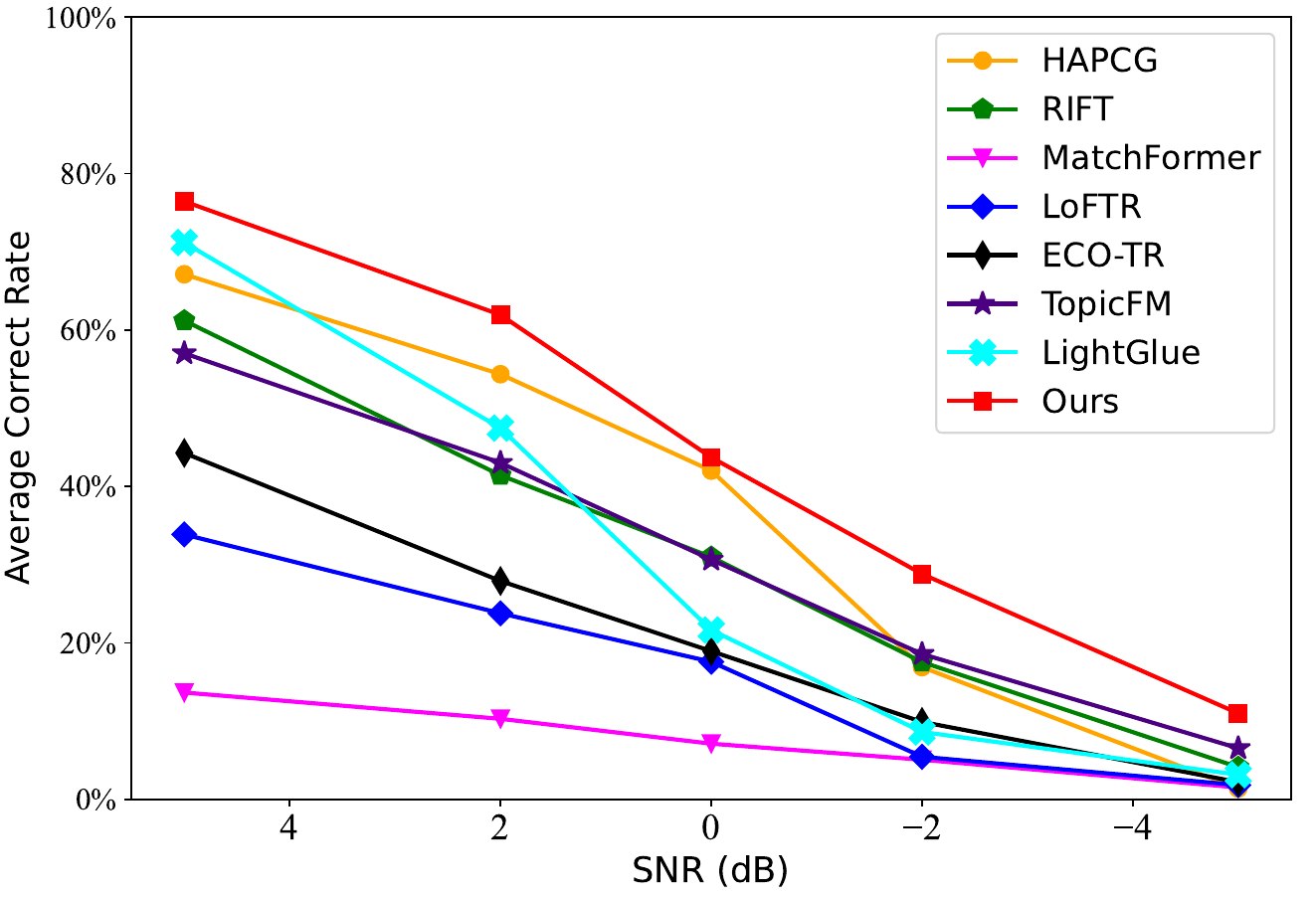}
		\label{fig10a}
		\includegraphics[width=0.245\linewidth]{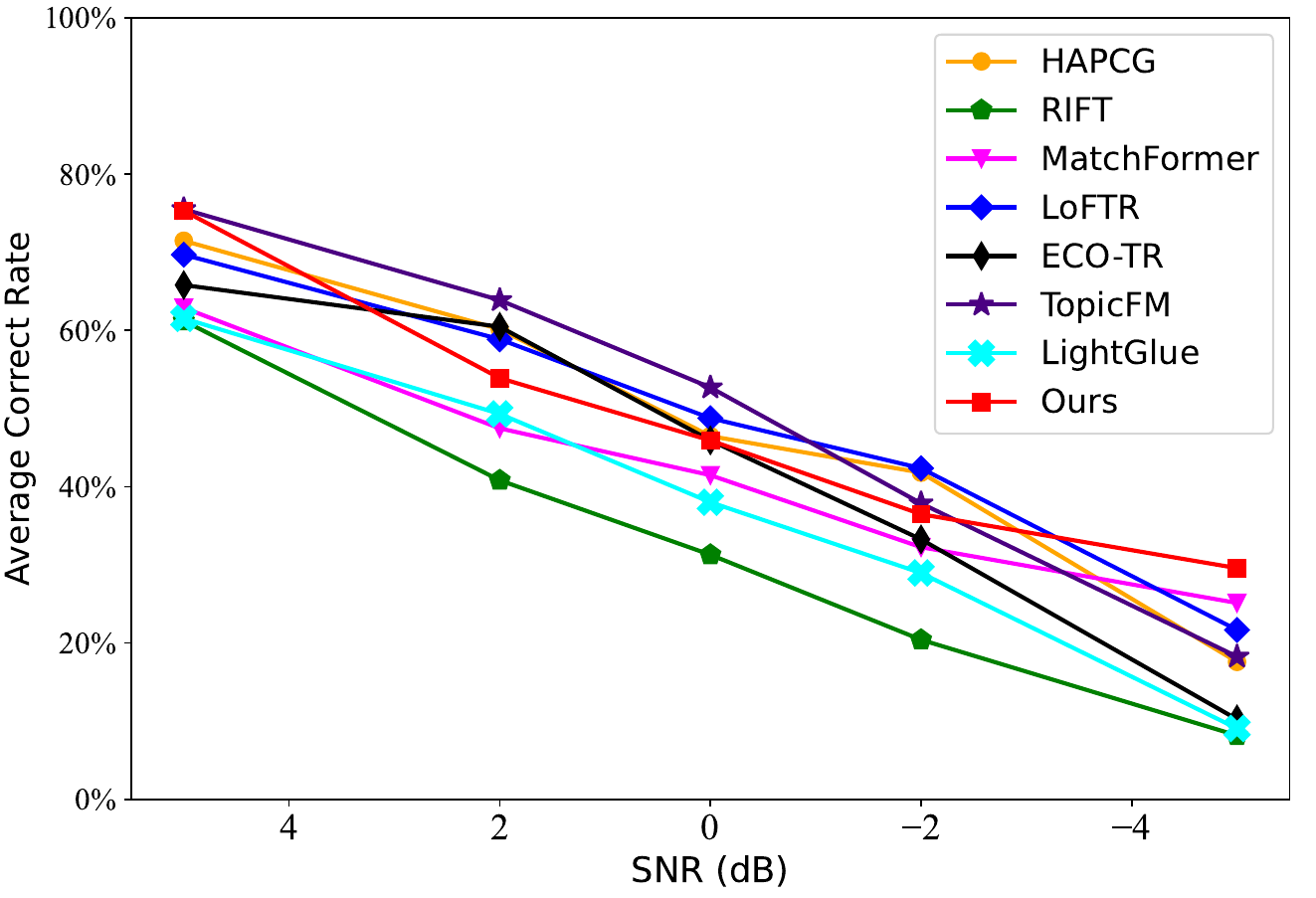}
		\label{fig10b}
		\includegraphics[width=0.245\linewidth]{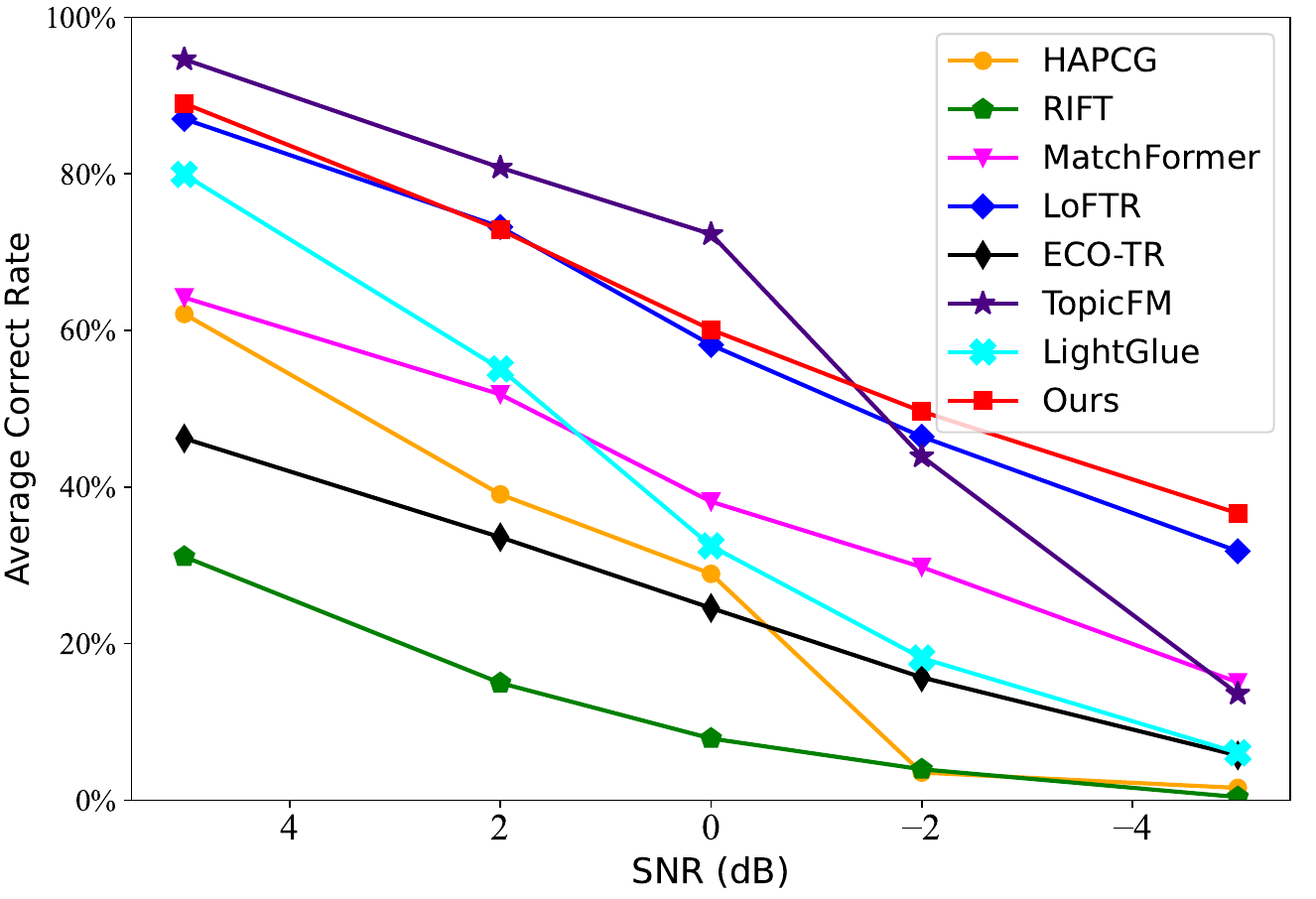}
		\label{fig10c}
		\includegraphics[width=0.245\linewidth]{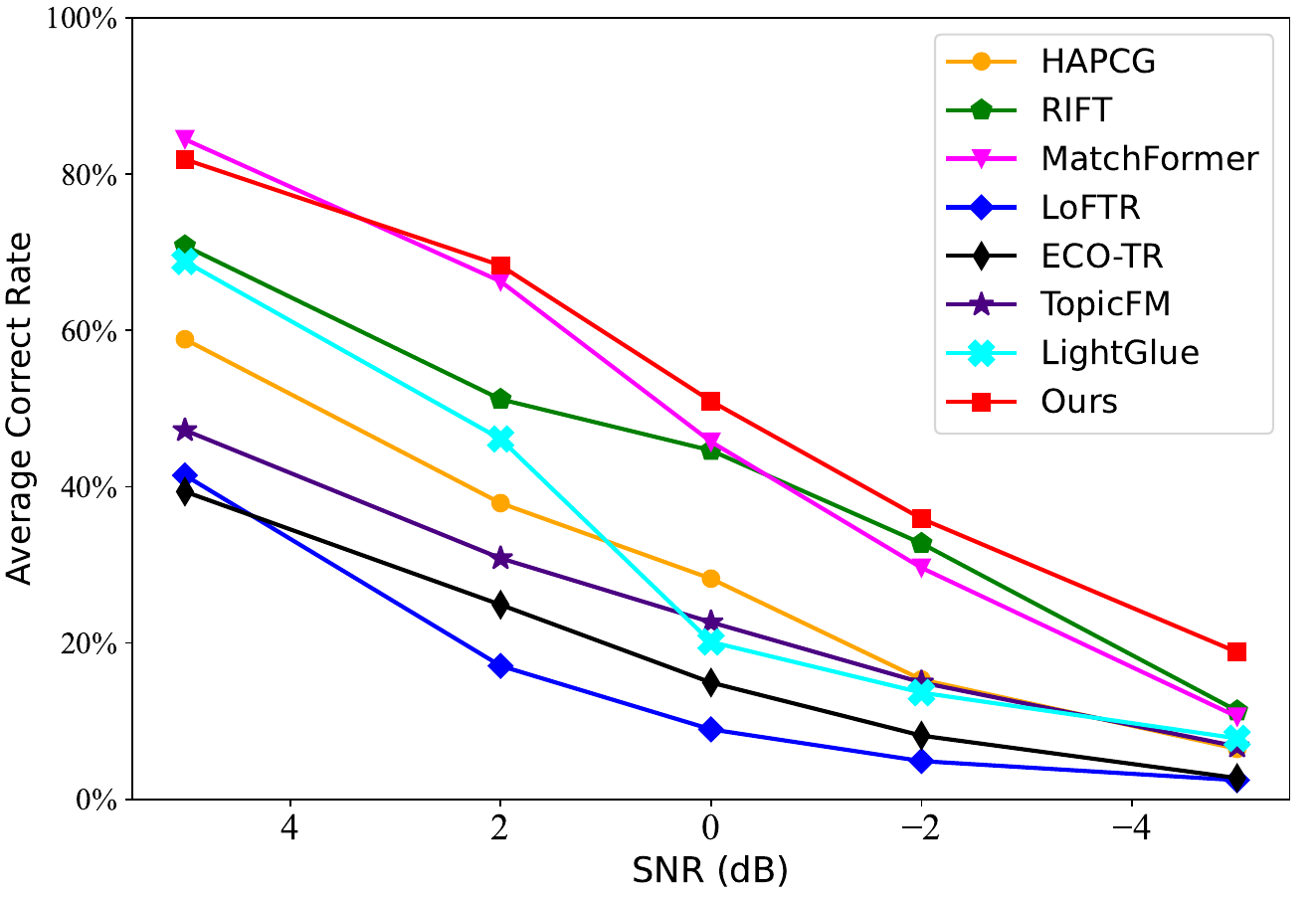}
		\label{fig10d}
		% \caption{histogram}
	\end{minipage}
	\caption{The average correct ratio of eight methods is compared as the intensity of additive random noise increases. From left to right, the datasets are optical-depth, infrared-optical, optical-map, and SAR-optical.}
	\label{Fig.10}
\end{figure*}

\textbf{\emph{1)} Additive Random Noise:} The causes of additive random noise are typically attributed to the following factors: 
(i) the image sensor works under low-light conditions, 
(ii) the sensor circuitry generates thermal noise and electromagnetic interference, 
and (iii) the image sensor operates for a long period at a high temperature.
\begin{equation}
	\label{deqn_ex25}
	J_a=I+n
\end{equation}	
where $I$ is the original image, $J_a$ is the additive noise image, 
and $n$ is Gaussian random noise with mean $u$ and variance $\sigma^2$. 
Its probability density function is expressed as:
\begin{equation}
	\label{deqn_ex26}
	\text{p}( {z} )=\frac{1}{\sqrt{2\pi\ }{\sigma}} \exp{\left( -\frac{{{\left( {z-{\mu}} \right)}^{2}}} {2 {\sigma}^{2}}\ \right)}
	% \text{p}\left(\ z\ \right)=\frac{1}{\sqrt{2\pi\ }\sigma\ }\exp\ \left(\ -\frac{{{\left(\ z-\mu\ \ \right)}^{2}}}{2{{\sigma\ }^{2}}}\ \right)
	% {G_s}(\alpha ) = {G_{\max }} - 3{\left( {\frac{\alpha }{{{\alpha _{3{\text{dB}}}}}}} \right)^2}
\end{equation}	
Signal-to-noise ratio (SNR) is defined as:
\begin{equation}
	\label{deqn_ex27}
	\text{SNR}=20\log{\left( {E\left( {I} \right)}/{\sigma^2} \right)} 
\end{equation}	
where $E\left( {I} \right)$ is the average image power, $\sigma^2$ is the noise power.	

Results Analysis and Comparison: Fig. \ref{Fig.9} presents a qualitative comparison of eight methods for additive random noise images when SNR is 0 dB. 
The results indicate that all methods have different degrees of performance decline due to the fact that noise interferes with the recognition and feature extraction on feature points. 
HAPCG and RIFT match failures on the first pair of images, RIFT only achieving an NCM of 16 on the third pair of images. 
The matching results of LoFTR and MatchFormer are similar, with a significant decrease in average NCM. 
ECO-TR fails to detect feature points on the second pair of images with significant scene changes. 
TopicFM only accomplishes sparse matching results on all images. 
LightGlue results are similar to LoFTR with moderate performance.
Compared to others, our method still achieves dense matching. 
Table \ref{TABLE II} summarizes the quantitative results of eight methods under different intensities of additive noise, and Fig. \ref{Fig.10} plots the curve of the average correct rate. 
(i) On Dataset 1, the ACR of MatchFormer is only 13.64\% and LoFTR is 33.85\% when SNR is 5dB; 
although the average NCM of RIFT decreases, its SR is 90\% and maintains stability as SNR continues to decrease. Even when SNR drops to -5dB, the SR remains at 60\%. 
Meanwhile, ECO-TR has an SR of 80\%, which is the best indicator among all methods, but the NCM is low. Our method follows ECO-TR with an SR of 40\%, but the NCM is higher than ECO-TR.
(ii) On dataset 2, TopicFM achieves the best metrics with an SR of 70\% when SNR is -5dB. 
Our method ranks second with an SR of 60\%; while all other algorithms are below 40\%.
(iii) On datasets 3 and 4, the SR of our method is more than 50\%, which is still the best, and the average correct rate is more than 5\% higher than others. 
The above comparison shows that our method can maintain robustness in noise image matching and improve the reliability and accuracy of remote sensing image matching tasks.

\begin{figure*}[hbp]
	\centering
	\hspace{-0.1cm}\subfigure[]{  % Optical-Depth
	\begin{minipage}[b]{0.24\textwidth}
	\includegraphics[width=1\linewidth,height=17mm]{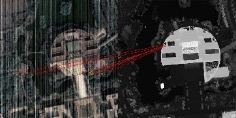}\vspace{1pt}
	\includegraphics[width=1\linewidth,height=17mm]{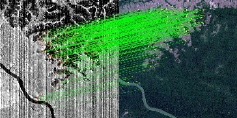}\vspace{1pt}
	\includegraphics[width=1\linewidth,height=17mm]{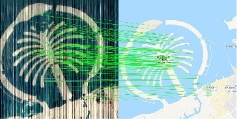}\vspace{1pt}
	\includegraphics[width=1\linewidth,height=17mm]{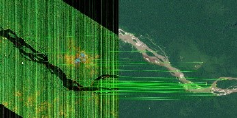}\vspace{1pt}
	\end{minipage}}
	\hspace{-0.1cm}\subfigure[]{  %Infrared-Optical
	\begin{minipage}[b]{0.24\textwidth}
	\includegraphics[width=1\linewidth,height=17mm]{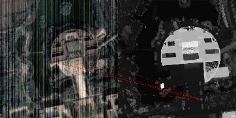}\vspace{1pt}
	\includegraphics[width=1\linewidth,height=17mm]{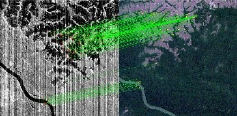}\vspace{1pt}
	\includegraphics[width=1\linewidth,height=17mm]{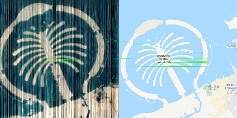}\vspace{1pt}
	\includegraphics[width=1\linewidth,height=17mm]{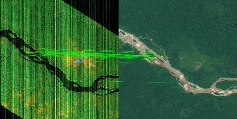}\vspace{1pt}
	\end{minipage}}
	\hspace{-0.1cm}\subfigure[]{  % Optical-Map
	\begin{minipage}[b]{0.24\textwidth}
	\includegraphics[width=1\linewidth,height=17mm]{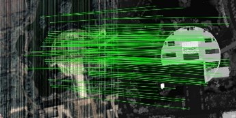}\vspace{1pt}
	\includegraphics[width=1\linewidth,height=17mm]{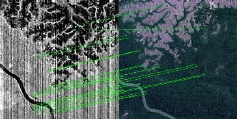}\vspace{1pt}
	\includegraphics[width=1\linewidth,height=17mm]{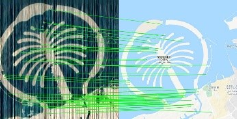}\vspace{1pt}
	\includegraphics[width=1\linewidth,height=17mm]{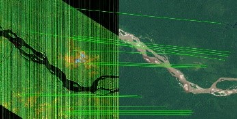}\vspace{1pt}
	\end{minipage}}
	\hspace{-0.1cm}\subfigure[]{  % SAR-Optical
	\begin{minipage}[b]{0.24\textwidth}
	\includegraphics[width=1\linewidth,height=17mm]{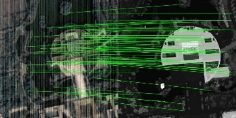}\vspace{1pt}
	\includegraphics[width=1\linewidth,height=17mm]{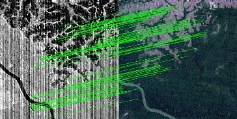}\vspace{1pt}
	\includegraphics[width=1\linewidth,height=17mm]{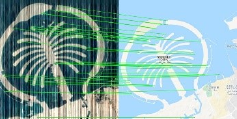}\vspace{1pt}
	\includegraphics[width=1\linewidth,height=17mm]{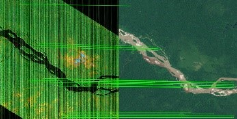}\vspace{1pt}
	\end{minipage}}
	\\
	\hspace{-0.1cm}\subfigure[]{  % SAR-Optical
	\begin{minipage}[b]{0.24\textwidth}
	\includegraphics[width=1\linewidth,height=17mm]{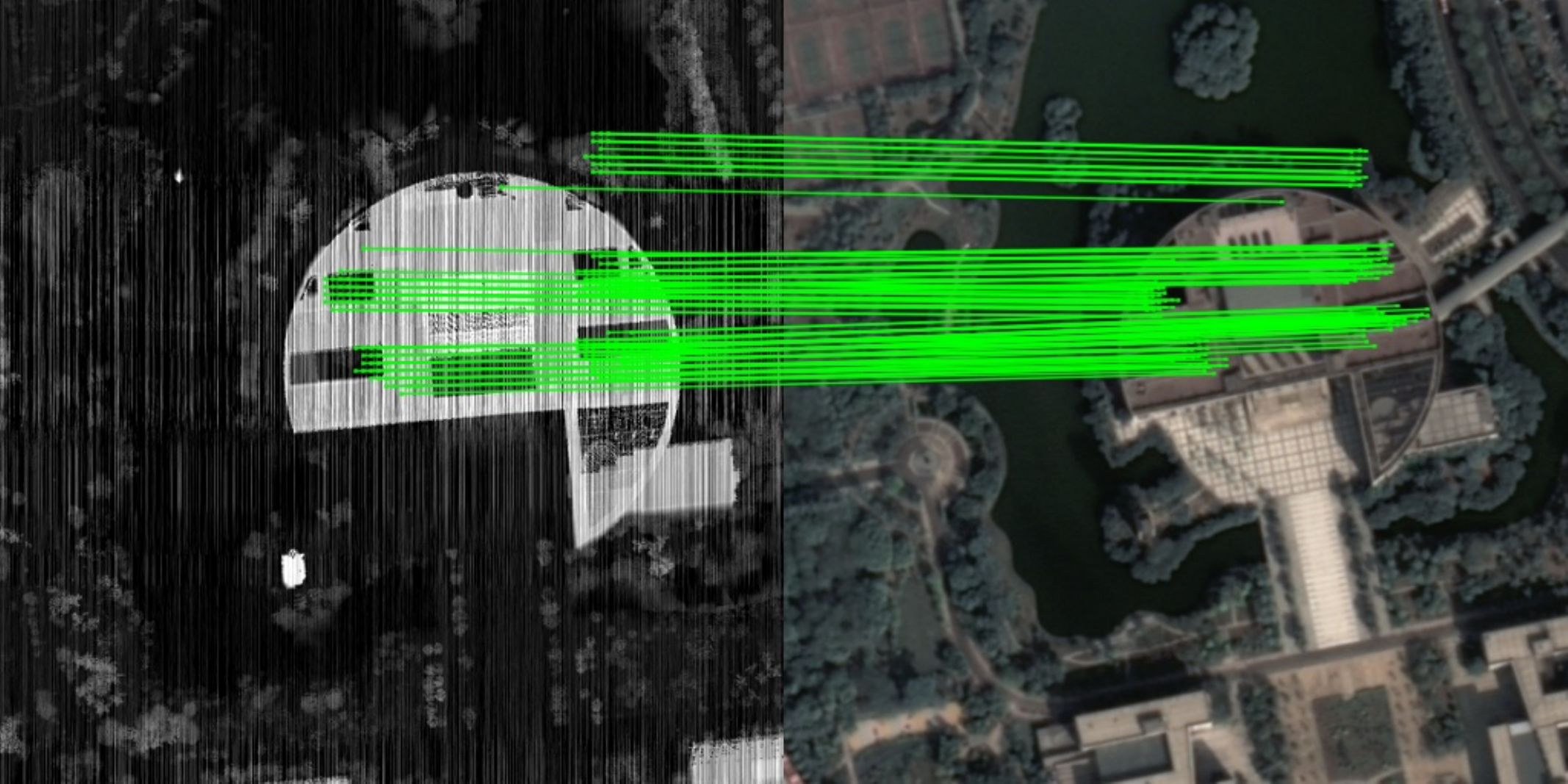}\vspace{1pt}
	\includegraphics[width=1\linewidth,height=17mm]{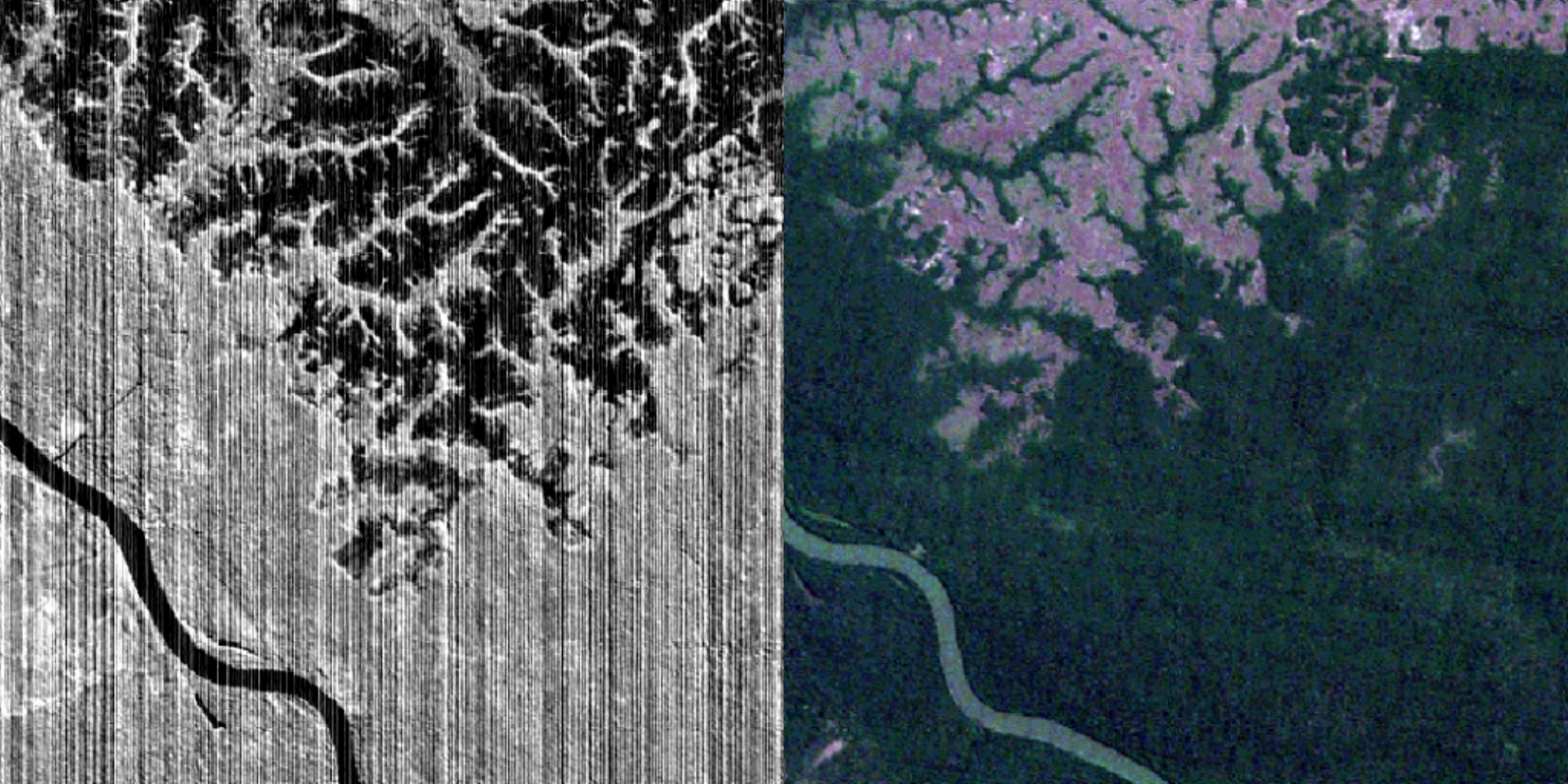}\vspace{1pt}
	\includegraphics[width=1\linewidth,height=17mm]{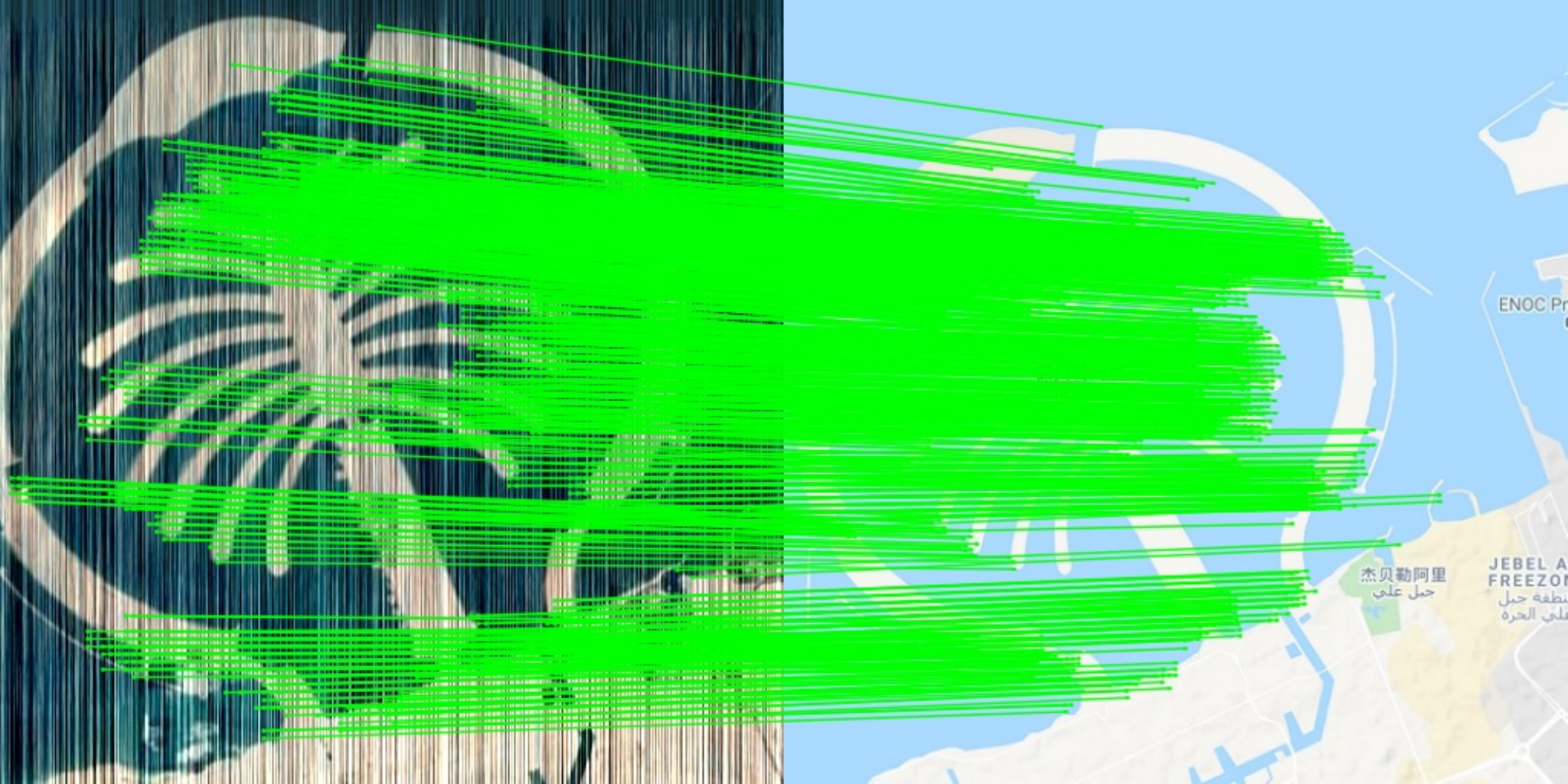}\vspace{1pt}
	\includegraphics[width=1\linewidth,height=17mm]{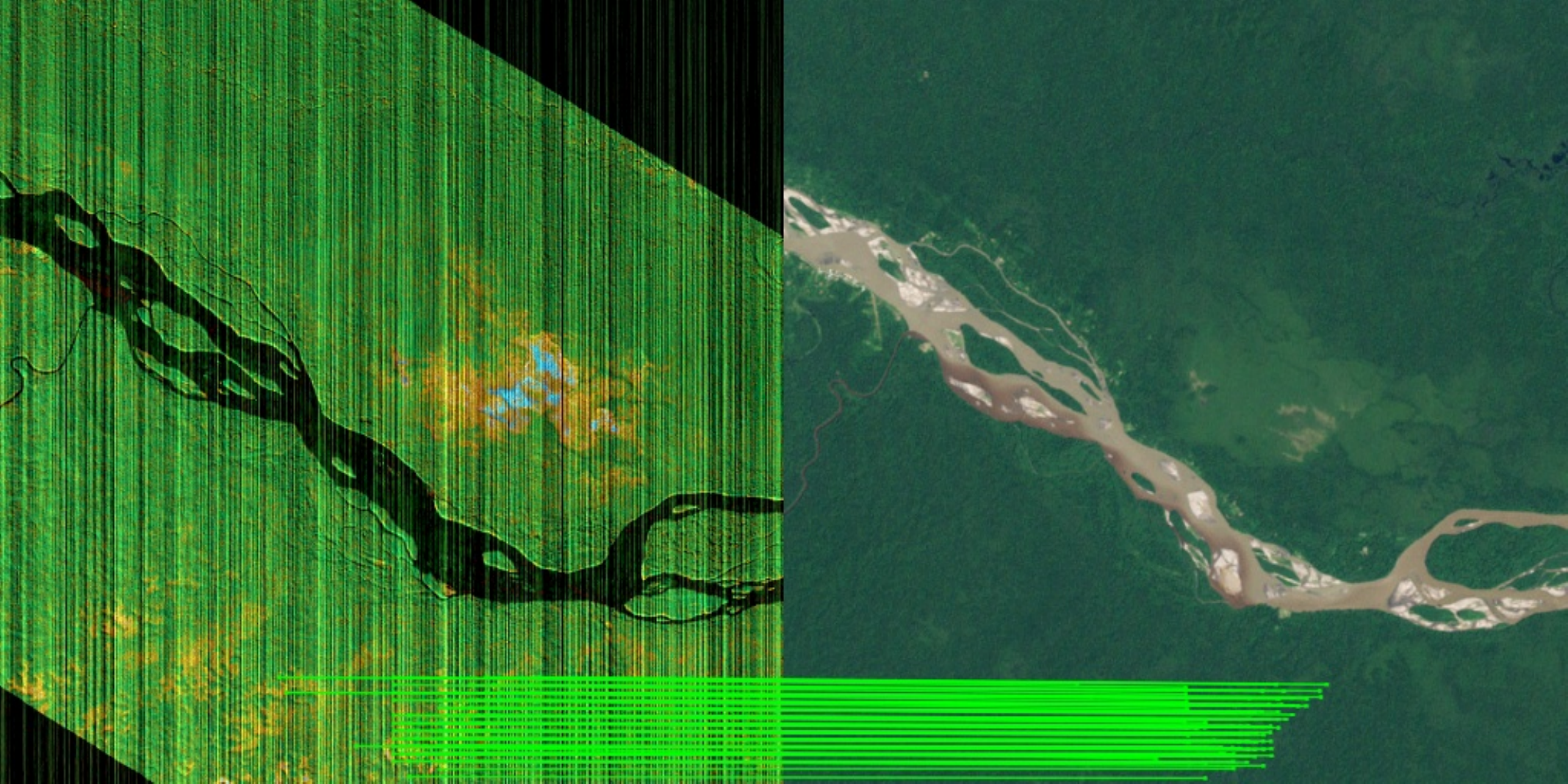}\vspace{1pt}
	\end{minipage}}
	\hspace{-0.1cm}\subfigure[]{  %Infrared-Optical
	\begin{minipage}[b]{0.24\textwidth}
	\includegraphics[width=1\linewidth,height=17mm]{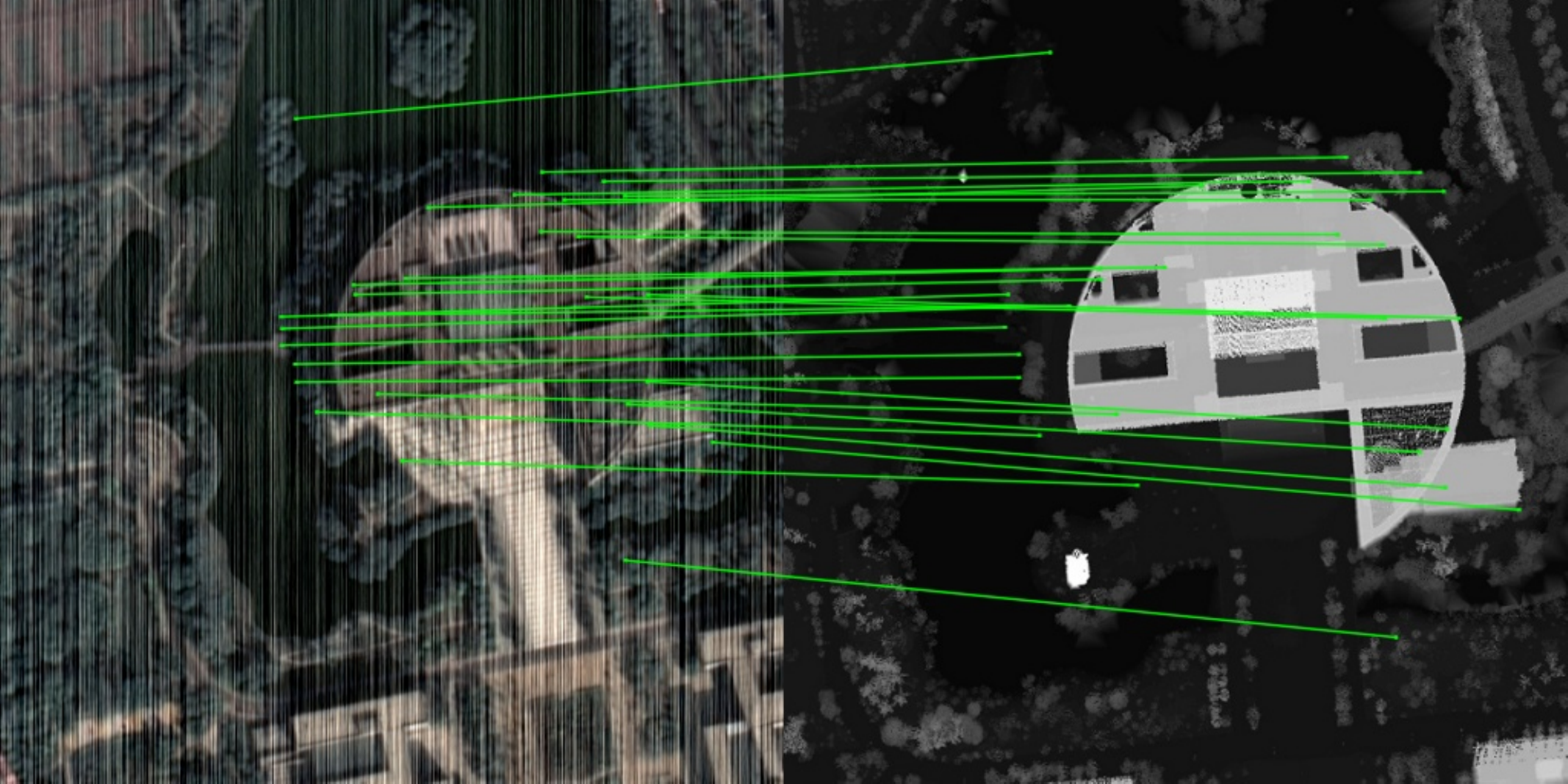}\vspace{1pt}
	\includegraphics[width=1\linewidth,height=17mm]{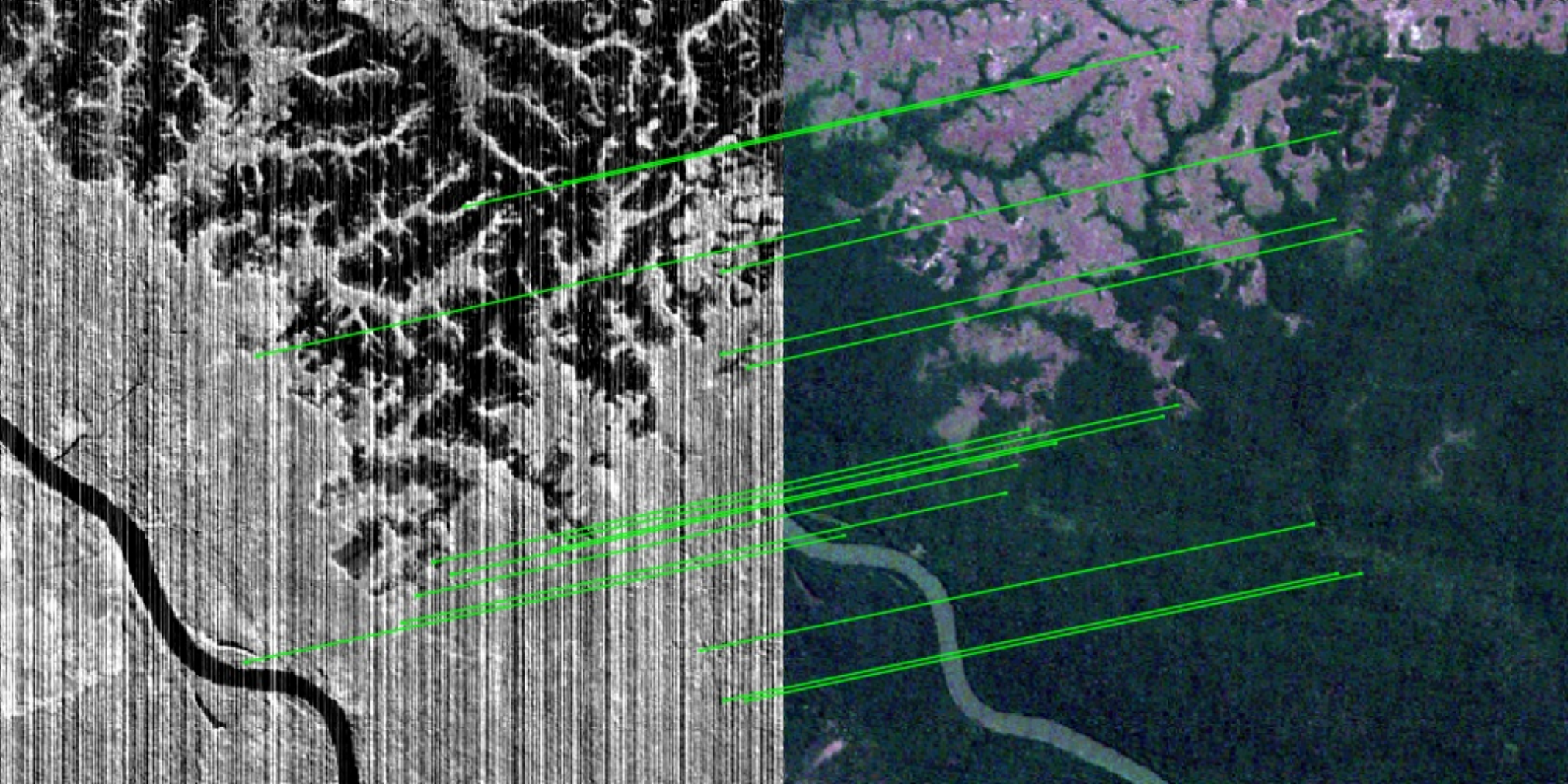}\vspace{1pt}
	\includegraphics[width=1\linewidth,height=17mm]{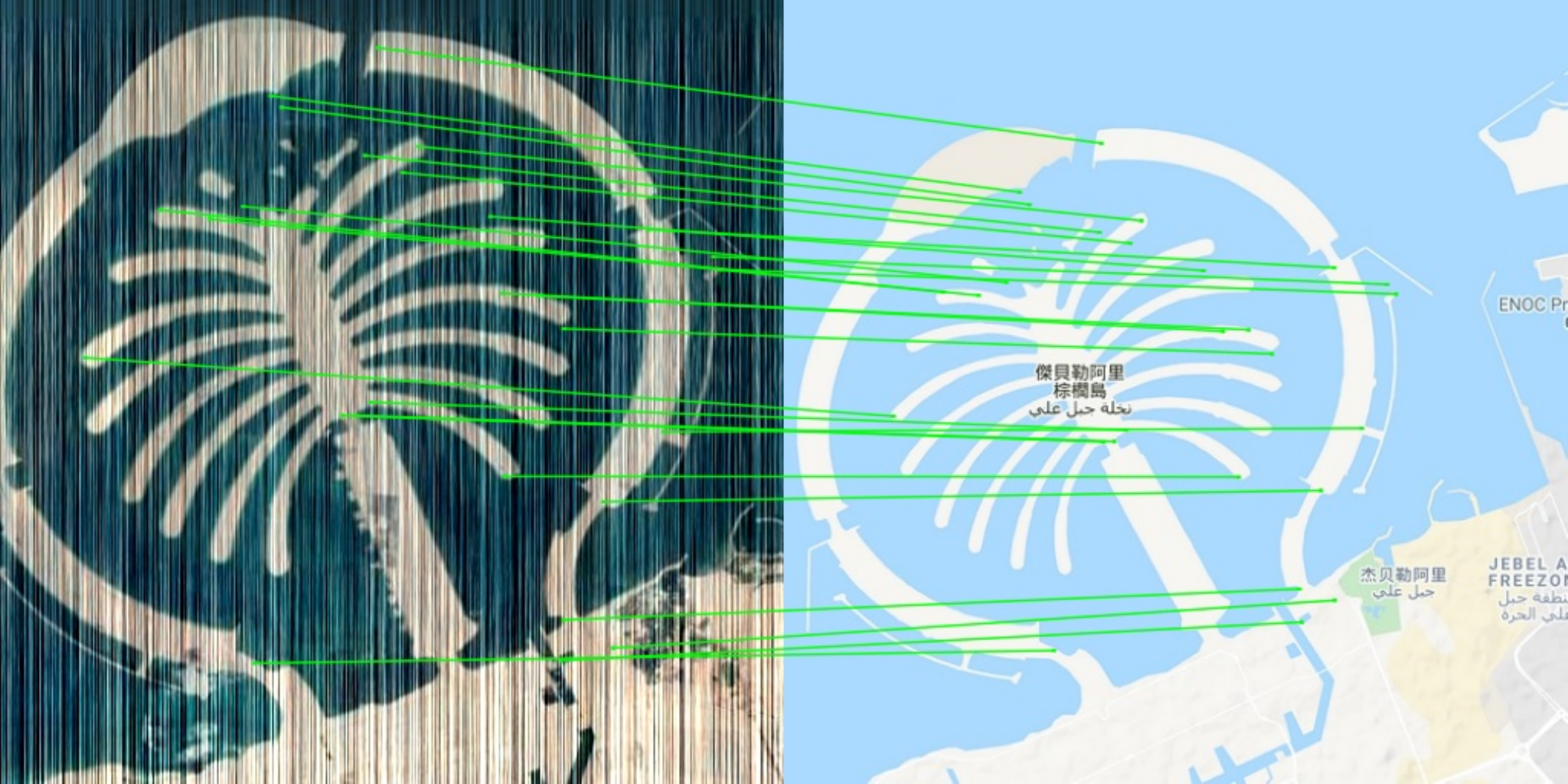}\vspace{1pt}
	\includegraphics[width=1\linewidth,height=17mm]{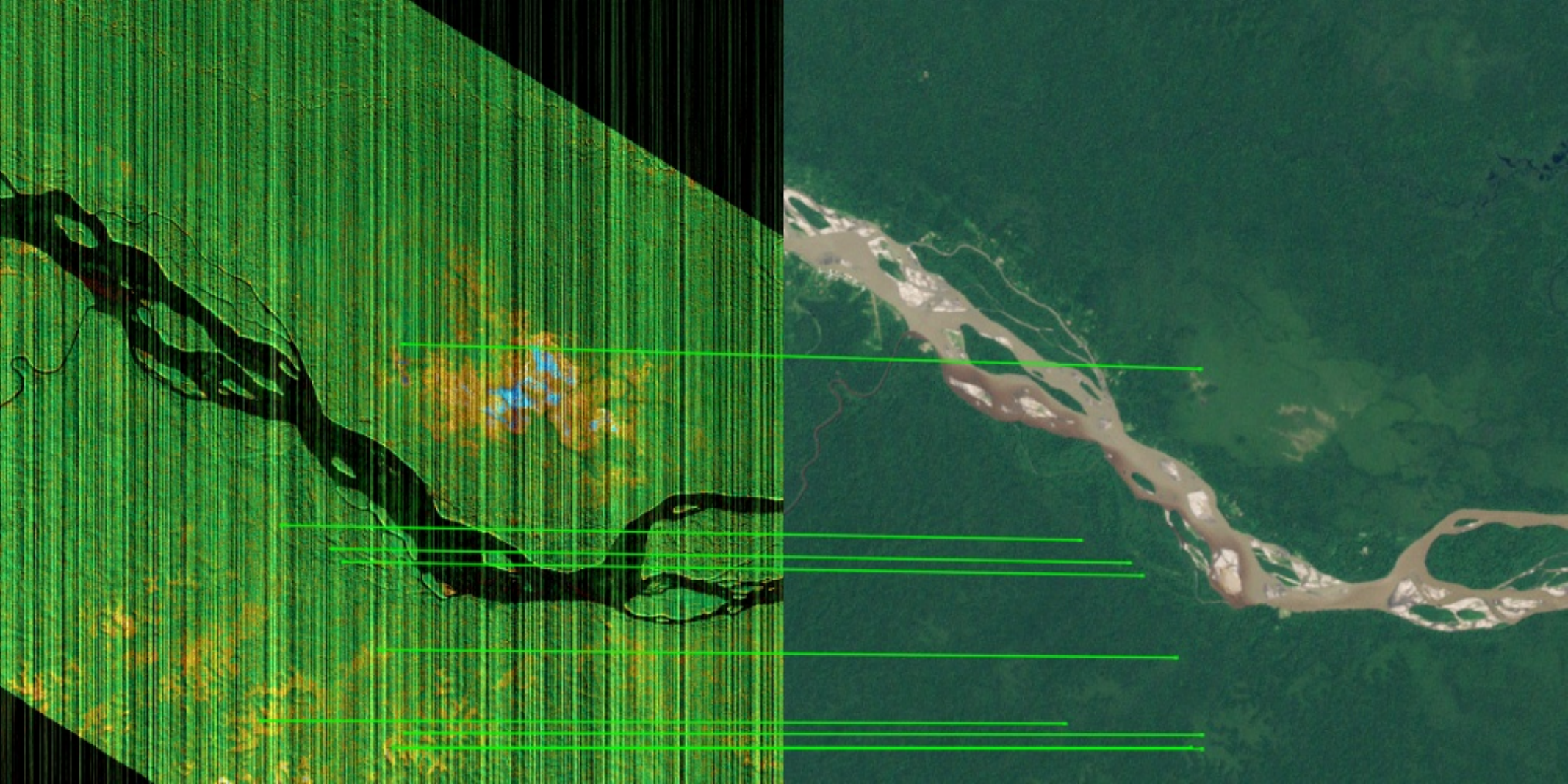}\vspace{1pt}
	\end{minipage}}
	\hspace{-0.1cm}\subfigure[]{  % Optical-Map
	\begin{minipage}[b]{0.24\textwidth}
	\includegraphics[width=1\linewidth,height=17mm]{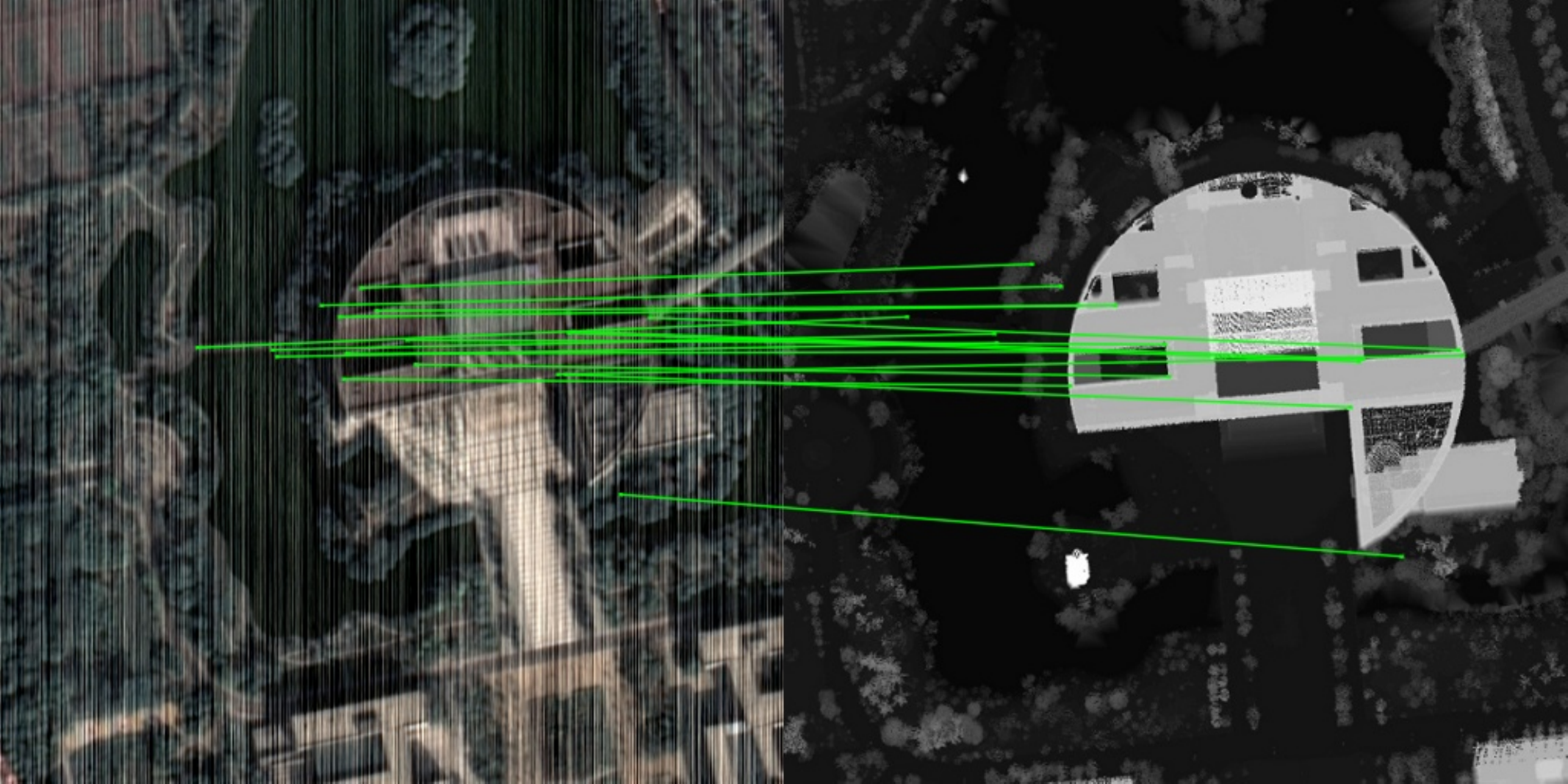}\vspace{1pt}
	\includegraphics[width=1\linewidth,height=17mm]{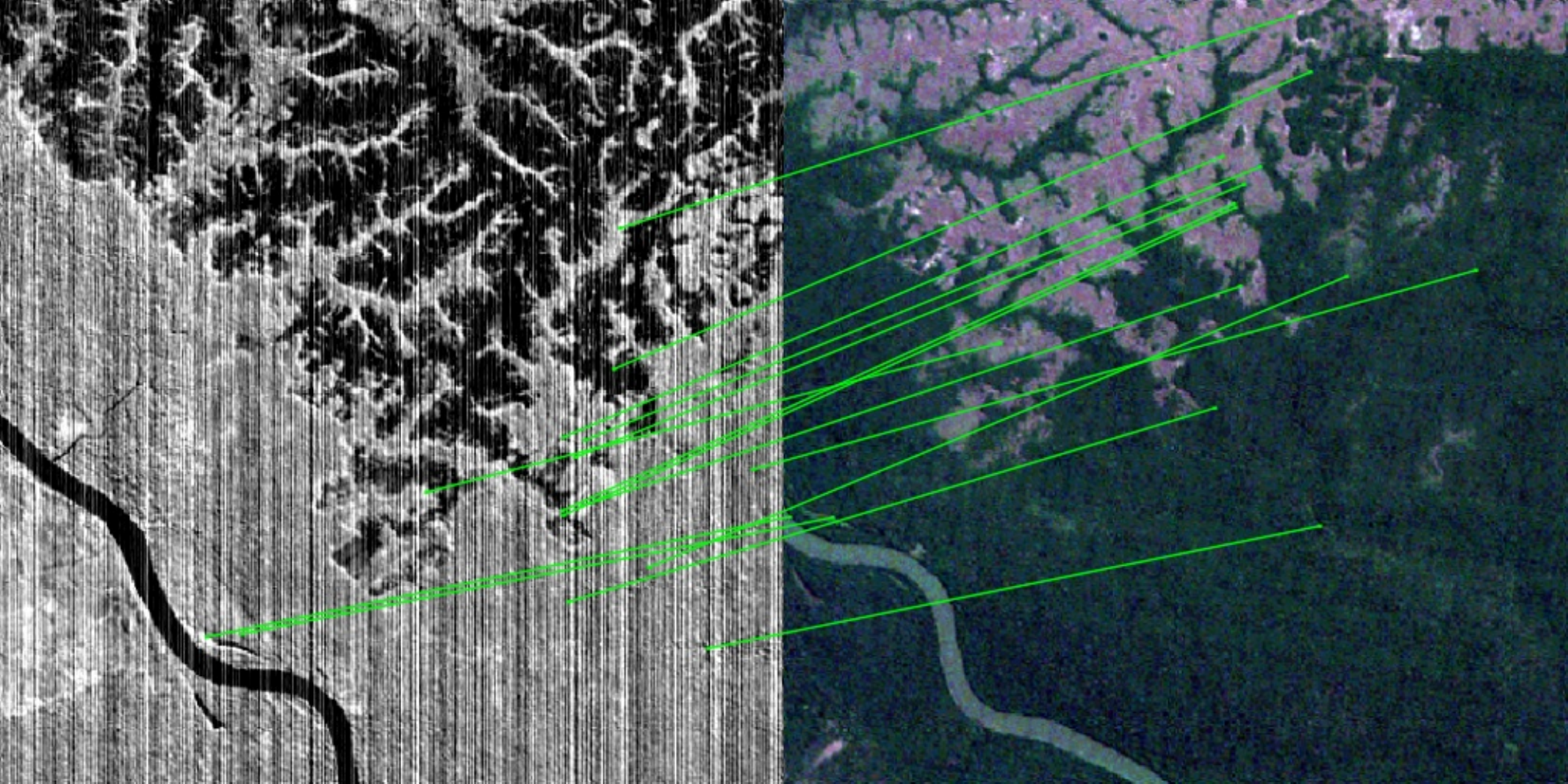}\vspace{1pt}
	\includegraphics[width=1\linewidth,height=17mm]{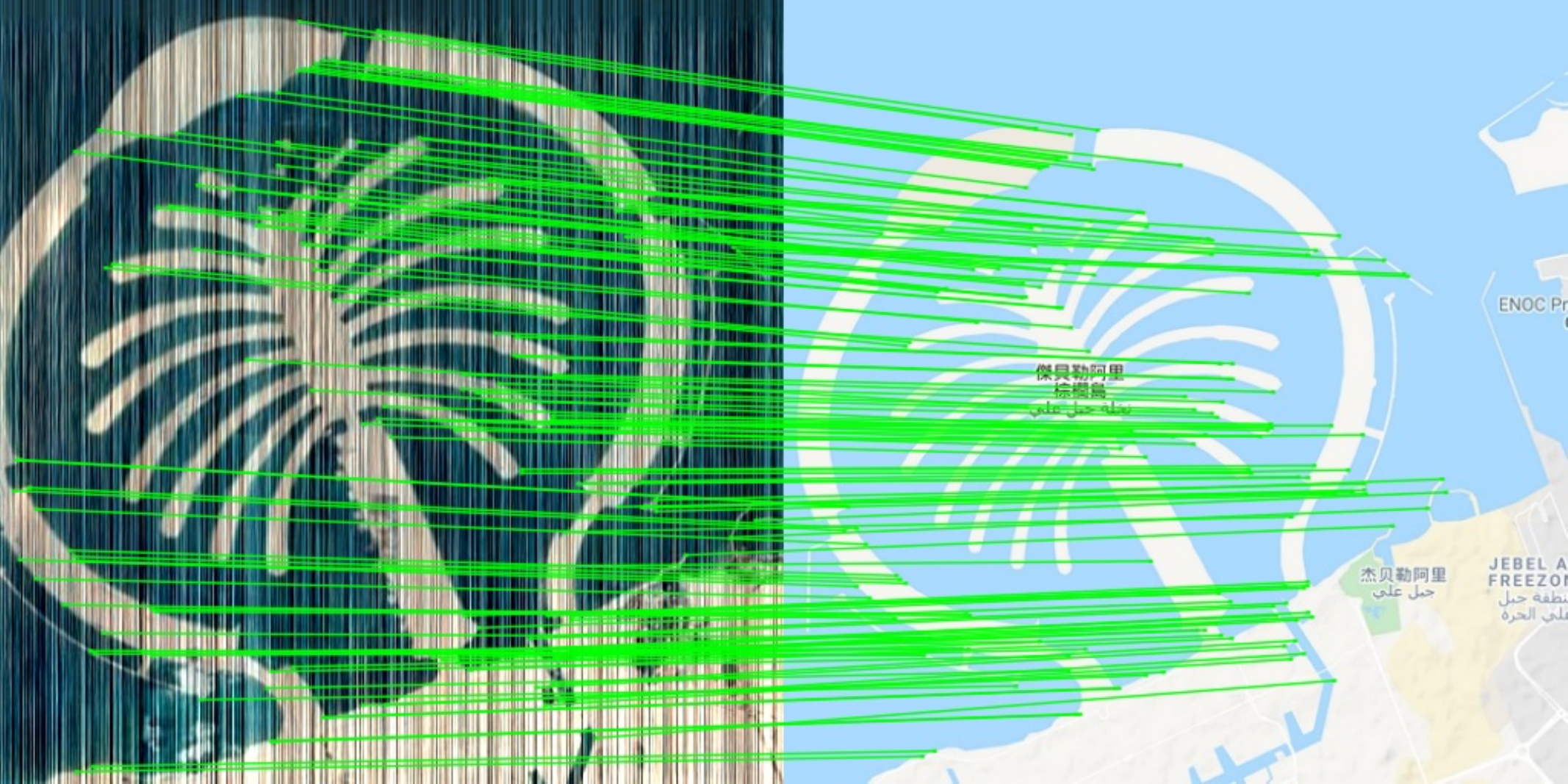}\vspace{1pt}
	\includegraphics[width=1\linewidth,height=17mm]{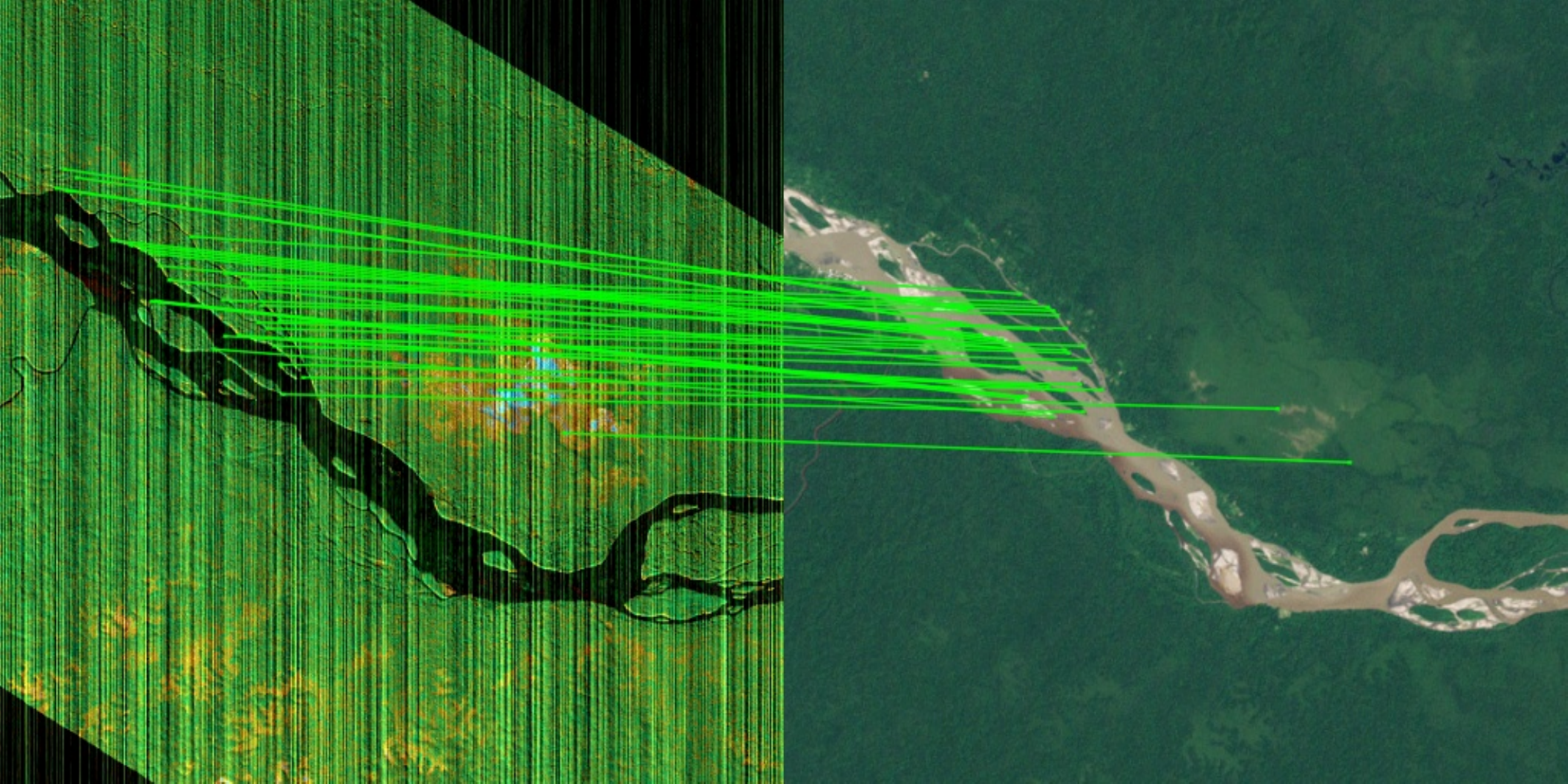}\vspace{1pt}
	\end{minipage}}
	\hspace{-0.1cm}\subfigure[]{  % SAR-Optical
	\begin{minipage}[b]{0.24\textwidth}
	\includegraphics[width=1\linewidth,height=17mm]{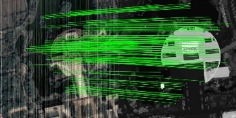}\vspace{1pt}
	\includegraphics[width=1\linewidth,height=17mm]{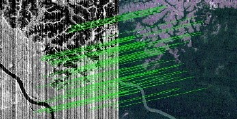}\vspace{1pt}
	\includegraphics[width=1\linewidth,height=17mm]{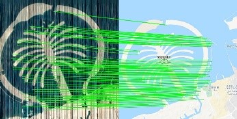}\vspace{1pt}
	\includegraphics[width=1\linewidth,height=17mm]{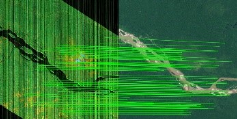}\vspace{1pt}
	\end{minipage}}

	\caption{Comparison results of eight methods when overlaying stripe noise with $\sigma^{2}=0.1 $ onto the query image (left image). 
	The red lines represent incorrect matches, and the green lines indicate correct matches. Compared to the additive random noise, stripe noise is more challenging for correct matching. 
	(a) HAPCG. (b) RIFT. (c) LoFTR. (d) MatchFormer. (e) ECO-TR. (f) TopicFM. (g) LightGlue. (h) Ours.}
	\label{Fig.11}
\end{figure*}

\textbf{\emph{2)} Periodic Stripe Noise:} Stripe noise is a specific remote sensing image noise that exhibits periodicity, directionality, and a striped distribution. 
This noise is caused by scanning differences between the forward and backward, mechanical movements of the satellite sensor, and temperature variations during the repetitive scanning process. 
The generation of stripe noise images is as follows:
\begin{equation}
	\label{deqn_ex28}
	J_s=I+n*I 
\end{equation}	
where $I$ is the original image, $J_s$ is the stripe noise image, 
and $n$ is the uniformly distributed noise with mean $u$ and variance $\sigma^2$, $\ast$ represent three-dimensional tensor multiplication.

\begin{figure*}[htpb]
	\centering
	\begin{minipage}{1\linewidth}
		\centering
		\includegraphics[width=0.245\linewidth]{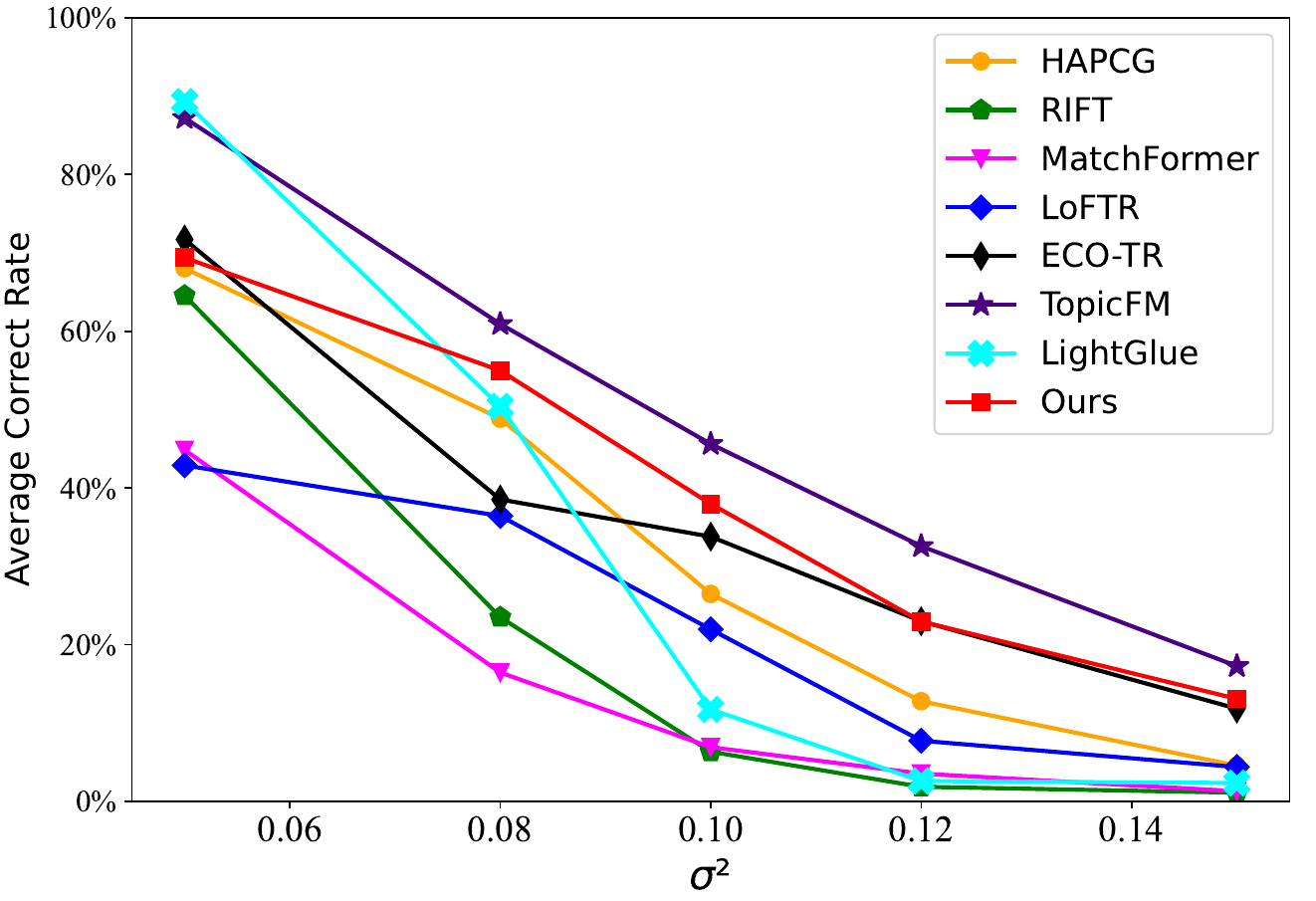}
		\label{fig12a}
		\includegraphics[width=0.245\linewidth]{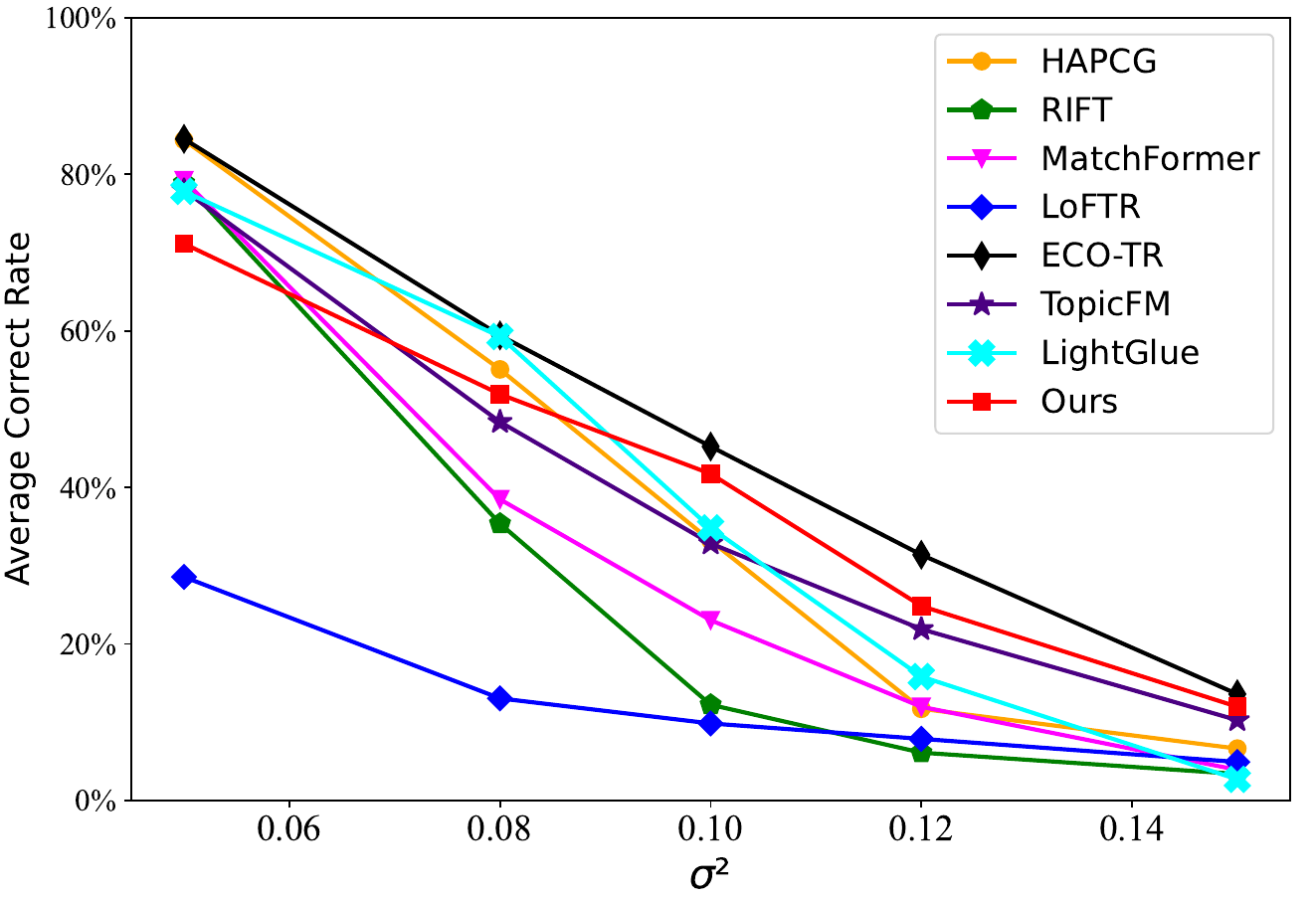}
		\label{fig12b}
		\includegraphics[width=0.245\linewidth]{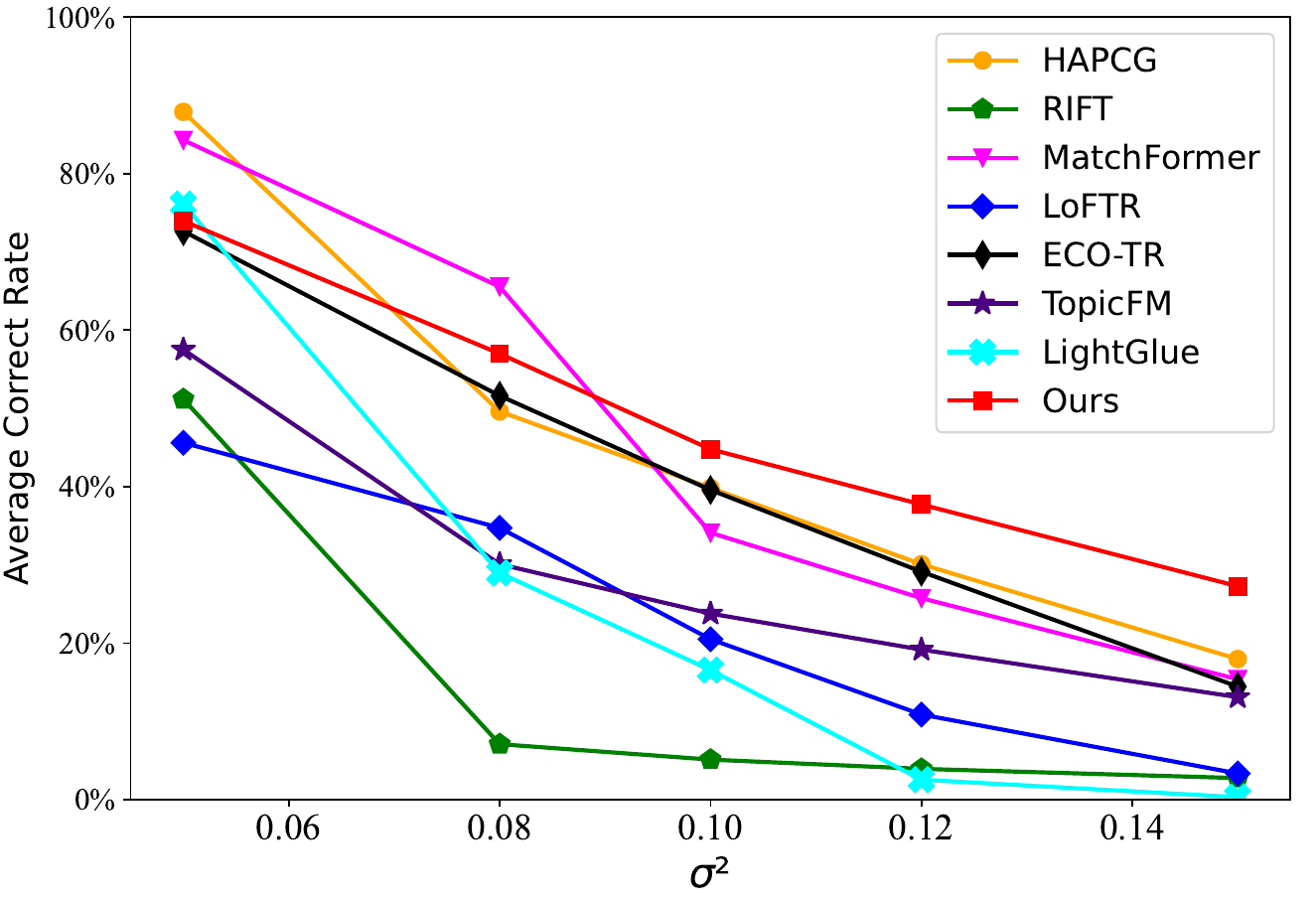}
		\label{fig12c}
		\includegraphics[width=0.245\linewidth]{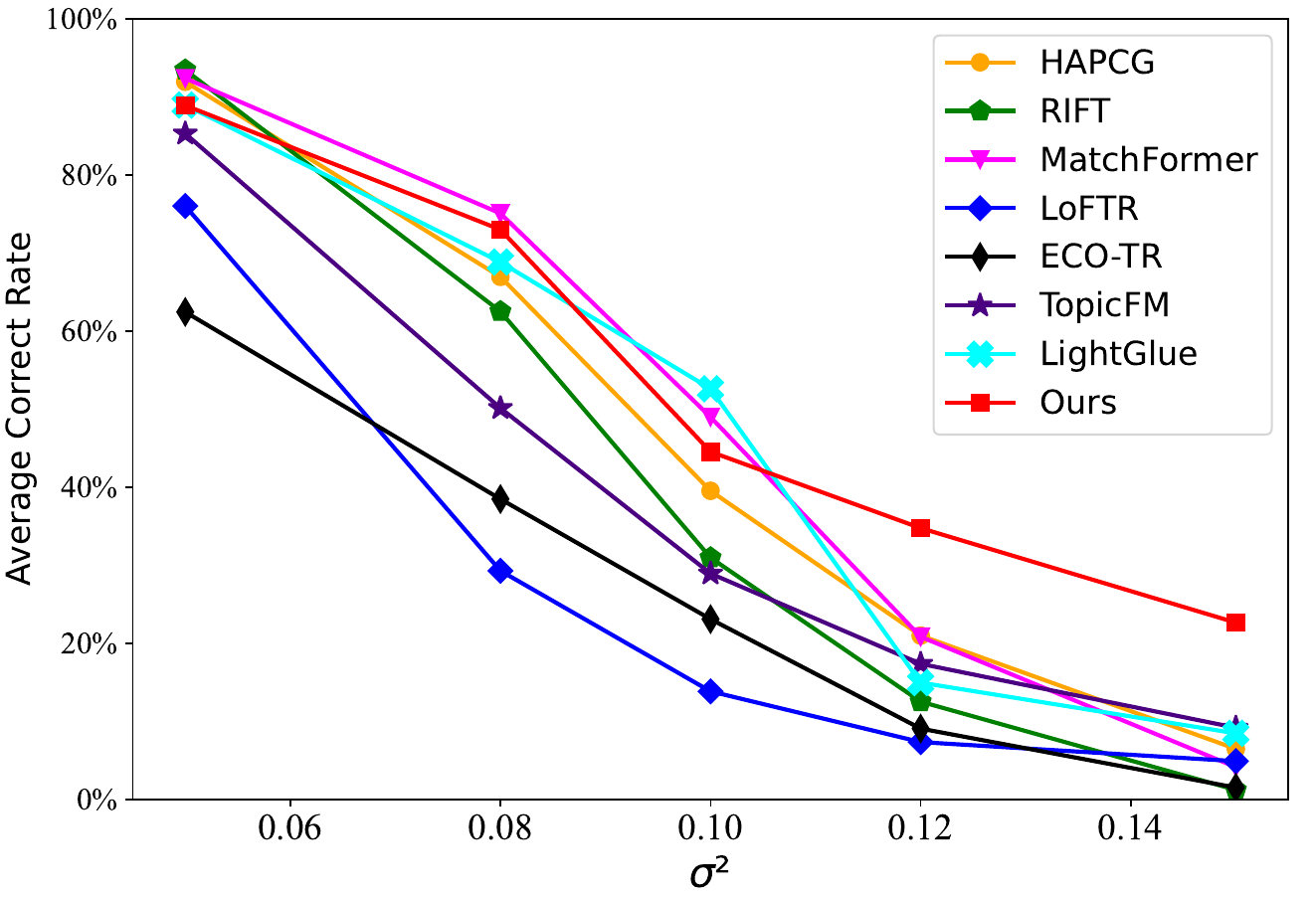}
		\label{fig12d}
		% \caption{histogram}
	\end{minipage}
	\caption{The average correct ratio of eight methods is compared as the intensity of stripe noise increases. From left to right, the datasets are optical-depth, infrared-optical, optical-map, and SAR-optical.}
	\label{Fig.12}
\end{figure*}

Results Analysis and Comparison: Fig. \ref{Fig.11} presents a qualitative comparison of eight methods for stripe noise images when $\sigma^2=0.1$. 
The results indicate that, compared to additive random noise, stripe noise is more challenging, leading to a more remarkable performance degradation for all methods. 
HAPCG and RIFT fail to match the first pair of images, and the NCM is smaller in the third and fourth pairs of images. 
The situation is similar to other methods. 
Table \ref{TABLE III} summarizes the quantitative results of eight methods under different intensities of stripe noise, and Fig. \ref{Fig.12} plots the corresponding curve of the average correct rate. 
(i) On Dataset 1, when $\sigma^2$ is set to 0.05, despite the NCM reduction for all methods, SR remains above 80\%, and the downward trend of the curves is consistent with the case of additive noise. 
(ii) On Dataset 2 and Dataset 3, when $\sigma^2$ is set to 0.15, the SR of RIFT is 0\%, and the SR of LightGlue is 10\% and 0\%, respectively; which demonstrates that both methods are more sensitive to noise. 
In contrast, most of the deep learning-based matching methods maintain SR above 30\%, which also suggests that the end-to-end deep learning matching methods are robust to noise interference.
(iii) On Dataset 4, when $\sigma^2$ is set to 0.5, the SR of ECO-TR is only 60\%, which is due to the fact that the noise weak texture image will reduce the performance of ECO-TR significantly, 
while the SR and the average correct rate of other methods is above 80\%. 
As $\sigma^2$ increases from 0.05 to 0.10, our method can maintain a high accuracy while other methods rapidly decline. 
For example, LoFTR decreases from 76.02\% to 13.82\%, RIFT decreases from 93.45\% to 30.95\%, TopicFM declined from 77.96\% to 32.85\% and LightGlue fell from 77.83\% to 34.84\%. 
When $\sigma^2$ increases from 0.10 to 0.15, the average correct rate of others is below 10\%, whereas ours maintains at 22.63\%. 
Overall, our proposed method has better robustness in matching noise images, outperforming other methods.  

\begin{table*}[htpb]
	\caption{THE QUANTITATIVE COMPARISON RESULTS WITH STRIPE NOISE}
	\fontsize{9}{11}\selectfont 
	\resizebox{\linewidth}{!}{
		\begin{tabular}{|p{2.0cm}<{\centering}|p{3.0cm}<{\centering}|p{2.3cm}<{\centering}|p{2.1cm}<{\centering}|p{2.1cm}<{\centering}|p{2.1cm}<{\centering}|p{2.1cm}<{\centering}|}
		\hline	
		\multicolumn{2}{|c|}{\diagbox[width=6cm]{\makecell{Data / Method / Metric \\ (NCM / SR / RMSE)}}{\makecell \\ \\ {${\sigma}^{2}$}}} & 0.05 & 0.08 & 0.10 & 0.12 & 0.15 \cr
		\hline	
		\multirow{8}*{\makecell{Dataset\ 1 \\ Optical-Depth}}
		&\textbf{HAPCG}                 &196 / 80 / 4.75	&107 / 70 / 7.35	&58 / 40 / 12.66	&28 / 20 / 16.32	&10 / 10 / 18.22 \\
		&\textbf{RIFT}                  &173 / 90 / 3.02	&63 / 80 / 5.47	    &17 / 70 / 7.20	    &5 / 20 / 16.34	    &3 / 20 / 16.36  \\
		&\textbf{LoFTR\ +\ FSC}         &166 / 90 / 3.24	&101 / 60 / 9.52	&77 / 50 / 11.12	&46 / 50 / 11.31	&22 / 40 / 13.06 \\
		&\textbf{MatchFormer\ +\ FSC}   &240 / 100 / 2.49	&88 / 80 / 5.26	    &37 / 60 / 9.63	    &19 / 50 / 11.53	&7 / 30 / 14.85  \\

		&\textbf{ ECO-TR\ +\ FSC}      & 951 / 100 / 2.51	& 616 / 100 / 2.58	& 448 / 100 / 2.60	& 305 / 100 / 2.69	 & 157 / 100 / 2.72  \\
		&\textbf{ TopicFM\ +\ FSC}     & 268 / 100 / 2.31	& 187 / 100 / 2.45	& 140 / 100 / 2.53	& 100 / 90 / 4.26    & 53 / 70 / 7.78    \\
		&\textbf{ LightGlue\ +\ FSC}   & 342 / 100 / 2.38	& 193 / 100 / 2.51	& 46 / 80 / 6.19    & 9   / 20 / 16.53   &  9 / 20 / 16.48   \\			
		
		&\textbf{Ours}                  &689 / 90 / 4.14	&546 / 80 / 6.30	&377 / 60 / 10.75	&228 / 50 / 12.16	&130 / 30 / 16.30 \\
		\hline
		\multirow{8}*{\makecell{Dataset\ 2 \\ Infrared-Optical}}
		&\textbf{HAPCG}                 &216 / 90 / 3.71	&141 / 80 / 5.89	&85 / 60 / 9.72	    &30 / 50 / 11.43	&17 / 40 / 13.15  \\
		&\textbf{RIFT}                  &116 / 90 / 3.49	&52 / 50 / 10.92	&18 / 20 / 16.35	&9 / 10 / 18.16	    &5 / 0 / 20       \\
		&\textbf{LoFTR\ +\ FSC}         &121 / 90 / 3.52	&53 / 70 / 7.23	    &40 / 70 / 7.85	    &32 / 60 / 9.51	    &20 / 30 / 14.72  \\
		&\textbf{MatchFormer\ +\ FSC}   &344 / 100 / 2.68	&167 / 80 / 6.07	&100 / 50 / 11.27	&52 / 40 / 13.04	&17 / 30 / 14.78  \\
		
		&\textbf{ ECO-TR\ +\ FSC }     & 727 / 70 / 7.70     & 511 / 70 / 7.78    & 389 / 60 / 9.51 	& 270 / 50 / 11.26 	  & 117 / 40 / 13.14   \\
		&\textbf{ TopicFM\ +\ FSC }    & 342 / 100 / 2.09    & 212 / 100 / 2.28   & 144 / 90 / 2.38 	& 96 / 90 / 4.13      & 45 / 70 / 7.68     \\
		&\textbf{ LightGlue\ +\ FSC }  & 172 / 90 / 4.14     & 131 / 50 / 11.07   & 77 / 50 / 11.22 	& 35 / 20 / 16.54     & 5 / 10 / 18.29    \\		
				
		&\textbf{Ours}                  &392 / 80 / 6.05	&286 / 70 / 8.33	&230 / 60 / 10.55	&137 / 50 / 12.20	&66 / 40 / 14.64  \\
		\hline
		\multirow{8}*{\makecell{Dataset\ 3 \\ Optical-Map}}
		&\textbf{HAPCG}                 &225 / 70 / 2.25	&127 / 60 / 9.45	&102 / 50 / 10.87	&60 / 30 / 14.46	&9 / 10 / 18.19  \\
		&\textbf{RIFT}                  &130 / 90 / 3.56	&18 / 30 / 14.53	&12 / 0 / 20	    &10 / 0 / 20	    &8 / 0 / 20      \\
		&\textbf{LoFTR\ +\ FSCFSC}      &109 / 60 / 9.13	&83 / 60 / 9.34	    &55 / 60 / 9.65	    &43 / 60 / 9.91	    &35 / 60 / 10.40 \\
		&\textbf{MatchFormer\ +\ FSC}   &252 / 80 / 5.71	&196 / 70 / 7.17	&102 / 70 / 7.48	&77 / 70 / 7.85	    &46 / 60 / 9.69  \\
		
		&\textbf{ ECO-TR\ +\ FSC }         & 1237 / 100 / 2.67    & 879 / 100 / 2.71   & 674 / 90 / 4.31 	& 496 / 90 / 4.46 	& 246 / 90 / 4.36  \\
		&\textbf{ TopicFM\ +\ FSC }        & 237 / 100 / 1.96     & 124 / 80 / 5.51    & 98 / 80 / 5.60     & 79 / 70 / 7.43    & 54 / 50 / 11.13  \\
		&\textbf{ LightGlue\ +\ FSC}       & 239 / 80 / 5.61	  &  91 / 60 / 9.40    & 52 / 40 / 12.91    & 7 / 30 / 14.80    &  1 / 0 / 20      \\
				
		&\textbf{Ours}                  &545 / 90 / 3.93	&420 / 80 / 6.29	&330 / 70 / 8.34	&278 / 60 / 10.27	&201 / 60 / 10.65 \\
		\hline
		\multirow{8}*{\makecell{Dataset\ 4 \\ SAR-Optical}}
		&\textbf{HAPCG}                 &114 / 80 / 5.52	&83 / 60 / 9.17	    &49 / 30 / 14.55	&26 / 20 / 16.36	&8 / 20 / 16.96   \\
		&\textbf{RIFT}                  &157 / 100 / 1.93	&105 / 70 / 7.34	&52 / 60 / 9.10	    &21 / 30 / 14.50	&2 / 10 / 18.13   \\
		&\textbf{LoFTR\ +\ FSCFSC}      &187 / 90 / 4.24	&72 / 80 / 5.95	    &34 / 70 / 7.81	    &18 / 60 / 9.46	    &12 / 40 / 12.59  \\
		&\textbf{MatchFormer\ +\ FSC}   &315 / 80 / 6.07	&256 / 70 / 7.81	&167 / 60 / 9.59	&71 / 50 / 11.30	&14 / 40 / 13.08  \\ 

		&\textbf{ ECO-TR\ +\ FSC }      & 138 / 60 / 9.55 	 & 85 / 60 / 9.64    & 51 / 50 / 11.26 	 & 20 / 30 / 14.69 	 & 3 / 20 / 16.47    \\
		&\textbf{ TopicFM\ +\ FSC  }    & 177 / 100 / 2.27 	 & 104 / 90 / 4.31 	 & 60 / 90 / 4.38    & 36 / 70 / 7.68    & 19 / 40 / 13.12   \\
		&\textbf{ LightGlue\ +\ FSC }   & 137 / 80 / 5.99    & 106 / 80 / 6.02 	 & 80 / 80 / 6.11    & 22 / 40 / 13.00   & 13 / 20 / 16.45   \\
		
		&\textbf{Ours}                  &609 / 90 / 4.21	&500 / 80 / 5.97	&305 / 60 / 10.13	&238 / 50 / 12.12	&155 / 40 / 14.01 \\
		\hline
		\end{tabular}
	}
	\label{TABLE III}
	% \end{center} 
\end{table*}

\subsection{Ablation Study}
We conducted ablation experiments on test dataset to demonstrate the necessity of each module. 
The model is trained using 10\% of the training data. 
This setting aims to improve efficiency as well as validate the stability of the model design with less training data. 
Table \ref{TABLE IV} summarizes the quantitative results for different module configurations. 

\emph{1)} Removing the feature preprocessing module resulted in a performance decrease, with NCM, SR, and RMSE decreasing by 20.51\%, 10\%, and 25.25\%, respectively. 
It indicates that the multi-scale feature representation is crucial for distinguishing objects or regions of different sizes and preserving the local features of various scales. 
This can improve the success rate of feature matching. 

\emph{2)} Replacing normalized positional encoding with absolute positional encoding will degrade the performance of the method, with NCM, SR, and RMSE reduced to 11.87\%, 5\%, and 10.56\%, respectively. 
This indicates that normalized positional encoding provides a more precise representation of the position information in the feature map, which is pivotal for distinguishing similar features. 

\emph{3)} With the increase of feature enhancement transformer layers at the coarse and fine levels, matching performance improves. 
This indicates that the alternating combination of multilevel self-attention and cross-attention is effective, and the local and global information within/between images can be aggregated into feature descriptors, 
which makes feature descriptors more discriminable, and enhances the performance of feature matching. 
However, as the number of layers increases, the performance improvement is not obvious. 
When $L_1=6,L_2=3$, the NCM and RMSE are only improved by 1.21\% and 0.77\%, respectively, and the RT is increased by 25.55\%; 
when $L_1=8,L_2=4$, the NCM and RMSE are only improved by 2.96\% and 1.54\%, respectively, and the RT is increased by 51.41\%; 
which shows that our methods have reasonableness in the setting of model parameters. 

\emph{4)} Replacing the outlier removal network with the FSC algorithm, the average processing time of the method increases by about 0.14s. 
Although the independent FSC algorithm can return a similar result of NCMs, our outlier removing network outperforms the FSC algorithm in terms of processing speed and efficiency.

\begin{table*}[htbp]
	\centering
	\caption{THE QUANTITATIVE COMPARISONS WITH ABLATION STUDY}
	\fontsize{9}{13.5}\selectfont  
	\resizebox{\linewidth}{!}{		
		\begin{tabular}{|l|c|c|c|c|}
		\hline	
		\multirow {2}{*}{Method}  & \multicolumn{4}{c|}{Average statistic value}\\
		\cline{2-5} & NCM \bm{$\uparrow$}   & SR(\%) \bm{$\uparrow$}  & RMSE(pixels) \bm{$\downarrow$}  & RT(s) \bm{$\downarrow$}  \\
		\hline
        1)\ Removing the feature preprocessing module     & 589   & 85  & 4.86  & 3.01  \\
		\hline
        2)\ Replacing normalized positional encoding with absolute positional encoding      & 653   & 90  & 4.29  & 3.35  \\
		\hline
		\multirow{2}{*}{\makecell{3) The number of feature enhancement transformer layers for coarse-level \\ and fine-level are increased to $L_1=6,L_2=3$ and $L_1=8,L_2=4$} }   & 750   & 95  & 3.85  & 3.98  \\
			                                          & 763   & 95  & 3.82  & 4.80  \\
		\hline						
		4)\ Replacing the outlier removal network with FSC                       & 754   & 95  & 3.93  & 3.31  \\
		\hline
		\quad \ Full robust matching method                   & 741   & 95  & 3.88  & 3.17 \\
		\hline	
		\end{tabular}
	}
	\label{TABLE IV}
	% \end{center}
\end{table*}

\subsection{Complexity and Efficiency Analysis}
\begin{table}[htbp]
	\centering
	\caption{COMPLEXITY AND EFFICIENCY COMPARISON}
	\fontsize{3}{5}\selectfont  
	\resizebox{\linewidth}{!}{
		% \begin{tabular}{|p{1cm}|p{2cm}<{\centering}|p{3cm}<{\centering}|}
	\begin{tabular}{!{\vrule width0.2pt}c!{\vrule width0.2pt}c!{\vrule width0.2pt}c!{\vrule width0.2pt}}
	\Xhline{0.2pt}	
	Method  & Params (MB) & GFLOPs \\
	% \textbf{Methods} & \textbf{Complexity} & \textbf{Params} & \textbf{GFLOPs} \\
	\Xhline{0.2pt}	
	\textbf{LoFTR}   & 11.56  & 307.64  \\
	\Xhline{0.2pt}
	\textbf{MatchFormer}   & 20.26  & 307.29  \\
	\Xhline{0.2pt}
	
	\textbf{ ECO-TR}    &  29.35 &  452.23  \\
	\Xhline{0.2pt}                                                                                                                                                 
	\textbf{ TopicFM}   &  11.60 &  180.43  \\
	\Xhline{0.2pt}                                                                                                                                                
	\textbf{ LightGlue} &  12.67 &  193.5   \\		
		
	\Xhline{0.2pt}
	\textbf{Ours}     & 14.86  & 319.58 \\
	\Xhline{0.2pt}
	\end{tabular} }
	\label{TABLE V}
	% \end{center}
\end{table}

To evaluate the space complexity and time complexity of the proposed method, we measured the number of parameters (Params) and computational cost (GFLOPs) of our method and the comparison methods; 
where FLOPs is also the number of floating point operations used for measuring the complexity of a neural network, $1$ GFLOPs = $10^6$ FLOPs. We calculate the average values of these metrics using testing dataset with image resolution of 512$\times$512. 
Since HAPCG and RIFT belong to the non-deep learning methods, they are not compared. 

As shown in Table \ref{TABLE V}, LoFTR is used as the evaluation baseline. 
(i) LoFTR has fewer model params and faster inference speed, whereas the matching performance needs to be improved, as shown on Table \ref{TABLE I}. 
(ii) Although the params of MatchFormer reaches 20.26MB, which is 75.25\% higher than the baseline; moreover, it uses an efficient and lightweight FPN decoder to reduce the space complexity of the model, resulting in less computational complexity and GFLOPs is equal to the baseline. 
(iii) ECO-TR is a three-stage matching network based on a coarse-to-fine transformer framework, where each stage is a transformer building block of three encoders and three decoders. Therefore, the params and the computational cost of ECO-TR network are relatively large. 
(iv) LightGlue adopts a lightweight transformer network with a low GFLOPs of 15.66. However, LightGlue uses SuperPoint \cite{ref23} for keypoint detection, which has a computational cost of 88.92 GFLOPs for a single image and increases LightGlue's overall GFLOPs to 193.5. In general, the params for LightGlue is approximate to the baseline and outperforms it in efficiency. 
(v) To achieve a fast computation, TopicFM is designed as an efficient, lightweight network for each coarse-to-fine step. In particular, TopicFM reduces computation by only focusing on the same semantic areas between the images to learn feature matching. Compared to the baseline, TopicFM achieved approximately a 40\% reduction in GFLOPs. 
(vi) Our method surpasses the baseline by 28.54\% and 3.88\% in Params and GFLOPs, respectively, 
however, the NCM outperforms the baseline by 89.51\% and MatchFormer by 84.32\%, and the computation time is close to each other. 
Combine with the above analysis, our method is more advantageous in terms of computational efficiency.

\section{Conclusion}
Noise could interfere with feature extraction and matching of remote sensing images, 
leading to degradation of matching method performance and inaccurate matching results. 
Aiming at the problems, we propose a robust matching method that improves feature descriptor discriminability and robustness through dense feature extraction, multi-scale feature aggregation, 
and feature enhancement with multi-level attention alternating combinations; 
Subsequently, we obtain highly efficient and accurate matching results by adopting a coarse-to-fine matching strategy and an outlier removal network with a binary classification mechanism. 
Experimental results demonstrate the comprehensive advantages of our method in multisource remote sensing image feature matching, and our method shows more significant performance than the state-of-the-art methods in challenging noise image matching.

The method proposed in this paper has two main limitations: 
(i) The RMSE is large, which is caused by large pixel errors of the matched points. 
The accurate matching between points depends on the manual setting of confidence threshold $\tau_c$ in the coarse matching estimation. 
As the threshold $\tau_c$ is smaller, the accuracy will be higher, but the density of matched points will decrease, and vice versa. 
Therefore, the optimal matching algorithm needs to be investigated to balance the relationship between correctness and denseness. 
(ii) The outlier removal network uses the coordinates of the matching points as input information, but when normalizing the coordinates of the matching points, it needs to use the intrinsic matrix provided by the dataset, which may be difficult to satisfy in some cases. 
Moreover, it restricts the network to be trained on some specific datasets. 
We should consider its improvement in the next work.

% In the future, we will focus on the effect of network generalization and robustness improvement after adding mixed noise to the training dataset during network training.

\ifCLASSOPTIONcaptionsoff
  \newpage
\fi


\begin{thebibliography}{}
\providecommand{\url}[1]{#1}
\csname url@samestyle\endcsname
\providecommand{\newblock}{\relax}
\providecommand{\bibinfo}[2]{#2}
\providecommand{\BIBentrySTDinterwordspacing}{\spaceskip=0pt\relax}
\providecommand{\BIBentryALTinterwordstretchfactor}{4}
\providecommand{\BIBentryALTinterwordspacing}{\spaceskip=\fontdimen2\font plus
\BIBentryALTinterwordstretchfactor\fontdimen3\font minus \fontdimen4\font\relax}
\providecommand{\BIBforeignlanguage}[2]{{%
\expandafter\ifx\csname l@#1\endcsname\relax
\typeout{** WARNING: IEEEtran.bst: No hyphenation pattern has been}%
\typeout{** loaded for the language `#1'. Using the pattern for}%
\typeout{** the default language instead.}%
\else
\language=\csname l@#1\endcsname
\fi
#2}}
\providecommand{\BIBdecl}{\relax}
\BIBdecl

\end{thebibliography}


\begin{thebibliography}{1}
	\bibliographystyle{IEEEtran}
	\bibitem{ref1}
	S. Xu, S. Chen, R. Xu, C. Wang, P. Lu, L. Guo, “Local feature matching using deep learning: A survey.” \textit{Inf. Fusion.}, vol. 107, no. 102344, pp. 1566-2535, 2024.
	% Ma, J., Jiang, X., Fan, A. \textit{et al}. “Image Matching from Handcrafted to Deep Features: A Survey.” \textit{Int J Comput Vis.},vol. 129, no. 1, pp. 23-79, Jan. 2021.

	\bibitem{ref2}
	% S. Zhao, Z. Zhang, W. Guo and Y. Luo, “An Automatic Ship Detection Method Adapting to Different Satellites SAR Images with Feature Alignment and Compensation Loss,” \textit{IEEE Trans. Geosci. Remote Sens.}, vol. 60, pp. 1-17, 2022.
	R. Zhou, D. Quan, S. Wang, C. Lv, X. Cao, J. Chanussot, Y. Li, L. Jiao, “A Unified Deep Learning Network for Remote Sensing Image Registration and Change Detection,” \textit{IEEE Trans. Geosci. Remote Sens.}, vol. 62, pp. 1-16, 2024.
	\bibitem{ref3}
	% X. Wan, Y. Shao, S. Zhang and S. Li, “Terrain Aided Planetary UAV Localization Based on Geo-referencing,” \textit{IEEE Trans. Geosci. Remote Sens.}, vol. 60, pp. 1-18, 2022.
	Z. Wang, D. Shi, C. Qiu, S. Jin, T. Li, Y. Shi, Z. Liu, Z. Qiao, “Sequence Matching for Image-Based UAV-to-Satellite Geolocalization,” \textit{IEEE Trans. Geosci. Remote Sens.}, vol. 62, pp. 1-15, 2024
	\bibitem{ref4}
	M. Tao, J. Li, J. Chen, Y. Liu, Y. Fan, J. Su, L. Wang, “Radio Frequency Interference Signature Detection in Radar Remote Sensing Image Using Semantic Cognition Enhancement Network,” \textit{IEEE Trans. Geosci. Remote Sens.}, vol. 60, pp. 1-14, 2022.

	\bibitem{ref5}
	C. Wang \textit{et al.}, “Translution-SNet: A Semisupervised Hyperspectral Image Stripe Noise Removal Based on transformer and CNN,” \textit{IEEE Trans. Geosci. Remote Sens.}, vol. 60, pp. 1-14, 2022.

	\bibitem{ref6}
	F. Dellinger, J. Delon, Y. Gousseau, J. Michel, and F. Tupin, “SAR-SIFT: A SIFT-like method for SAR images,” \textit{IEEE Trans. Geosci. Remote Sens.}, vol. 53, no. 1, pp. 453-466, Jan. 2015.
	% J. Li, Q. Hu, and M. Ai, “RIFT: Multi-modal image matching based on radiation-variation insensitive feature transform,” \textit{IEEE Trans. Image Process.}, vol. 29, pp. 3296-3310, 2020.

	\bibitem{ref7}
	K. M. Yi, E. Trulls, V. Lepetit, and P. Fua, “LIFT: Learned invariant feature transform,” in \textit{Proc. Eur. Conf. Comput. Vis.}, Berlin, Germany: Springer, 2016, pp. 467-483.
	% J. Li, W. Xu, P. Shi, Y. Zhang and Q. Hu, “LNIFT: Locally Normalized Image for Rotation Invariant Multimodal Feature Matching,” \textit{IEEE Trans. Geosci. Remote Sens.}, vol. 60, pp. 1-14, 2022.
	\bibitem{ref8}
	W. Ma {et al}., “Remote Sensing Image Registration With Modified SIFT and Enhanced Feature Matching,” \textit{IEEE Geosci. Remote Sens. Lett.}, vol. 14, no. 1, pp. 3-7, Jan. 2017.
	% Y. Yao, Y. Zhang, Y. Wan, X. Liu, X. Yan and J. Li, "Multi-Modal Remote Sensing Image Matching Con-sidering Co-Occurrence Filter,” \textit{IEEE Trans. Image Process.}, vol. 31, pp. 2584-2597, 2022.
	\bibitem{ref9}
	% B. Zhu, C. Yang, J. Dai, J. Fan, Y. Qin and Y. Ye, “R2FD2: Fast and Robust Matching of Multimodal Remote Sensing Images via Repeatable Feature Detector and Rotation-Invariant Feature Descriptor,” \textit{IEEE Trans. Geosci. Remote Sens.}, vol. 61, pp. 1-15, 2023.
	Y. Xiang, F. Wang and H. You, “OS-SIFT: A Robust SIFT-Like Algorithm for High-Resolution Optical-to-SAR Image Registration in Suburban Areas,” \textit{IEEE Trans. Geosci. Remote Sens.}, vol. 56, no. 6, pp. 3078-3090, June 2018.
	\bibitem{ref10}
	Y. Ye, J. Shan, L. Bruzzone and L. Shen, “Robust Registration of Multimodal Remote Sensing Images Based on Structural Similarity,” \textit{IEEE Trans. Geosci. Remote Sens.}, vol. 55, no. 5, pp. 2941-2958, May 2017.
	\bibitem{ref11}
	J. Fan, Y. Wu, M. Li, W. Liang and Y. Cao, “SAR and Optical Image Registration Using Nonlinear Diffusion and Phase Congruency Structural Descriptor,” \textit{IEEE Trans. Geosci. Remote Sens.}, vol. 56, no. 9, pp. 5368-5379, Sept. 2018.
	\bibitem{ref12}
	Y. Ye, L. Bruzzone, J. Shan, F. Bovolo and Q. Zhu, “Fast and Robust Matching for Multimodal Remote Sensing Image Registration,” \textit{IEEE Trans. Geosci. Remote Sens.}, vol. 57, no. 11, pp. 9059-9070, Nov. 2019.
	\bibitem{ref13}
	C. Gao, W. Li, R. Tao and Q. Du, “MS-HLMO: Multiscale Histogram of Local Main Orientation for Remote Sensing Image Registration,” \textit{IEEE Trans. Geosci. Remote Sens.}, vol. 60, pp. 1-14, 2022.
	% Y. Ye, Q. Wang, H. Zhao, X. Teng, Y. Bian and Z. Li, "Fast and Robust Optical-to-SAR Remote Sensing Image Registration Using Region-Aware Phase Descriptor," \textit{IEEE Trans. Geosci. Remote Sens.}, vol. 62, pp. 1-12, 2024.
	\bibitem{ref14}
	% S. Wang, D. Quan, X. Liang, M. Ning, Y. Guo, and L. Jiao, “A deep learning framework for remote sensing image registration,” \textit{ISPRS J. Photogramm. Remote Sens.}, vol. 145145, pp. 148-164, Nov. 2018.
	W. Ma, J. Zhang, Y. Wu, L. Jiao, H. Zhu and W. Zhao, “A Novel Two-Step Registration Method for Remote Sensing Images Based on Deep and Local Features,” \textit{IEEE Trans. Geosci. Remote Sens.}, vol. 57, no. 7, pp. 4834-4843, July 2019.

	\bibitem{ref15}
	Y.He, Y.Hu, W.Zhao, J.Li, Y.Liu, Y.Han and J. Wen, “DarkFeat: noise-robust feature detector and descriptor for extremely low-light RAW images.” \textit{Proc. AAAI Conf. Artif. Intell.} Vol. 37. No. 1. 2023.

	\bibitem{ref16}
	Y. Zhang, C. Lan, H. Zhang, G. Ma, and H. Li, “Multimodal Remote Sensing Image Matching via Learning Features and Attention Mechanism,” \textit{IEEE Trans. Geosci. Remote Sens.}, vol. 62, pp. 1-20, 2024.
	
	\bibitem{ref17}
	K. Dai, T. Xie, K. Wang, Z. Jiang, R. Li, and L. Zhao, “FMAP: Learning robust and accurate local feature matching with anchor points,” \textit{Expert Systems with Applications}, p. 121328, 2023.
	\bibitem{ref18}
	Y. Ye, C. Yang, G. Gong, P. Yang, D. Quan and J. Li, “Robust Optical and SAR Image Matching Using Attention-Enhanced Structural Features,” \textit{IEEE Trans. Geosci. Remote Sens.}, vol. 62, pp. 1-12, 2024.
	
	\bibitem{ref19}
	Z. Shen, J. Sun, Y. Wang, X. He, H. Bao, X. Zhou, “Semi-Dense Feature Matching With Transformers and its Applications in Multiple-View Geometry”, \textit{IEEE Trans. Pattern Anal. Mach. Intell.}, vol.45, no.6, pp.7726-7738, 2023.
	
	\bibitem{ref20}
	Y. Liu, W. He, and H. Zhang, “GLoCNet: Robust Feature Matching with Global-Local Consistency Network for Remote Sensing Image Registration,” \textit{IEEE Trans. Geosci. Remote Sens.}, vol. 61, pp. 1-13, 2023.
	\bibitem{ref21}
	A. Vaswani \textit{et al}., “Attention is all you need,” in \textit{Proc. Adv. Neural Inf. Process. Syst.}, Long Beach, CA, USA, Jun. 2017, pp. 5998-6008.
	
	\bibitem{ref22}
	K. Han \textit{et al}., “A Survey on Vision Transformer,” \textit{IEEE Trans. Pattern Anal. Mach. Intell.}, vol. 45, no. 1, pp. 87-110, 1 Jan. 2023.

	\bibitem{ref23}
	P. Sarlin, D. DeTone, T. Malisiewicz, and A. Rabinovich, “SuperGlue: Learning feature matching with graph neural networks,” in \textit{Proc. IEEE Conf. Comput. Vis. Pattern Recognit. (CVPR)}, Seattle, WA,USA, Jun. 2020, pp. 4937-4946.

	\bibitem{ref24}
	J. Sun, Z. Shen, Y. Wang, H. Bao and X. Zhou, “LoFTR: Detector-Free Local Feature Matching with transformers,” in \textit{Proc. IEEE Conf. Comput. Vis. Pattern Recognit. (CVPR)}, Nashville, TN, USA, 2021, pp. 8918-8927.

	\bibitem{ref25}
	A. Katharopoulos, A. Vyas, N. Pappas, and F. Fleuret, “Transformers are RNNs: Fast autoregressive transformers with linear attention,” in \textit{Proc. IEEE Int. Conf. Mach. Learn. (ICML)}, Jul. 2020, pp. 5156-5165.

	\bibitem{ref26}
	D. Quan \textit{et al}., “Deep Feature Correlation Learning for MultiModal Remote Sensing Image Registration,” \textit{IEEE Trans. Geosci. Remote Sens.}, vol. 60, pp. 1-16, 2022.

	\bibitem{ref27}
	J. Guo, G. Xiao, Z. Tang, S. Chen, S. Wang and J. Ma, “Learning for Feature Matching via Graph Context Attention,” \textit{IEEE Trans. Geosci. Remote Sens.}, vol. 61, pp. 1-14, 2023.

	\bibitem{ref28}
	Z. Qin \textit{et al}., “GeoTransformer: Fast and Robust Point Cloud Registration With Geometric Transformer,” \textit{IEEE Trans. Pattern Anal. Mach. Intell.}, vol. 45, no. 8, pp. 9806-9821, Aug. 2023.

	\bibitem{ref29}
	L. Zhao, Y. Liu, C. Men and Y. Men, “Double Propagation Stereo Matching for Urban 3D Reconstruction From Satellite Imagery,” \textit{IEEE Trans. Geosci. Remote Sens.}, vol. 60, pp. 1-17, 2022.

	\bibitem{ref30}
	L. Li, L. Han, M. Ding and H. Cao, “Multimodal Image Fusion Framework for End-to-End Remote Sensing Image Registration,” \textit{IEEE Trans. Geosci. Remote Sens.}, vol. 61, pp. 1-14, 2023.

	\bibitem{ref31}
	Y. Yu, X. Yang, J. Li and X. Gao, “A Refined Hybrid Network for Object Detection in Aerial Images,” \textit{IEEE Trans. Geosci. Remote Sens.}, vol. 61, pp. 1-15, 2023.

	\bibitem{ref32}
	Y. Xiang, F. Wang, L. Wan, N. Jiao and H. You, “OS-Flow: A Robust method for Dense Optical and SAR Image Registration,” \textit{IEEE Trans. Geosci. Remote Sens.}, vol. 57, no. 9, pp. 6335-6354, Sept. 2019.

	\bibitem{ref33}
	H. Zhang \textit{et al}., “Optical and SAR Image Dense Registration Using a Robust Deep Optical Flow Framework,” \textit{IEEE J. Sel. Topics Appl.Earth Observ. Remote Sens.}, vol. 16, pp. 1269-1294, 2023.

	\bibitem{ref34}
	I. Rocco, M. Cimpoi, R. Arandjelović, A. Torii, T. Pajdla and J. Sivic, “NCNet: Neighbourhood Consensus Networks for Estimating Image Correspondences,” \textit{IEEE Trans. Pattern Anal. Mach. Intell.}, vol. 44, no. 2, pp. 1020-1034, 1 Feb. 2022.

	\bibitem{ref35}
	X. Li, K. Han, S. Li, and V. Prisacariu, “Dual-resolution correspondence networks,” in \textit{Proc. Adv. Neural Inf. Process. Syst.}, vol. 33, Dec. 2020, pp. 1-14.

	\bibitem{ref36}
	P. Truong, M. Danelljan and R. Timofte, “GLU-Net: Global-Local Universal Network for Dense Flow and Correspondences,” in \textit{Proc. IEEE Conf. Comput. Vis. Pattern Recognit. (CVPR)}, Seattle, WA, USA, 2020, pp. 6257-6267.

	\bibitem{ref37}
	Q. Zhou, T. Sattler and L. Leal-Taixé, “Patch2Pix: Epipolar-Guided Pixel-Level Correspondences,” in \textit{Proc. IEEE Conf. Comput. Vis. Pattern Recognit. (CVPR)}, Nashville, TN, USA, 2021, pp. 4667-4676.

	\bibitem{ref38}
	P. Truong, M. Danelljan, R. Timofte and L. Van Gool, “PDC-Net+: Enhanced Probabilistic Dense Correspondence Network,” \textit{IEEE Trans. Pattern Anal. Mach. Intell.}, vol. 45, no. 8, pp. 10247-10266, Aug. 2023.

	\bibitem{ref39}
	M. A. Fischler and R. C. Bolles, “Random sample consensus: A paradigm for model fitting with applications to image analysis and automated cartography”, \textit{Commun. ACM}, vol. 24, no. 6, pp. 381-395, 1981.

	\bibitem{ref40}
	P. H. S. Torr and A. Zisserman, “MLESAC: A new robust estimator with application to estimating image geometry”, \textit{Comput. Vis. Image Understand.}, vol. 78, no. 1, pp. 138-156, Apr. 2000.

	\bibitem{ref41}
	O. Chum and J. Matas, “Matching with PROSAC-Progressive sample consensus”, in \textit{Proc. IEEE Conf. Comput. Vis. Pattern Recognit. (CVPR)}, pp. 220-226, Jun. 2005.

	\bibitem{ref42}
	R. Raguram, O. Chum, M. Pollefeys, J. Matas and J.M. Frahm, “USAC: A universal framework for random sample consensus”, \textit{IEEE Trans. Pattern Anal. Mach. Intell.}, vol. 35, no. 8, pp. 2022-2038, Aug. 2013.

	\bibitem{ref43}
	D. Barath, J. Matas and J. Noskova, “MAGSAC: Marginalizing sample consensus”, in \textit{Proc. IEEE Conf. Comput. Vis. Pattern Recognit. (CVPR)}, pp. 10197-10205, Jun. 2019.

	\bibitem{ref44}
	J. Ma, J. Zhao, J. Jiang, H. Zhou, and X. Guo, “Locality preserving matching,” \textit{Int. J. Comput. Vis.}, vol. 127, no. 5, pp. 512-531, 2019.

	\bibitem{ref45}
	J. Ma, J. Jiang, H. Zhou, J. Zhao, and X. Guo, “Guided locality preserving feature matching for remote sensing image registration,” \textit{IEEE Trans. Geosci. Remote Sens.}, vol. 56, no. 8, pp. 4435-4447, Aug. 2018.

	\bibitem{ref46}
	L. Shen, Q. Xin, J. Zhu, X. Huang and T. Jin, “Frame-Based Locality Preservation Matching for Images Involving Large-Scale Transformations,” \textit{IEEE Trans. Geosci. Remote Sens.}, vol. 60, pp. 1-13, 2022.

	\bibitem{ref47}
	K. M. Yi, E. Trulls, Y. Ono, V. Lepetit, M. Salzmann and P. Fua, “Learning to Find Good Correspondences,” in \textit{Proc. IEEE Conf. Comput. Vis. Pattern Recognit. (CVPR)}, Salt Lake City, UT, USA, 2018, pp. 2666-2674.

	\bibitem{ref48}
	J. Zhang \textit{et al}., “OANet: Learning Two-View Correspondences and Geometry Using Order-Aware Network,” \textit{IEEE Trans. Pattern Anal. Mach. Intell.}, vol. 44, no. 6, pp. 3110-3122, June 2022.

	% K. He, X. Zhang, S. Ren and J. Sun, “Deep Residual Learning for Image Recognition,” in \textit{Proc. IEEE Conf. Comput. Vis. Pattern Recognit. (CVPR)}, Las Vegas, NV, USA, 2016, pp. 770-778.

	\bibitem{ref49}
	J. Ma, X. Jiang, J. Jiang, J. Zhao and X. Guo, “LMR: Learning a Two-Class Classifier for Mismatch Removal,” \textit{IEEE Trans. Image Process.}, vol. 28, no. 8, pp. 4045-4059, Aug. 2019.

	% Howard, Andrew G. \textit{et al}. “MobileNets: Efficient Convolutional Neural Networks for Mobile Vision Applications,” 2017, \textit{ArXiv:1704.04861}.

	\bibitem{ref50}
	J. Chen \textit{et al}., “LSV-ANet: Deep Learning on Local Structure Visualization for Feature Matching,” \textit{IEEE Trans. Geosci. Remote Sens.}, vol. 60, pp. 1-18, 2022.

	% Nicolas Carion, Francisco Massa, Gabriel Synnaeve, Nicolas Usunier, Alexander Kirillov, and Sergey Zagoruyko. “End-to-end object detection with transformers.” In \textit{Proc. Eur. Conf. Comput. Vis. (ECCV)}, Berlin, Germany: Springer-Verlag, 2020, pp. 213-229., 2020.

	\bibitem{ref51}
	J. Guo, G. Xiao, Z. Tang, S. Chen, S. Wang and J. Ma, “Learning for Feature Matching via Graph Context Attention,” \textit{IEEE Trans. Geosci. Remote Sens.}, vol. 61, pp. 1-14, 2023.

	% S. Li, Q. Zhao and Z. Xia, “Sparse-to-Local-Dense Matching for Geometry-Guided Correspondence Estimation,” \textit{IEEE Trans. Image Process.}, vol. 32, pp. 3536-3551, 2023.

	\bibitem{ref52}
	J. Li, Q. Hu, Y. Zhang, “Multimodal image matching: A scale-invariant algorithm and an open dataset,” \textit{ISPRS J. Photogramm. Remote Sens.}, vol. 201, pp. 77-88, Oct. 2023.
	% T. Y. Lin, P. Dollár, R. Girshick, K. He, B. Hariharan and S. Belongie, “Feature Pyramid Networks for Object Detection,” in \textit{Proc. IEEE Conf. Comput. Vis. Pattern Recognit. (CVPR)}, Honolulu, HI, USA, 2017, pp. 936-944.

	% A. Nibali, Z. He, S. Morgan, and L. Prendergast, “Numerical coordinate regression with convolutional neural networks,” 2018, \textit{ArXiv:1801.07372}.

	\bibitem{ref53}
	D. Quan, H. Wei, S. Wang, Y. Gu, B. Hou and L. Jiao, “A Novel Coarse-to-Fine Deep Learning Registration Framework for Multimodal Remote Sensing Images,” \textit{IEEE Trans. Geosci. Remote Sens.}, vol. 61, pp. 1-16, 2023.
	% Q. Wang, X. Zhou, B. Hariharan and N. Snavely. “Learning feature descriptors using camera pose supervision.” In \textit{Proc. Eur. Conf. Comput. Vis. (ECCV)},2020, pp. 757-774.

	\bibitem{ref54}
	Y. Xu, J. Li, C. Du and H. Chen, “NBR-Net: A Nonrigid Bidirectional Registration Network for Multitemporal Remote Sensing Images,” \textit{IEEE Trans. Geosci. Remote Sens.}, vol. 60, pp. 1-15, 2022.
	% Z. Li and N. Snavely, “MegaDepth: Learning Single-View Depth Prediction from Internet Photos,” in \textit{Proc. IEEE Conf. Comput. Vis. Pattern Recognit. (CVPR)}, Salt Lake City, UT, USA, 2018, pp. 2041-2050.

	\bibitem{ref55}
	L. Shi, R. Zhao, B. Pan, Z. Zou and Z. Shi, “Unsupervised Multimodal Remote Sensing Image Registration via Domain Adaptation,” \textit{IEEE Trans. Geosci. Remote Sens.}, vol. 61, pp. 1-11, 2023.
	% Jiayuan Li, Qingwu Hu, and Yongjun Zhang. “Multimodal image matching: A scale-invariant method and an open dataset.” \textit{ISPRS J. Photogramm. Remote Sens.}, vol. 204, pp. 77-88, Sep 2023.

	\bibitem{ref56}
	Y. Xiao, C. Zhang, Y. Chen, B. Jiang and J. Tang, “ADRNet: Affine and Deformable Registration Networks for Multimodal Remote Sensing Images,” \textit{IEEE Trans. Geosci. Remote Sens.}, vol. 62, pp. 1-13, 2024.

	% Y. Zhang \textit{et al}., “Histogram of the orientation of the weighted phase descriptor for multi-modal remote sensing image matching,” \textit{ISPRS J. Photogramm. Remote Sens.}, vol. 196196, pp. 11-15, Feb. 2023.

	\bibitem{ref57}
	J. Li, Q. Hu, and M. Ai, “RIFT: Multi-modal image matching based on radiation-variation insensitive feature transform,” \textit{IEEE Trans. Image Process.}, vol. 29, pp. 3296-3310, 2020.

	\bibitem{ref58}
	J. Li, W. Xu, P. Shi, Y. Zhang and Q. Hu, “LNIFT: Locally Normalized Image for Rotation Invariant Multimodal Feature Matching,” \textit{IEEE Trans. Geosci. Remote Sens.}, vol. 60, pp. 1-14, 2022.

	% Yongxiang Yao, Yongjun Zhang, Yi Wan, Xinyi Liu, Haoyu Guo. Heterologous Images Matching Considering Anisotropic Weighted Moment and Absolute Phase Orientation[J]. \textit{Geomatics and Information Science of Wuhan University}, 2021, 46(11): 1727-1736. doi: 10.13203/j.whugis20200702.

	\bibitem{ref59}
	Z. Fan, M. Wang, Y. Pi, Y. Liu and H. Jiang, “A Robust Oriented Filter-Based Matching Method for Multisource, Multitemporal Remote Sensing Images,” \textit{IEEE Trans. Geosci. Remote Sens.}, vol. 61, pp. 1-16, 2023.
	% Qing Wang, Jiaming Zhang, Kailun Yang, Kunyu Peng, Rainer Stiefelhagen, “MatchFormer: Interleaving Attention in Transformers for Feature Matching,” In \textit{Proc. Asian Conf. Comput. Vis.}, 2022, pp. 2746-2762.

	\bibitem{ref60}
	T. Y. Lin, P. Dollár, R. Girshick, K. He, B. Hariharan and S. Belongie, “Feature Pyramid Networks for Object Detection,” in \textit{Proc. IEEE Conf. Comput. Vis. Pattern Recognit. (CVPR)}, Honolulu, HI, USA, 2017, pp. 936-944.
	% Y. Wu, W. Ma, M. Gong, L. Su and L. Jiao, “A Novel Point-Matching method Based on Fast Sample Consensus for Image Registration,” \textit{IEEE Geosci. Remote Sens. Lett.}, vol. 12, no. 1, pp. 43-47, Jan. 2015.

	\bibitem{ref61}
	K. He, X. Zhang, S. Ren and J. Sun, “Deep Residual Learning for Image Recognition,” in \textit{Proc. IEEE Conf. Comput. Vis. Pattern Recognit. (CVPR)}, Las Vegas, NV, USA, 2016, pp. 770-778.

	\bibitem{ref62}
	Howard, Andrew G. \textit{et al}. “MobileNets: Efficient Convolutional Neural Networks for Mobile Vision Applications,” 2017, \textit{ArXiv:1704.04861}.

	\bibitem{ref63}
	Nicolas Carion, Francisco Massa, Gabriel Synnaeve, Nicolas Usunier, Alexander Kirillov, and Sergey Zagoruyko. “End-to-end object detection with transformers.” In \textit{Proc. Eur. Conf. Comput. Vis. (ECCV)}, Berlin, Germany: Springer-Verlag, 2020, pp. 213-229.
	
	\bibitem{ref64}
	S. Li, Q. Zhao and Z. Xia, “Sparse-to-Local-Dense Matching for Geometry-Guided Correspondence Estimation,” \textit{IEEE Trans. Image Process.}, vol. 32, pp. 3536-3551, 2023.

	\bibitem{ref65}
	A. Nibali, Z. He, S. Morgan, and L. Prendergast, “Numerical coordinate regression with convolutional neural networks,” 2018, \textit{ArXiv:1801.07372}.

	\bibitem{ref66}
	Q. Wang, X. Zhou, B. Hariharan and N. Snavely. “Learning feature descriptors using camera pose supervision.” In \textit{Proc. Eur. Conf. Comput. Vis. (ECCV)},2020, pp. 757-774.

	\bibitem{ref67}
	Z. Li and N. Snavely, “MegaDepth: Learning Single-View Depth Prediction from Internet Photos,” in \textit{Proc. IEEE Conf. Comput. Vis. Pattern Recognit. (CVPR)}, Salt Lake City, UT, USA, 2018, pp. 2041-2050.

	\bibitem{ref68}
	Y. Liu, X. Gong, J. Chen, S. Chen, and Y. Yang, “Rotation-invariant Siamese network for low-altitude remote-sensing image registration,” \textit{IEEE J. Sel. Topics Appl. Earth Observ. Remote Sens.}, vol. 13, pp. 5746-5758, 2020.
	% Y. Xiang, R. Tao, F. Wang, H. You, and B. Han, “Automatic registration of optical and SAR images via improved phase congruency model,” \textit{IEEE J. Sel. Topics Appl. Earth Observ. Remote Sens.}, vol. 13, pp. 5847-5861, 2020.
	
	% Jiayuan Li, Qingwu Hu, and Yongjun Zhang. “Multimodal image matching: A scale-invariant method and an open dataset.” \textit{ISPRS J. Photogramm. Remote Sens.} vol. 204, pp. 77-88, Sep 2023.

	\bibitem{ref69}
	Y. Zhang \textit{et al}., “Histogram of the orientation of the weighted phase descriptor for multi-modal remote sensing image matching,” \textit{ISPRS J. Photogramm. Remote Sens.}, vol. 196196, pp. 11-15, Feb. 2023.

	\bibitem{ref70}
	Y. Yao, Y. Zhang, Y. Wan, X. Liu, H. Guo. Heterologous Images Matching Considering Anisotropic Weighted Moment and Absolute Phase Orientation[J]. \textit{Geomatics and Information Science of Wuhan University}, 2021, 46(11): 1727-1736. doi: 10.13203/j.whugis20200702.

	\bibitem{ref71}
	Q. Wang, J. Zhang, K. Yang, K. Peng, R. Stiefelhagen, “MatchFormer: Interleaving Attention in Transformers for Feature Matching,” In \textit{Proc. Asian Conf. Comput. Vis.}, 2022, pp. 2746-2762.

	\bibitem{ref72}
	D. Tan, \textit{et al}., “ECO-TR: Efficient correspondences finding via coarse-to-fine refinement,” In \textit{Proc. Eur. Conf. Compute. Vis. (ECCV)}, Oct. 2022, pp. 317-334.

	\bibitem{ref73}
	K. Truong Giang, S. Song, and S. Jo. “TopicFM: Robust and Interpretable Topic-Assisted Feature Matching.” in \textit{Proc. AAAI Conf. Artif. Intell.}, vol. 37, no. 2, pp. 2447-2455, Jun. 2023.
	
	\bibitem{ref74}
	P. Lindenberger, P. -E. Sarlin and M. Pollefeys, “LightGlue: Local Feature Matching at Light Speed,” in \textit{Proc. IEEE/CVF Int. Conf. Comput. Vis. (ICCV)}, Paris, France, 2023, pp. 17581-17592.

	\bibitem{ref75}
	Y. Wu, W. Ma, M. Gong, L. Su and L. Jiao, “A Novel Point-Matching method Based on Fast Sample Consensus for Image Registration,” \textit{IEEE Geosci. Remote Sens. Lett.}, vol. 12, no. 1, pp. 43-47, Jan. 2015.


\end{thebibliography}
\end{document}